\documentclass[final,3p,times]{elsarticle}

\usepackage{amssymb}
\usepackage{amsmath}
\usepackage{subfigure} 
\usepackage{booktabs}
\usepackage{multirow} 
\usepackage{hyperref}  
\usepackage{rotating}
\usepackage{array}
\usepackage{xcolor}

\journal{Knowledge-Based Systems}

\begin{document}

\begin{frontmatter}



\title{MIAFEx: An Attention-based Feature Extraction Method for Medical Image Classification}

\author[inst1]{Oscar Ramos-Soto}
\ead{oscar.ramos9279@alumnos.udg.mx}
\author[inst2]{Jorge Ramos-Frutos}
\ead{jramos.estudiantepicyt@ciatec.mx}
\author[inst3]{Ezequiel Perez-Zarate}
\ead{isaias.perez@alumnos.udg.mx}
\author[inst1]{Diego Oliva\corref{cor1}}
\ead{diego.oliva@cucei.udg.mx}
\author[inst1]{Sandra E. Balderas-Mata}
\ead{sandra.balderas@academicos.udg.mx}

\affiliation[inst1]{
    organization={Departamento de Ingeniería Electro-Fotónica, Universidad de Guadalajara, CUCEI},
    addressline={Av. Revolución 1500},
    city={Guadalajara},
    postcode={44430},
    state={Jalisco},
    country={México}
}

\affiliation[inst2]{organization={Posgrados, Centro de Innovación Aplicada en Tecnologías Competitivas},
            addressline={Omega 201}, 
            city={León},
            postcode={37545}, 
            state={Guanajuato},
            country={México}}

\affiliation[inst3]{
    organization={School of Computer Science and Technology, Zhejiang Gongshang University},
    city={Hangzhou},
    postcode={310018},
    state={Zhejiang},
    country={China}
}

\cortext[cor1]{Corresponding author}

\begin{abstract}
Feature extraction techniques are crucial in medical image classification; however, classical feature extractors, in addition to traditional machine learning classifiers, often exhibit significant limitations in providing sufficient discriminative information for complex image sets. While Convolutional Neural Networks (CNNs) and Vision Transformer (ViT) have shown promise in feature extraction, they are prone to overfitting due to the inherent characteristics of medical imaging data, including small sample sizes or high intra-class variance. In this work, the Medical Image Attention-based Feature Extractor (MIAFEx) is proposed, a novel method that employs a learnable refinement mechanism to enhance the classification token within the Transformer encoder architecture. This mechanism adjusts the token based on learned weights, improving the extraction of salient features and enhancing the model’s adaptability to the challenges presented by medical imaging data. The MIAFEx output feature quality is compared against classical feature extractors using traditional and hybrid classifiers. Also, the performance of these features is compared against modern CNN and ViT models in classification tasks, demonstrating their superiority in accuracy and robustness across multiple complex medical imaging datasets. This advantage is particularly pronounced in scenarios with limited training data, where traditional and modern models often struggle to generalize effectively. The source code of this proposal can be found at  \href{https://github.com/Oscar-RamosS/Medical-Image-Attention-based-Feature-Extractor-MIAFEx}{github.com/Oscar-RamosS/Medical-Image-Attention-based-Feature-Extractor-MIAFEx}.
\end{abstract}



\begin{keyword}
Medical image classification \sep Feature extraction \sep Transformer encoder architecture
\end{keyword}

\end{frontmatter}

\section{Introduction}

Automated image classification in the medical field significantly improves diagnostics and healthcare by separating image sets into distinct categories corresponding to various diseases or conditions. This process, usually done by artificial intelligence and Machine Learning (ML) systems, provides medical professionals with a tool to identify diseases accurately \cite{ahmed2020artificial} through imaging techniques such as Magnetic Resonance Imaging (MRI),  Computed Tomography (CT) scans, among many others \cite{rana2023machine}. Nevertheless, access to large and labeled image datasets is often a significant challenge \cite{an2023comprehensive}. These sets are frequently limited in image quantity due to privacy concerns and the specialized nature of medical expertise required to annotate images. For example, medical fields such as ophthalmology, radiology, and dermatology may consist of only a few hundred or thousands images, which is significantly less than the millions of images typically used to train general-purpose computer vision models \cite{tsipras2020imagenet}. Additionally, many medical images do not provide explicit visual features, making feature extraction challenging.


In computer vision, image feature extraction is a fundamental tool used to capture essential characteristics of an image in terms of numerical values \cite{nixon2019feature}, enabling classifiers to differentiate between classes or objects effectively by identifying and quantifying key visual attributes such as edges, textures, shapes, or color patterns. These features represent the image in a lower-dimensional space, allowing ML and hybrid models to process and classify images based on the extracted information. Effective feature extraction ensures that classifiers can generalize well to unseen data and avoid overfitting, especially in tasks with complex datasets or limited training data. When applied to medical datasets, classical feature extractors face significant problems due to the complexity of the images. These traditional techniques are specifically designed to capture simple local patterns like edges and gradients, but still struggle to detect the high-level features often needed in medical imaging tasks, such as detecting small lesions or understanding tissue structures \cite{li2018large}. Additionally, they cannot capture global context and relationships between different parts of an image, which is crucial for interpreting medical scans. These methods are also sensitive to noise and artifacts \cite{routray2017analysis}, and they do not scale well to the high-dimensional nature of medical data. As a result, these handcrafted methods often lead to poor performance in these feature classifications when compared to more modern approaches, such as approaches like Deep Learning (DL). 

In DL models such as Convolutional Neural Networks (CNNs) and Vision Transformers (ViT), overfitting and limited generalization are the main issues that arise when they are trained on tiny medical datasets \cite{li2020analyzing}. 
The latter issue leads the model to ``memorize'' the training images instead of learning patterns that may be applied to other datasets. Because of this problem, a model may perform well on a certain dataset portion but poorly when used with additional patient images or in different imaging conditions. The CNNs rely heavily on large datasets to learn robust and meaningful features from the data. However, when training on small medical datasets, the model may learn small correlations or noise, leading to overfitting. Similarly, ViTs, which rely on self-attention mechanisms to capture global dependencies within an image, can struggle when insufficient data allows the model to learn the relationships between patches effectively. The nature of medical images complicates this issue, as they do not often provide easily identifiable features that a model can extract and learn from. Unlike natural images, where objects like cars, animals, or buildings have clear edges, shapes, and textures, as the ones in which modern techniques are trained, medical images can be highly abstract and require domain expertise to interpret. 

This work presents the Medical Image Attention-based Feature Extractor (MIAFEx), a novel approach that enhances feature extraction by refining the classification token ([CLS] token) within the Transformer encoder. This refinement mechanism dynamically adjusts this token using learnable weights, allowing the model to prioritize salient features, crucial for medical image classification. The main contributions of this work can be summarized as follows:

\begin{itemize} 
    \item \textbf{Introduction of the MIAFEx framework:} A novel approach to medical image classification for complex, small, and imbalanced medical datasets. 

     \item \textbf{Learnable refinement mechanism:} The MIAFEx introduces a dynamic and learnable refinement mechanism that adjusts the [CLS] token during training, enabling the model to focus on the most relevant features for classification, improving model performance.
     
    \item \textbf{Comparative analysis:} The proposed method is compared against classical feature extraction techniques combined with ML classifiers, emphasizing its effectiveness. 

    \item \textbf{Extensive experimental validation:} The obtained results were evaluated across diverse medical imaging datasets, presenting superior performance compared to traditional and modern methods like CNNs and ViT, especially on smaller datasets. 
    
\end{itemize}

The remainder of this paper is organized as follows: Section \ref{sec:relat} reviews related work relevant to this study, while Section \ref{sec:miafex} provides a detailed explanation of the MIAFEx framework, detailing its structure and functionality. Section \ref{sec:experimental} outlines the experimental framework designed to evaluate the performance of MIAFEx, including a description of the dataset and the techniques employed for comparative analysis. Section \ref{sec:results} presents comprehensive results from the evaluation, highlighting the method's effectiveness through rigorous comparison. Finally, Section \ref{sec:conclusions} summarizes the findings and conclusions of this work, along with potential directions for future research.

\section{Related work}
\label{sec:relat}
In recent decades, feature extraction techniques have played an important role in medical image classification for disease detection. Classical methods for feature extraction, such as Scale-Invariant Feature Transform (SIFT) \cite{el2013comparative}, Local Binary Patterns (LBP) \cite{he1990texture}, and Histogram of Oriented Gradients (HOG) \cite{dalal2005histograms}, have long been used in traditional computer vision and also in the medical area \cite{chowdhary2020segmentation}.  The performance of capturing the most relevant image features highly impacts the classifier's accuracy. Numerous approaches have been developed for a wide range of medical imaging applications using these classical techniques, among many others. Some of the most relevant approaches are presented below.

\subsection{Classical and hybrid approaches}
In 2018, Alhindi et al. \cite{alhindi2018comparing} compared LBP, HOG, and deep features for histopathological image classification, finding that the best performance was achieved using LBP combined with a Support Vector Machine (SVM), with an accuracy of 90.52\% across 20 classes. That same year, Liebgott et al. publicly released a robust toolbox called ImFEATbox \cite{liebgott2018imfeatbox}, providing MATLAB users with a comprehensive set of local and global feature descriptors, including geometrical, intensity, region, texture, and moment-based features. ImFEATbox was tested on MRI prostate cancer and FDG-PET/CT lung cancer images, performing statistical analyses using an SVM. Concurrently, Raj et al. \cite{raj2020optimal} utilized both the Gray-level Co-occurrence Matrix (GLCM) and Gray-level Run Length Matrix (GLRLM) to extract features from three medical datasets to differentiate between Alzheimer’s disease and various cancers, selecting 15 features for classification using the metaheuristic-based Opposition Crow Search (OCS).

In 2021, Aswiga et al. \cite{rv2021augmenting} tackled the issue of limited datasets by introducing Feature Extraction Based Transfer Learning (FETL), which combines three different feature extraction techniques to enhance basic multilevel transfer learning (MLTL) for mammography datasets, classifying benign, malignant, and normal images.

Several proposals also demonstrate successful hybridization strategies to address feature extraction challenges in medical image datasets. For instance, metaheuristic optimization has been widely applied to reduce feature complexity and aid in feature selection. In 2016, Nagarajan et al. \cite{nagarajan2016hybrid} proposed a hybrid genetic algorithm using branch and bound techniques and the artificial bee colony algorithm, applied to thyroid, breast cancer, and brain tumor images. Their approach incorporated the Intrinsic Pattern Extraction Algorithm (IPEA), Texton-based Contour Gradient Extraction (TCGR), and SIFT for robust feature extraction.

In 2018, Echegaray et al. \cite{echegaray2018quantitative} introduced the Quantitative Image Feature Engine (QIFE) for volumetric medical image feature extraction using multiprocessor parallelization, primarily leveraging Haralick features and GLCM applied to DICOM files. In 2022, Kuwil \cite{kuwil2022new} developed the Feature Extraction Based on Region of Mines (FE\_mines), which extracts features from the black, green, and red channels based on statistical measurements such as mean, standard deviation, and coefficient of variance. It was tested on MRI, X-ray, and eye fundus images, achieving high performance across all types. Also in 2023, Sharma et al. \cite{sharma2023hog} hybridized HOG with a variant of ResNet-50 for brain tumor detection in MRI images, achieving an accuracy of 88\%, outperforming several other techniques. That same year, Narayan et al. \cite{narayan2023fuzzynet} introduced FuzzyNet, a model for image classification that uses GLCM features and particle swarm optimization to perform binary differentiation in musculoskeletal radiographs, incorporating noise reduction and contrast enhancement before model implementation.

Ranjitha and Pushphavathi (2024) proposed a Gaussian-Wiener filter to reduce noise in images while maintaining their quality. Feature extraction from preprocessed images is performed using the GLCM technique, identifying texture patterns in tumors in the images, improving the classification performed by ML and DL algorithms~\cite{ranjitha2024analysis}. In a 2022 work reported by Hage Chehade et al.~\cite{hage2022lung}, statistical features, GLCM, and Hu invariant moments were used to classify lung and colon cancer cases. Combining these techniques achieved an accuracy of $99\%$ and an F1-score of $98.8\%$ using XGBoost. HOG and LBP are used for brain tumor image classification in the work reported by Basthikodi et al. in 2024~\cite{basthikodi2024enhancing}. These feature extraction techniques achieve competitive classification metrics when combined with PCA and SVM classification.

\subsection{CNN and ViT approaches}

On the other hand, DL-based methods such as convolutional layers in CNNs and self-attention mechanisms in ViT have gained interest for feature extraction in modern models, hence the number of papers regarding the classification of medical images using CNN or ViT models and variants has highly increased  \cite{mauricio2023comparing}. 

One of the most relevant works is MedViT, presented by Manzari et al. \cite{manzari2023medvit}. This proposal introduces a CNN-Transformer hybrid model designed to reduce model complexity while enhancing high-level information by considering feature variance and mean on large-scale datasets, such as those in MedMNIST-2D. In the same year, Ding et al. developed FTransCNN \cite{ding2023ftranscnn}, which combines fuzzy logic, CNN, and transformer architectures to enhance local details and eliminate irrelevant regions. Their model incorporates a Fuzzy Attention Fusion Module (FAFM) to capture both high and low-level features in chest X-ray and endoscopy images, achieving high performance on both datasets. More recently, in 2024, Huo et al. introduced the HiFuse model \cite{huo2024hifuse}, which employs a parallel hierarchical structure, an adaptive Hierarchical Feature Fusion (HFF) block, and an Inverted Residual Multi-layer Perceptron (IRMLP) to build a robust model that avoids introducing noise. HiFuse was tested on various image types, including skin lesions, endoscopic, chest CT, and microscopic images. In the same year, Yue and Li designed the MedMamba \cite{yue2024medmamba}, mainly based on the hybrid basic block named SS-Conv-SSM, which implies using convolutional layers for local feature extraction. It was tested on sixteen datasets with 411,007 images across ten different imaging modalities. Another work incorporating neural networks for feature extraction is attributed to Maheswari and Gopalakrishnan in 2024. They propose an Advanced Attention Capsule Network (AACNet) incorporating a multi-level capsule network, dynamic channel attention modules, and a multi-feature extractor with an SPP layer, experimenting on X-ray and CT images~\cite{maheswari2024dynamic}. Remzan et al. in 2024 proposed feature extraction using six pre-trained convolutional neural network architectures for different brain images: Positron Emission Tomography (PET), Magnetic Resonance Imaging (MRI), and Computed Tomography (CT). The six architectures were DenseNet-121, VGG-19, Inception V3, MobileNet, EfficientNetB1, ResNet-50, and EfficientNetV2B1. Of these six architectures, only the three best ones are used for feature extraction. They managed to identify the three architectures that most frequently achieved accuracy above $96\%$; these architectures were VGG-19, EfficientNetV2B1, and ResNet-50~\cite{remzan2024ensemble}. In another paper published in 2024~\cite{navaneethan2024enhancing}, Navaneethan and Devarajan developed a method for diabetic retinopathy detection using fundus image datasets. A CNN feature extraction process is employed, and these features are augmented using a Generative Adversarial Network (GAN) to assist the classifiers in delivering better results. Finally, another recent study published by G{\"u}ler and Naml{\i} on brain tumor detection using MR images achieved $100\%$ accuracy in classification. Four architectures were used as feature extractors, corresponding to VGG, ResNet, DenseNet, and SqueezeNet. The extracted features are subsequently classified using less complex classification algorithms~\cite{guler2024brain}.

As a complement to what was mentioned above, a summary of the work carried out on classification in medical images is shown in Table~\ref{tab:comparison_studies}. Seven types of images are observed and classified using classical descriptors and CNN-based descriptors. In this way, the classification work on each type of image becomes clearer.

\begin{table}[h]
\centering
\caption{Detailed comparison of MIAFEx with previous studies across medical imaging domains.}
\label{tab:comparison_studies}
\resizebox{0.85\textwidth}{!}{%
\begin{tabular}{llcccc}
\hline
\textbf{Medical Domain} & \textbf{Author(s)} & \textbf{Year} & \textbf{Method} & \textbf{Accuracy (\%)} & \textbf{Reference} \\
\hline
\multirow{3}{*}{Histological Biopsy} 
  & Alhindi et al. & 2018 & LBP + SVM & 90.52 & \cite{alhindi2018comparing} \\
  & Raj et al. & 2018 & GLCM + OCS & 91.00 & \cite{raj2020optimal} \\
\hline
\multirow{3}{*}{Ocular Alignment} 
  & Kuwil & 2022 & FE-Mines + SVM & 82.00 & \cite{kuwil2022new} \\
  & Ranjitha et al. & 2024 & GLCM + ML/DL & 85.00 & \cite{ranjitha2024analysis} \\
\hline
\multirow{3}{*}{Eye Fundus} 
  & Navaneethan et al. & 2024 & CNN + GAN & 91.00 & \cite{navaneethan2024enhancing} \\
  & Kuwil & 2022 & FE-Mines + ML & 85.00 & \cite{kuwil2022new} \\
\hline
\multirow{3}{*}{Breast Ultrasound} 
  & Sharma et al. & 2023 & HOG + ResNet-50 & 88.00 & \cite{sharma2023hog} \\
  & Nagarajan et al. & 2016 & (GA, ABC) + (SIFT + IPEA + TCGR) & 85.00 & \cite{nagarajan2016hybrid} \\
\hline
\multirow{3}{*}{Chest CT} 
  & Echegaray et al. & 2018 & QIFE + SVM & 92.00 & \cite{echegaray2018quantitative} \\
  & Chehade et al. & 2022 & GLCM + XGBoost & 99.00 & \cite{hage2022lung} \\

\hline
\multirow{3}{*}{Brain MRI} 
  & Remzan et al. & 2024 & VGG-19, ResNet-50, EffNetV2B1 & 96.00 & \cite{remzan2024ensemble} \\
  & G{\"u}ler and Naml{\i} & 2024 & VGG, ResNet, DenseNet, SqueezeNet + ML & 100.00 & \cite{guler2024brain} \\
\hline
\multirow{3}{*}{Gastrointestinal Endoscopy} 
  & Huo et al. & 2024 & HiFuse (HFF + IRMLP) & 92.00 & \cite{huo2024hifuse} \\
  & Yue and Li & 2024 & MedMamba (SS-Conv-SSM) & 94.00 & \cite{yue2024medmamba} \\
\hline
\end{tabular}%
}
\end{table}

As observed, several previously proposed methods are related to medical image feature extraction and classification. However, most of them are tested on large datasets or use several feature extractors, improving the classification performance in classical, hybrid, and DL approaches. Also, the best-performing ones require significant computational resources to obtain excellent results, making clinical implementation difficult.

\section{MIAFEx design}
\label{sec:miafex}
This section describes the MIAFEx in detail, emphasizing the refinement mechanism. This approach generally processes input train images, dividing them into patches and extracting feature classification embeddings [CLS] through a transformer encoder using the ViT architecture. Then, this [CLS] token enters the refinement mechanism, first multiplying it with learnable refinement weights to prioritize relevant features. This refined token is passed through a fully connected layer, and a softmax function processes the output to produce class probabilities. The model is trained using cross-entropy loss, and during backpropagation, both the refinement weights and the model parameters are updated to improve classification performance. This is defined as the training part. Then, the trained model is used for a test dataset refined feature extraction and for a later feature selection using any classifier, named the inference step. The above-described general MIAFEx process is summarized in Fig. \ref{fig:gen_diag}. A more detailed explanation is presented below.

\begin{figure}[htb!]
    \centering
    \includegraphics[width=0.8\linewidth]{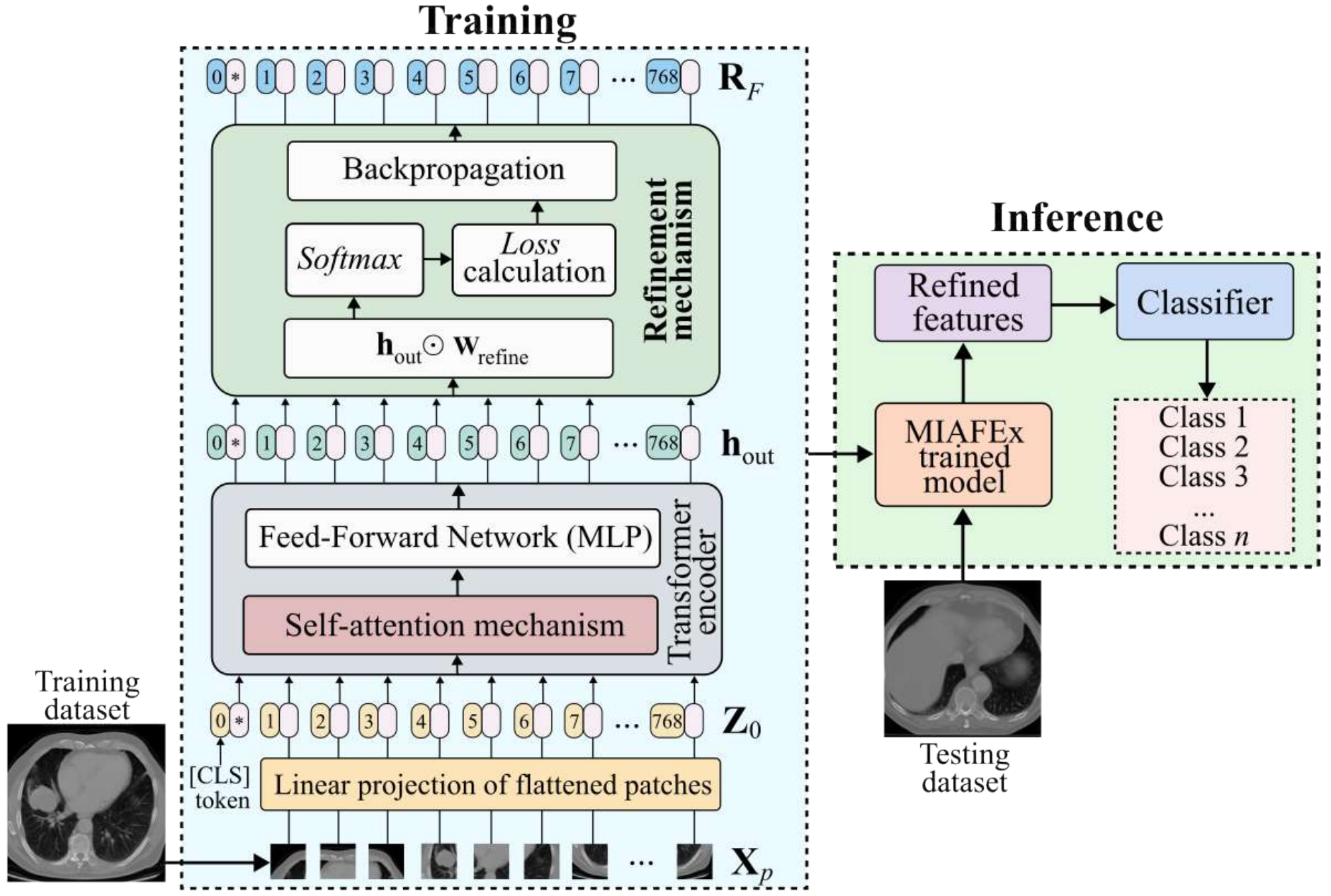}
    \caption{General diagram of the MIAFEx.}
    \label{fig:gen_diag}
\end{figure}

\subsection{Patch embedding}

In the original ViT model \cite{dosovitskiy2020image} (whose core element is the transformer encoder), the input image $\mathbf{X} \in \mathbb{R}^{H \times W \times C}$ (where $H$, $W$, and $C$ represent the height, width, and number of channels, respectively) is divided into fixed-size patches. Each patch is flattened and linearly projected into a $D$-dimensional vector. Let $\mathbf{X}_p \in \mathbb{R}^{N \times D}$ represent the sequence of $N$ patch embeddings, where $N = \frac{H \cdot W}{P^2}$ is the total number of patches, and $P \times P$ is the patch size.

In addition to the patch embeddings, a learnable [CLS] token is prepended to the sequence, which serves as a summary representation for the entire image. Mathematically, the input to the transformer encoder is:

\begin{equation}
    \mathbf{Z}_0 = [\text{CLS}; \mathbf{X}_p] + \mathbf{E}_{\text{pos}}
\end{equation}

\noindent where $\mathbf{Z}_0 \in \mathbb{R}^{(N+1) \times D}$ is the initial input, and $\mathbf{E}_{\text{pos}}$ represents position embeddings that encode the location of each patch.

In this approach, the selected input image size is $224 \times 224 \times 3$ (height, width, and RGB channels). Then, the image is divided into patches of size $16 \times 16$, and each patch is treated as a token, similar to how tokens are used in natural language processing with transformers. After flattening each patch, a vector of size $P^2 \times C$ is obtained. In this case, the size of each flattened patch is 
$16 \times 16 \times 3 = 768$. These flattened patches are then linearly projected into a $D$-dimensional space in ViT's base version. This projection ensures that all patches have a consistent embedding dimension that the transformer can process.

\subsection{Transformer encoder}

The transformer encoder consists of $L$ layers, each composed of a multi-head self-attention mechanism and a Feed-forward Network, in this case, a multi-layer perceptron (MLP). The [CLS] token and the patch embeddings are updated in each layer based on interactions between all tokens. Let $\mathbf{Z}_L \in \mathbb{R}^{(N+1) \times D}$ represent the output of the $l$-th transformer layer. After processing through all $L$ layers, the output corresponding to the [CLS] token is extracted as the global representation of the image:

\begin{equation}
    \mathbf{Z}_L = [\mathbf{h}_{\text{out}}; \mathbf{X}_{p, \text{out}}]
\end{equation}

\noindent where $\mathbf{h}_{\text{out}}$ is the final output of the [CLS] token after it has interacted with all the patch tokens through the transformer layers.

The processed [CLS] token, from now on referred to in this document as $\mathbf{h}_{\text{out}}$, works as a summary token that gathers global information from all patches via the self-attention mechanism. It serves as both an input (to initiate the encoding process) and as an output, representing the holistic image features. The $\mathbf{h}_{\text{out}}$ token is passed to the final classification head or further refined using additional mechanisms to optimize performance.

\subsection{Refinement mechanism}

After passing through $L$ transformer encoder layers, the output corresponding to the $\mathbf{h}_{\text{out}}$ token is extracted from the last layer, mathematically defined as:

\begin{equation}
\mathbf{h}_{\text{out}} = [h_1, h_2, \dots, h_{D}]
\end{equation}

\noindent where $D$ is the dimensionality of the token, in this approach $D = 768$, each $h_i$ represents a scalar feature corresponding to the $i$-th dimension of the $\mathbf{h}_{\text{out}}$ token.

In the refinement mechanism, a set of learnable weights $\mathbf{w}_{\text{refine}}$ is introduced, which is a vector of the same dimensionality as the $\mathbf{h}_{\text{out}}$. These weights are initialized randomly and are trained along with the rest of the model. The refinement weights can be represented as:

\begin{equation}
    \mathbf{w}_{\text{refine}} = [w_1, w_2, \dots, w_D]
\end{equation}

Each $w_i$ is a scalar weight corresponding to the $i$-th dimension of the $\mathbf{h}_{\text{out}}$ token. These weights allow the model to amplify, attenuate, or ignore certain features in the $\mathbf{h}_{\text{out}}$ during training. The refinement process involves applying an element-wise multiplication between the $\mathbf{h}_{\text{out}}$ and the refinement weights $\mathbf{w}_{\text{refine}}$. This operation modifies each dimension of the $\mathbf{h}_{\text{out}}$ token according to the learned importance of that feature. Mathematically, this is expressed as:

\begin{equation}
    \mathbf{R}_F = \mathbf{h}_{\text{out}} \odot \mathbf{w}_{\text{refine}}
\end{equation}

\noindent where $\odot$ represents element-wise multiplication. In expanded form, this becomes:

\begin{equation}
    \mathbf{R}_F = [h_1 \cdot w_1, h_2 \cdot w_2, \dots, h_D \cdot w_D]
\end{equation}

Each element of the $\mathbf{h}_{\text{out}}$ token is scaled by its corresponding refinement weight in $\mathbf{w}_{\text{refine}}$. The refinement weights allow the model to prioritize certain features based on their relevance to the task. This refined feature vector $\textbf{R}_F$ is passed to a fully connected layer for classification:

\begin{equation}
    \hat{y} = \text{softmax}(a \cdot \mathbf{R}_F + b)
\end{equation}

\noindent where $a$ and $b$ are the weights and bias of the classifier, and $\hat{y}$ represents the predicted class probabilities.

To train the model, the predicted class probabilities $\hat{y}$ are compared to the true labels using the cross-entropy loss function, defined as:

\begin{equation}
    \mathcal{L}_{\text{CE}}(\hat{y}, y) = -\sum_{i=1}^{C} y_i \log(\hat{y}_i)
\end{equation}

\noindent where $y$ is the one-hot encoded true label, $C$ is the number of classes, and $\hat{y}_i$ is the predicted probability for class $i$. This loss function penalizes the model based on how far the predicted probabilities are from the true labels. After computing the loss, backpropagation is performed to update the learnable parameters of the model, including the refinement weights $\mathbf{w}_{\text{refine}}$, as well as the weights and biases in the classifier. Through multiple training iterations, this process allows the model to learn and refine the features extracted from the $\mathbf{h}_{\text{out}}$ token, improving classification accuracy over time. The final refined hidden state, $\mathbf{R}_F$, now represents the final feature vector extracted.

The refinement mechanism can be understood as a process that adjusts the contribution of each feature dimension in the $\mathbf{h}_{\text{out}}$ token. If a refinement weight $w_i$ is close to $0$, the corresponding feature $h_i$ will be suppressed. Conversely, if $w_i > 1$, the feature $h_i$ will be amplified. The key idea behind it is to allow the model to adjust the importance of different token dimensions dynamically. Each dimension in the hidden state represents a specific feature of the input image, and not all features are equally relevant to the classification task. By learning the refinement weights, the model can down-weight irrelevant features and up-weight the important ones. 

The refinement mechanism plays a crucial role in adjusting the feature representation of the transformer encoder output, ensuring that the most relevant features for the classification task are emphasized while less critical features are down-weighted. This improves the overall performance of the Transformer encoder by allowing the model to refine the learned features during training dynamically.


\section{Experimental framework}
\label{sec:experimental}
In this section, the selected datasets for the MIAFEx testing are presented, followed by the description of the ML classifiers and DL used for comparison. Also, the classification performance metrics and the experimental configuration for the model test are detailed.

\subsection{Datasets}
A selection of diverse datasets was chosen to evaluate the efficiency of the MIAFEx framework, each representing unique challenges in medical image analysis. These datasets were selected based on their varying image acquisition methods, dataset sizes, and the visual complexity of the images. Below, we present the details of the datasets used in the study.

\subsubsection{Histological biopsy}
\begin{figure}[htbp]
    \centering
    \subfigure[CLL]{
        \includegraphics[width=0.22\textwidth]{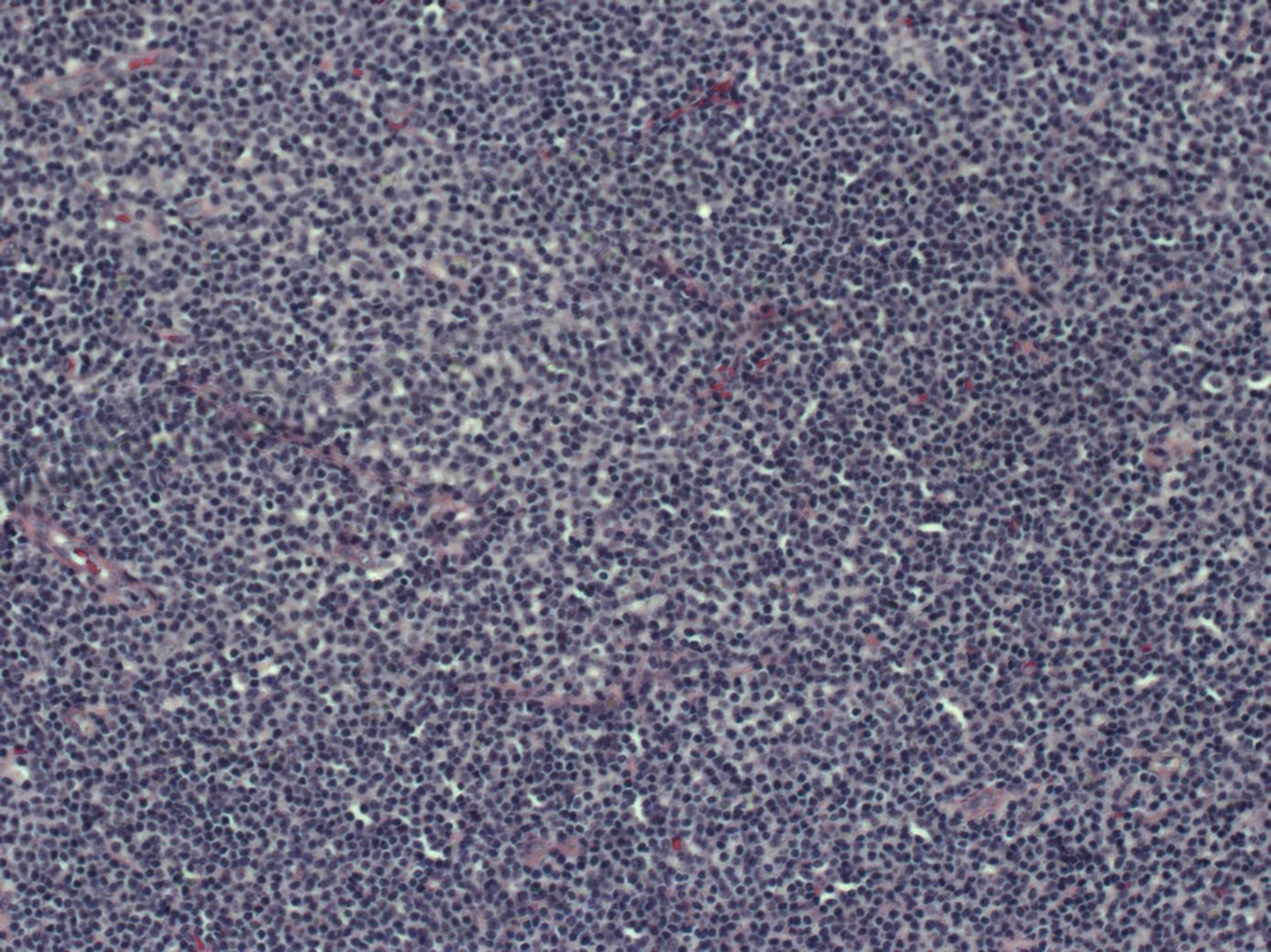}
    }
    \subfigure[FL]{
        \includegraphics[width=0.22\textwidth]{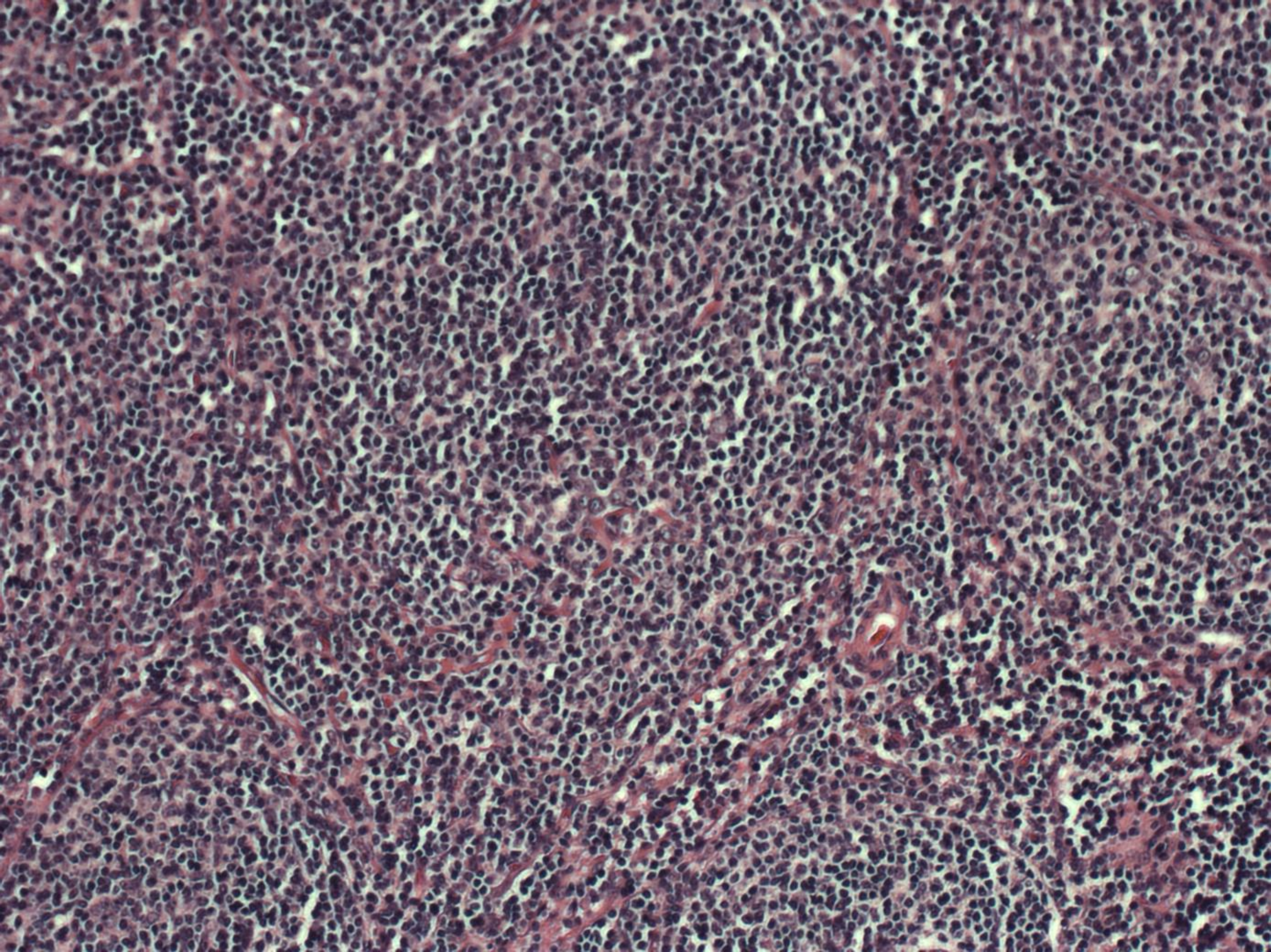}
    }
        \subfigure[MCL]{
        \includegraphics[width=0.22\textwidth]{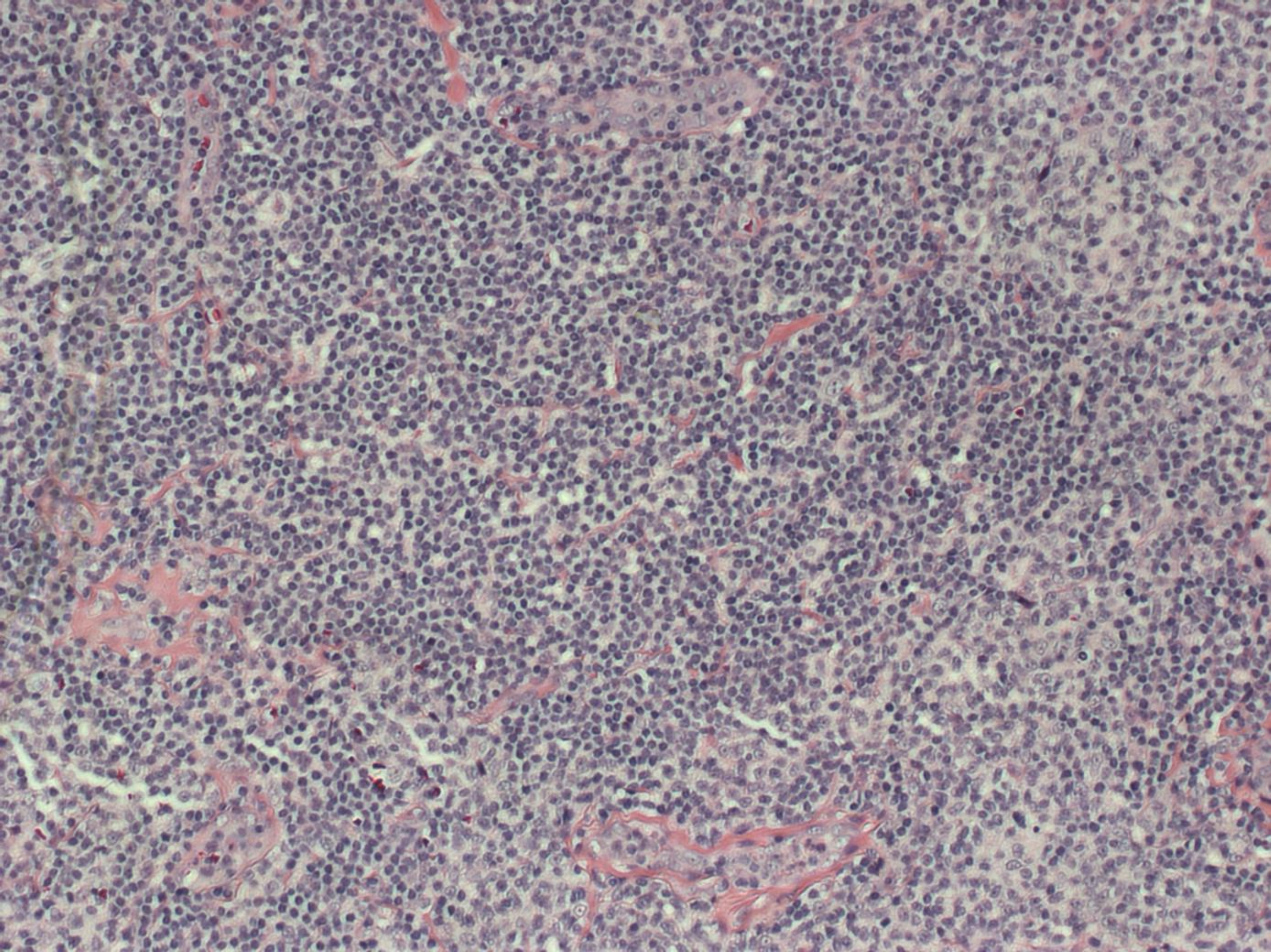}
    }
    \caption{Histological biopsy images.}
  \label{fig:Histological biopsy}
\end{figure}

The histological biopsy is a widely used microscopy-based imaging study to observe different tissues at a cellular level. The chosen dataset\footnote{https://www.kaggle.com/datasets/andrewmvd/malignant-lymphoma-classification} includes images from three classes, namely Chronic Lymphocytic Leukemia (CLL), Follicular Lymphoma (FL), and Mantle Cell Lymphoma (MCL). Images were acquired with an AxioCam MR5 CCD camera, a Zeiss Axioscope white light microscope, and $20 \text{x}$ objective. $57$ training images per class of $1040 \times 1388$ pixels were taken which were divided into $30$ mosaics of $208 \times 231$ pixels from each image giving a total of $1710$ images per class. $56$, $82$, and $65$ images were used in the CLL, FL, and MCL testing, respectively. Images of the three classes mentioned are shown in Fig.~\ref{fig:Histological biopsy}. 

\subsubsection{Ocular alignment}
\begin{figure}[htbp]
    \centering
    \subfigure[Normal]{
        \includegraphics[width=0.25\textwidth]{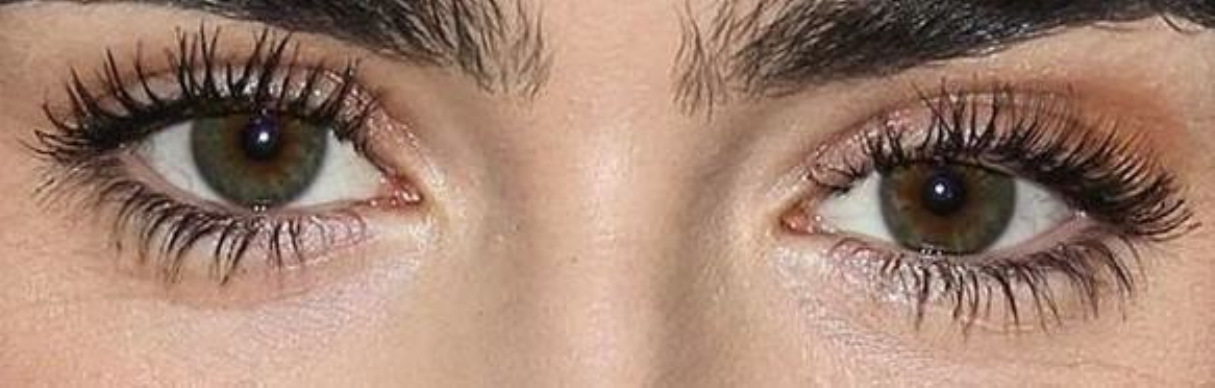}
    }
    \subfigure[Esotropia]{
        \includegraphics[width=0.318\textwidth]{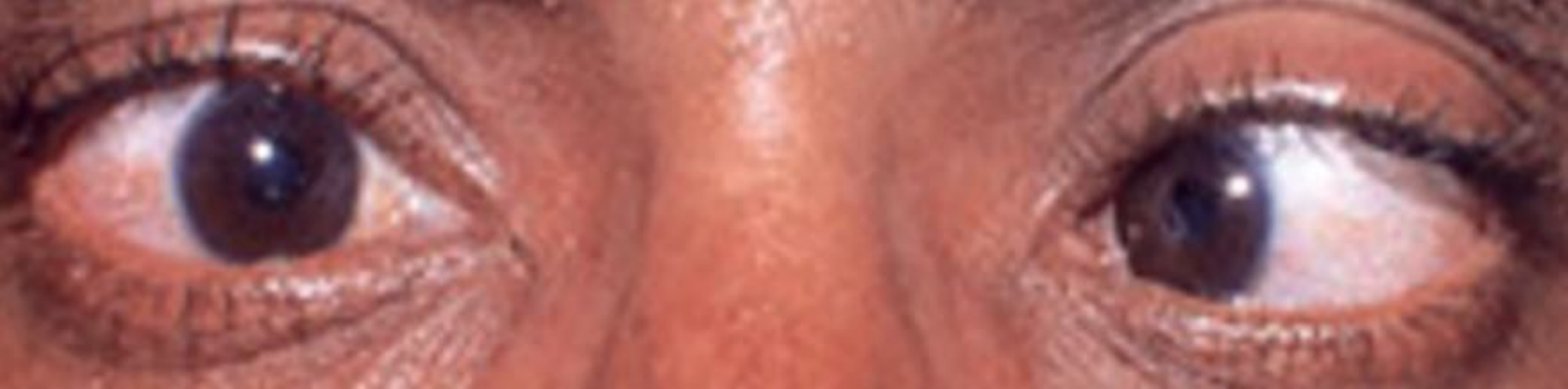}
    }
    \subfigure[Exotropia]{
        \includegraphics[width=0.25\textwidth]{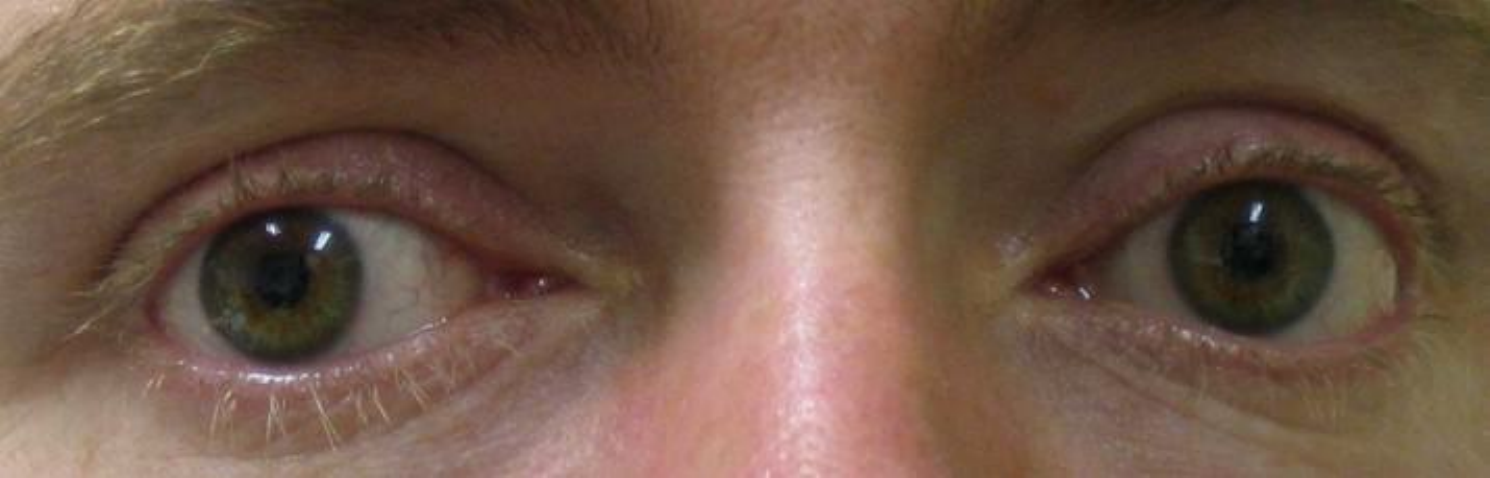}
    }
    \subfigure[Hypertropia]{
        \includegraphics[width=0.25\textwidth]{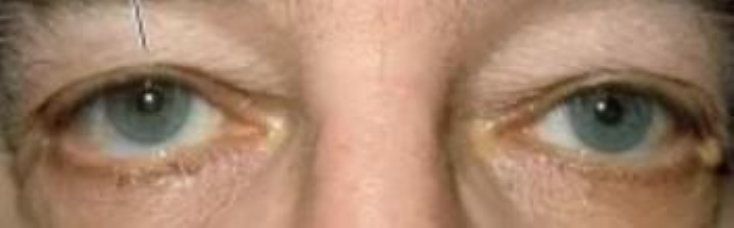}
    }
        \subfigure[Hypotropia]{
        \includegraphics[width=0.318\textwidth]{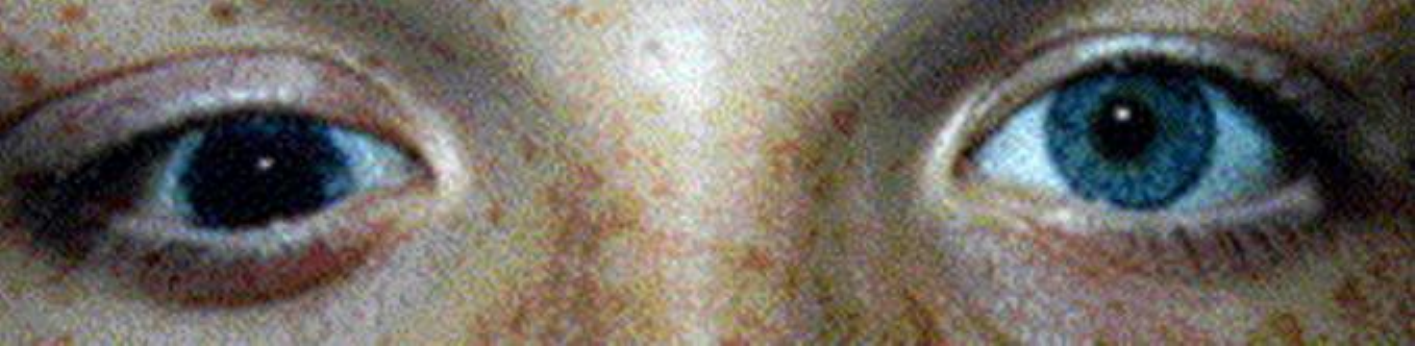}
    }
    \caption{Ocular alignment images.}
  \label{fig:ocular_alignment}
\end{figure}

Strabismus is a condition where the eyes are not aligned and can cause the brain to ignore signals from the eye or not perceive depth in the environment.  In this work, the dataset selected for strabismus\footnote{https://www.kaggle.com/datasets/ananthamoorthya/strabismus} consists of four distinct classes, each representing a specific type of ocular misalignment, along with a fifth class for correctly aligned eyes. The first is Esotropia, shown when the eyes deviate inwards (100 images). Exotropia, which causes the eyes to deviate outwards (104 images). The third is Hypertropia, which causes the eyes to deviate upwards (104 images). The last is Hypotropia, where the eyes deviate downwards (93 images). The fifth classification is where the eyes are aligned correctly (110 images). The image dataset consists of 509 total images of different sizes. Figure~\ref{fig:ocular_alignment} illustrates images representing the five specified classes.

\subsubsection{Eye fundus}
\begin{figure}[htbp]
    \centering
    \subfigure[Normal]{
        \includegraphics[width=0.22\textwidth]{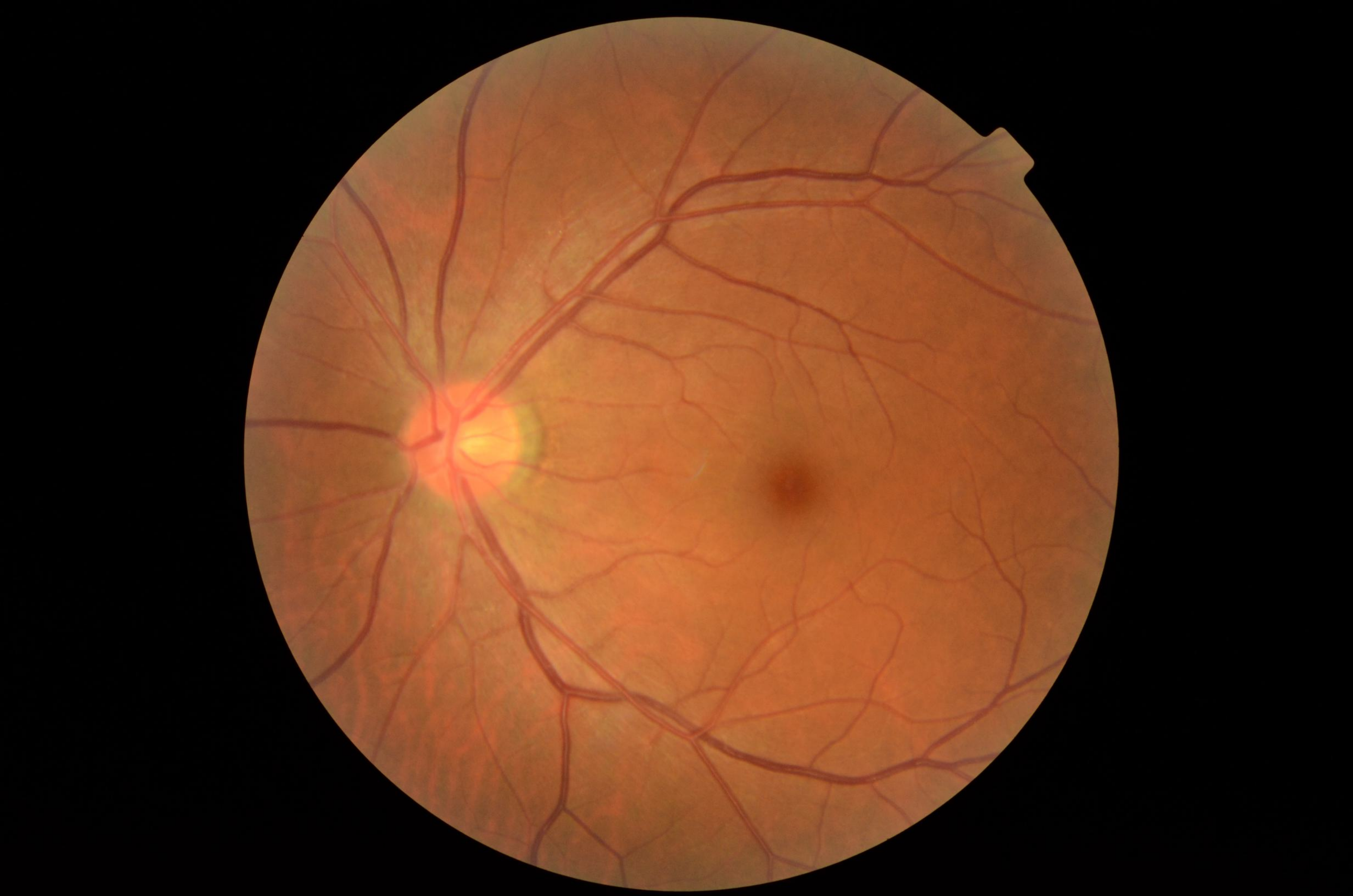}
    }
    \subfigure[Cataract]{
        \includegraphics[width=0.22\textwidth]{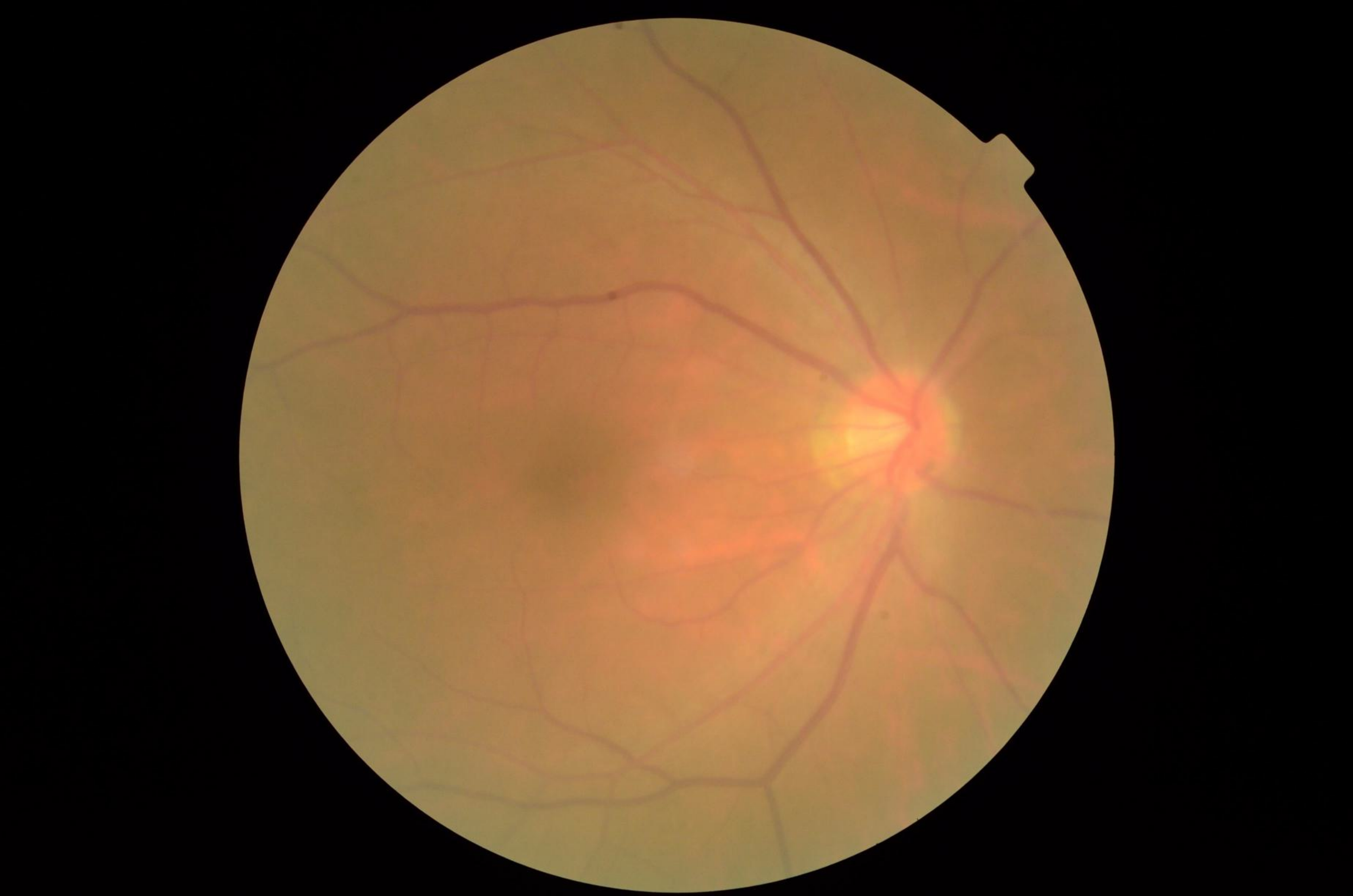}
    }
    \subfigure[Glaucoma]{
        \includegraphics[width=0.22\textwidth]{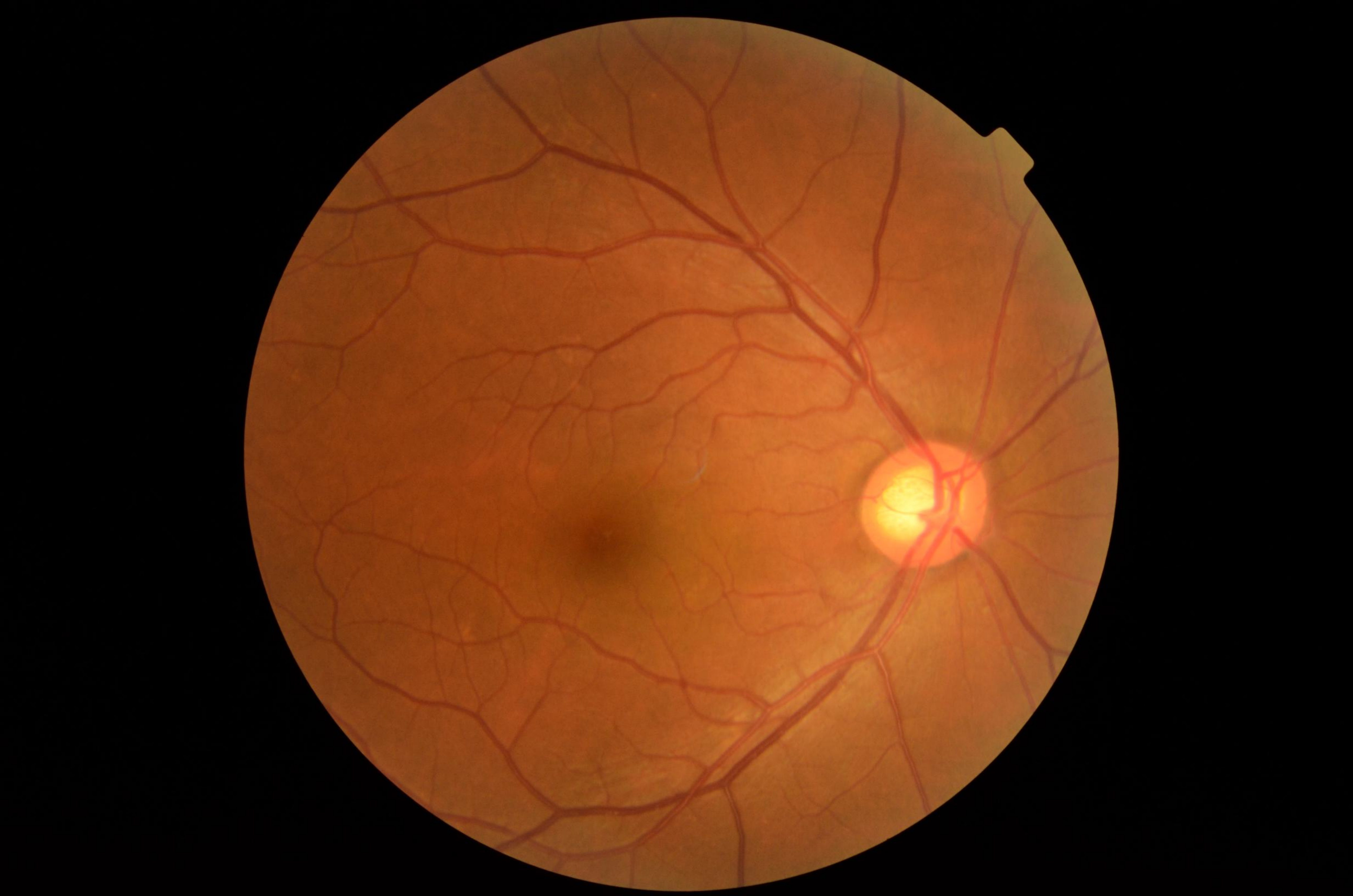}
    }
    \subfigure[Retinal disease]{
        \includegraphics[width=0.22\textwidth]{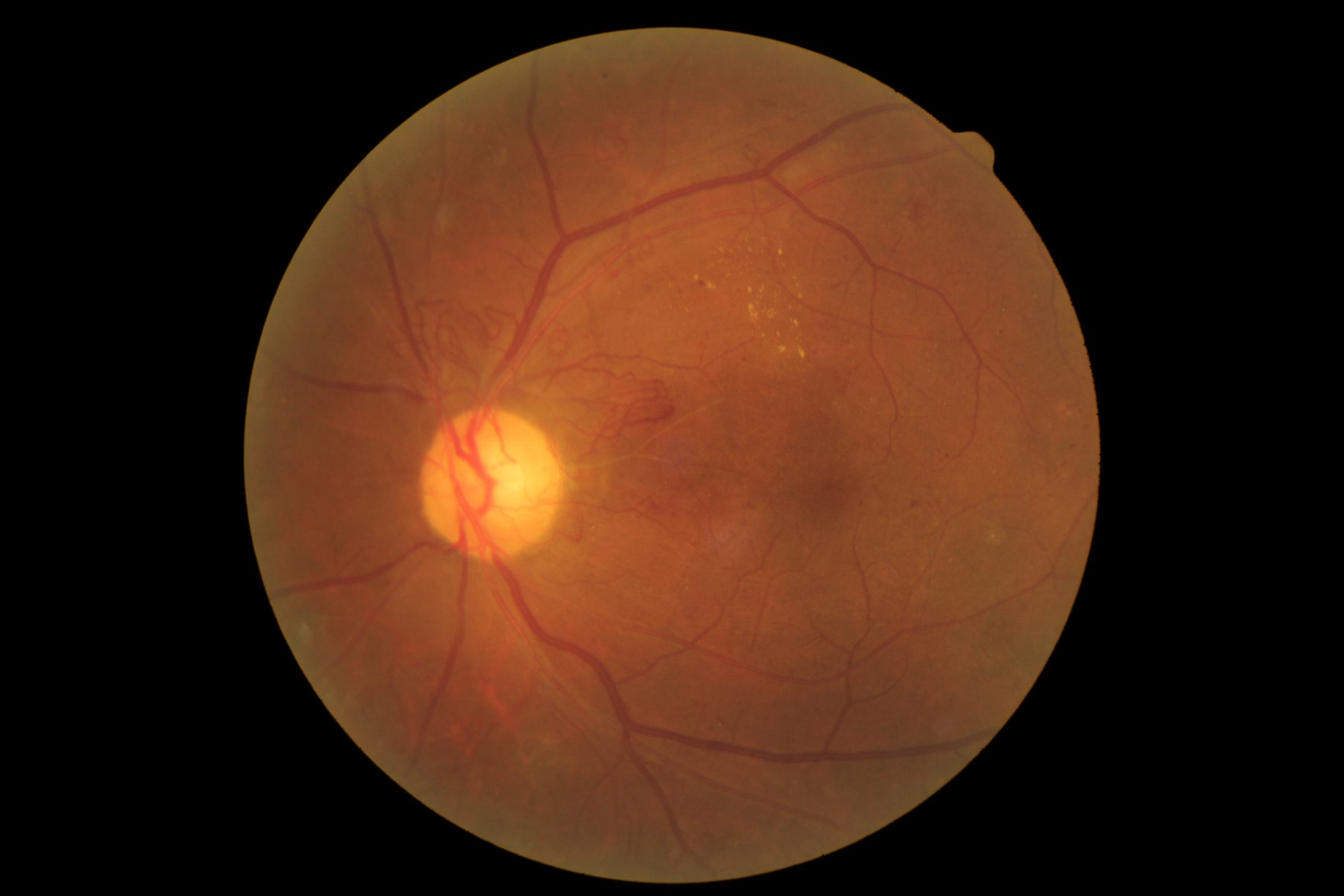}
    }
    \caption{Eye fundus images.}
  \label{fig:eye_fundus}
\end{figure}

By using an ophthalmoscope, it is possible to observe the inner part of the human eye, specifically the retina, which plays a vital role in the eye's sensorial function. This study is known as eye fundus imaging.  The dataset used\footnote{https://www.kaggle.com/datasets/jr2ngb/cataractdataset} contains eye fundus images of some diseases that affect this organ and that increase the risk of human beings going blind. There are four categories in the dataset: normal, cataract, glaucoma, and retinal disease. There are 300 images of the normal class and 300 images distributed in groups of 100 in each disease class, adding one more to the glaucoma class. The sizes of the images are varied. Images corresponding to the four mentioned classes are displayed in Fig.~\ref{fig:eye_fundus}.

\subsubsection{Ultrasound breast cancer}

\begin{figure}[htbp]
    \centering
    \subfigure[Normal]{
        \includegraphics[width=0.251\textwidth]{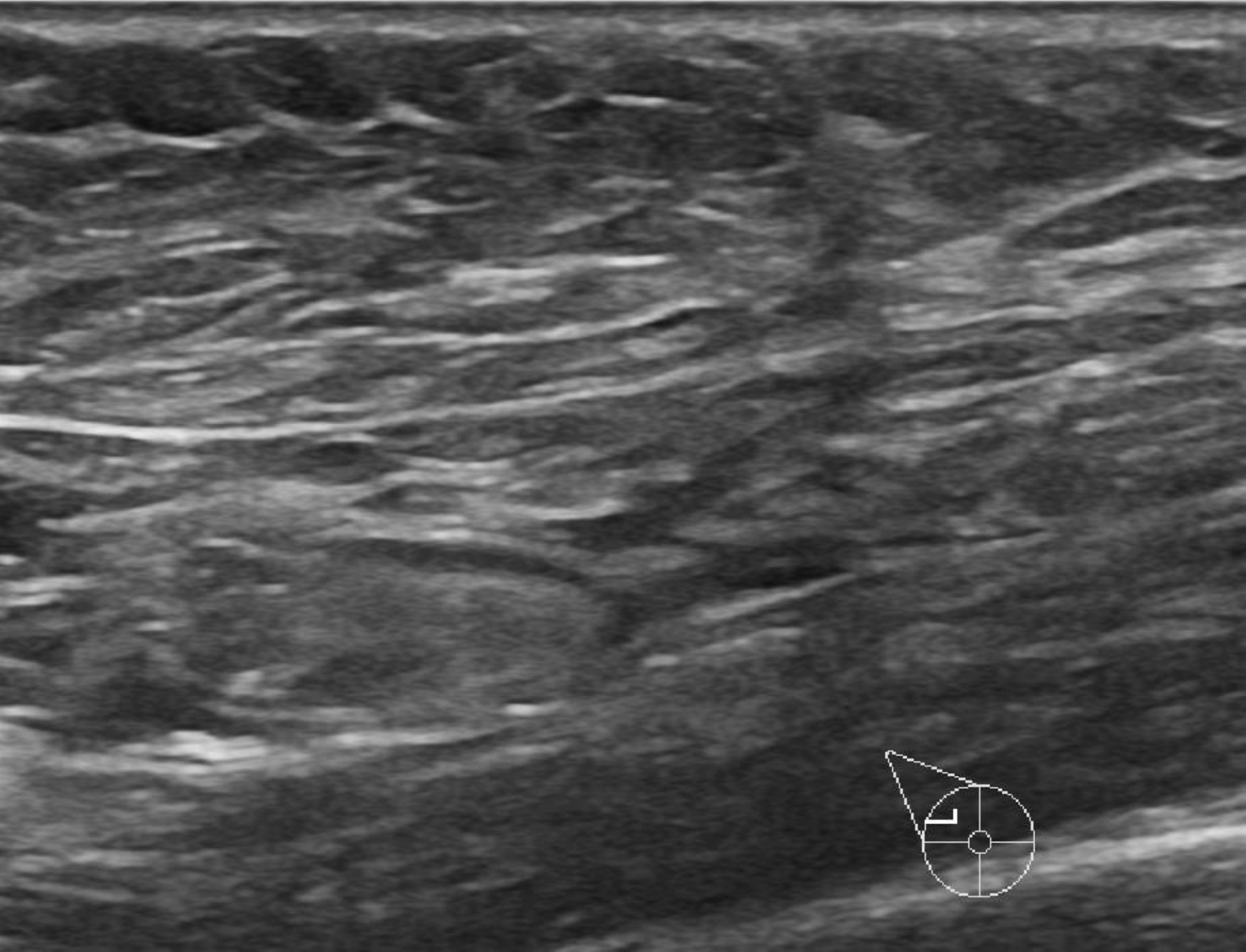}
    }
    \subfigure[Benign]{
        \includegraphics[width=0.256\textwidth]{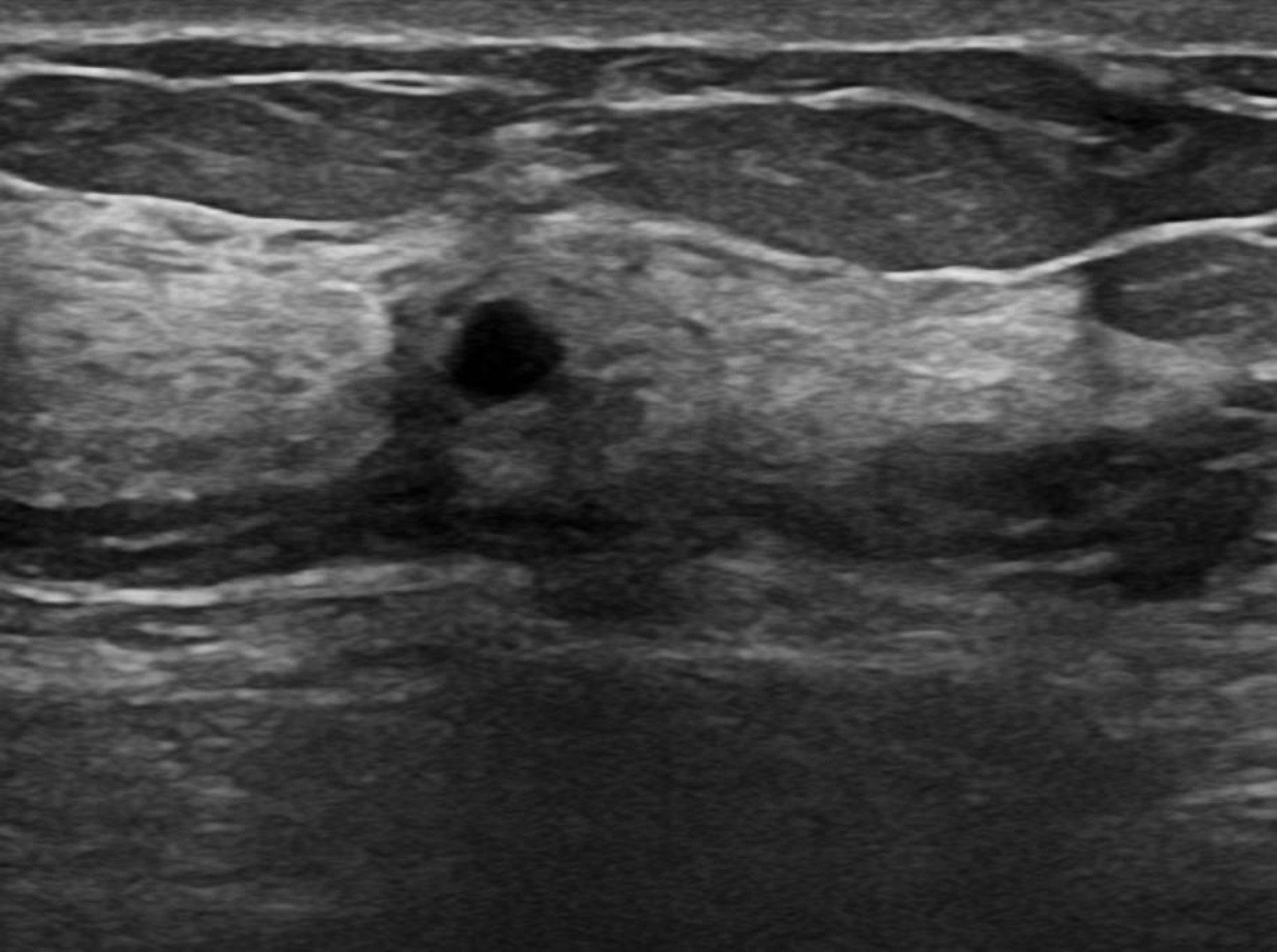}
    }
    \subfigure[Malign]{
        \includegraphics[width=0.261\textwidth]{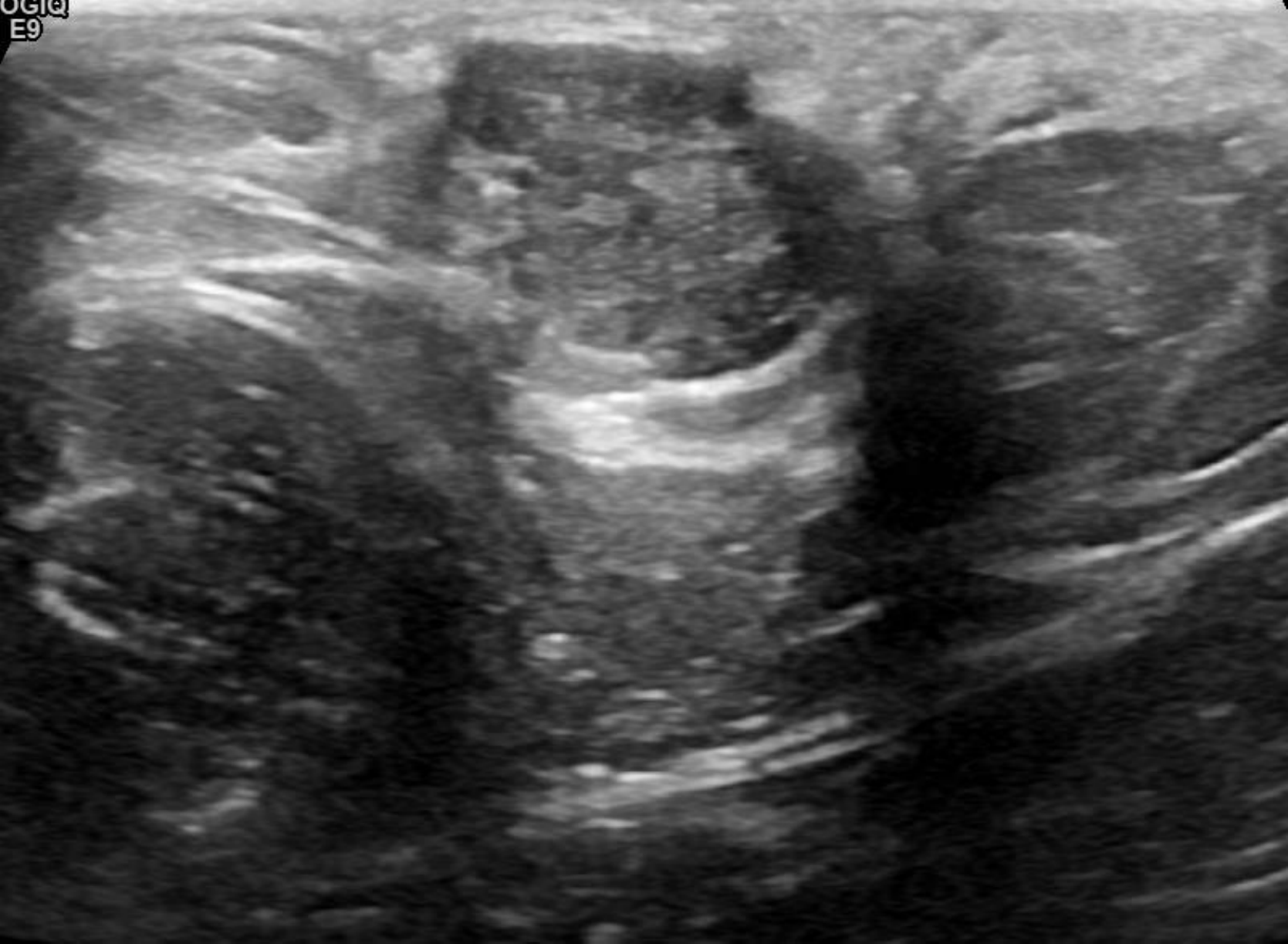}
    }
    \caption{Breast ultrasound images.}
  \label{fig:breas_us}
\end{figure}

Breast cancer has gained significant attention in recent times due to its high mortality rate and the increasing number of reported cases. Using the ultrasound imaging technique makes it possible to observe the tissue distribution for cancer visual detection, making it an early detection alternative to diminish the public health impact. The dataset\footnote{https://www.kaggle.com/datasets/sabahesaraki/breast-ultrasound-images-dataset} used in this work contains 780 images with three categories: benign with 442 images, malignant with 206, and normal with 132. The scans were collected from 600 women between 25 and 75 years old. It is a highly unbalanced dataset with different image sizes, having an average of $500\times 500$ pixels. Figure~\ref{fig:breas_us} showcases examples from each of the three described classes.


\subsubsection{Chest CT}
\begin{figure}[htbp]
    \centering
    \subfigure[Normal]{
        \includegraphics[width=0.196\textwidth]{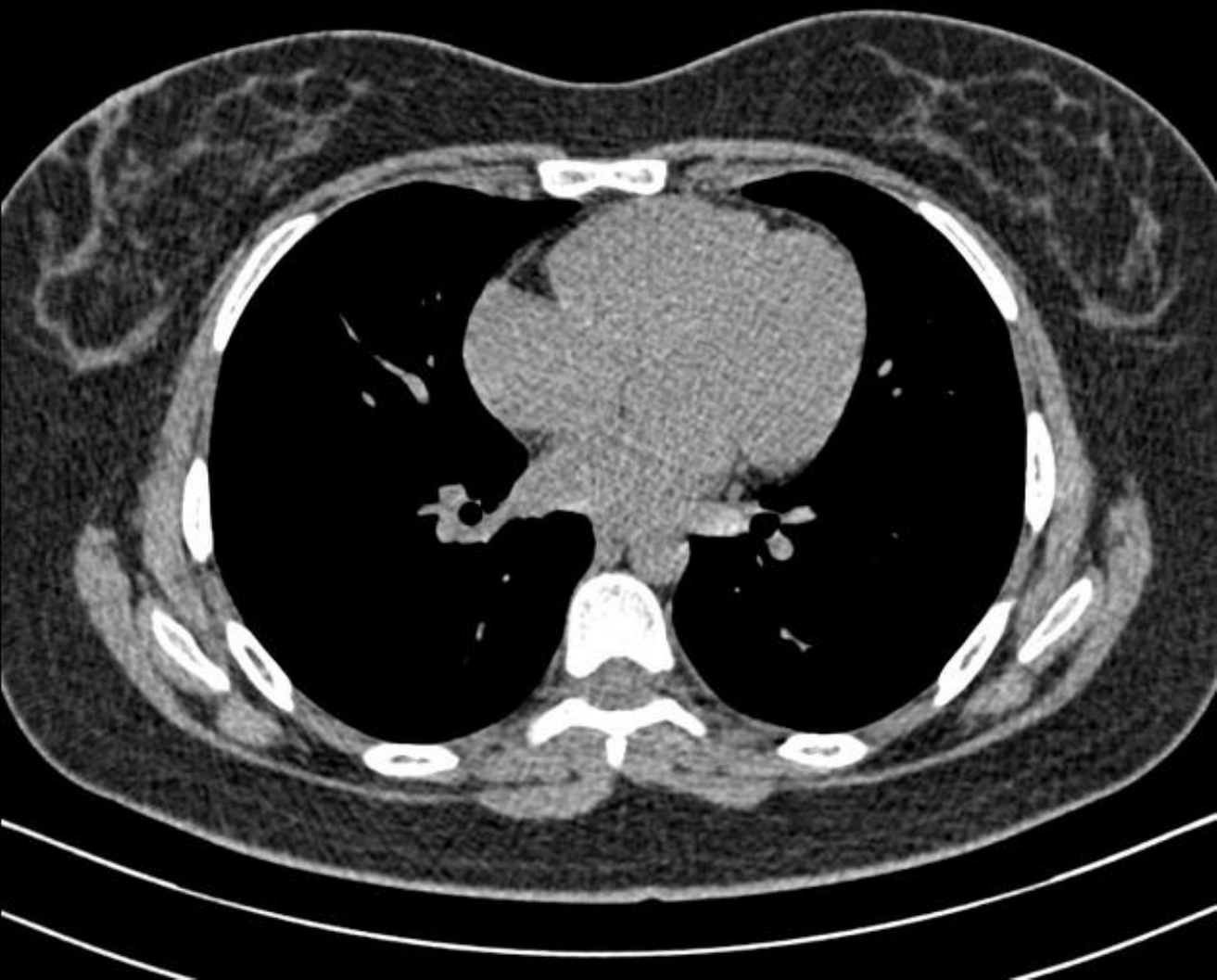}
    }
    \subfigure[Adenocarcinoma]{
        \includegraphics[width=0.205\textwidth]{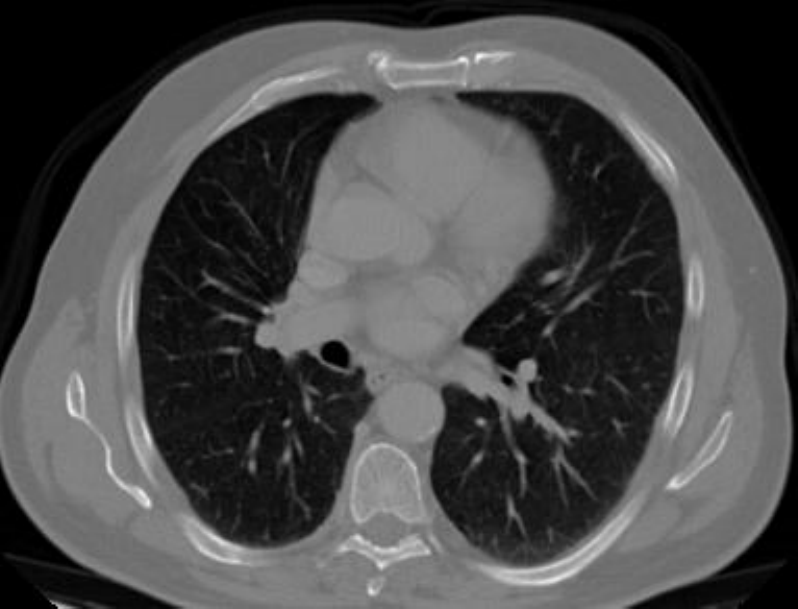}
    }
    \subfigure[Squamous cell carcinoma]{
        \includegraphics[width=0.20\textwidth]{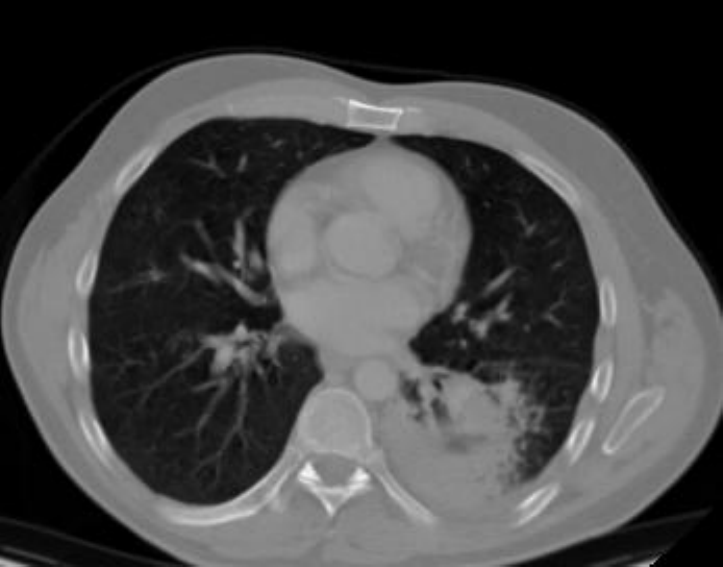}
    }
    \subfigure[Large cell carcinoma]{
        \includegraphics[width=0.23\textwidth]{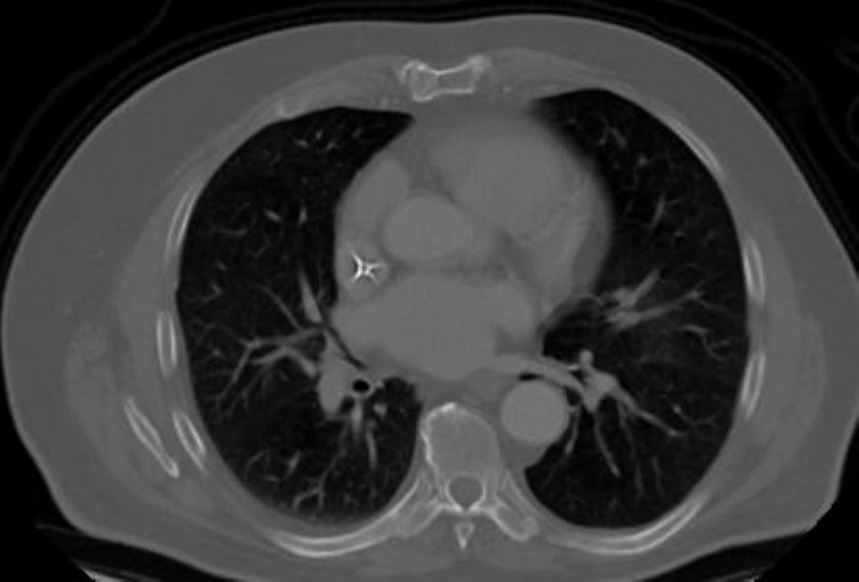}
    }
    \caption{Chest CT scans.}
  \label{fig:chestct}
\end{figure}

Chest CT scans play a relevant role in diagnosing a wide range of thoracic diseases, from infections to malignancies. By analyzing those scans, it is possible to detect diseases in the chest region, including pneumonia and cancer. The selected chest CT dataset \footnote{https://www.kaggle.com/datasets/mohamedhanyyy/chest-ctscan-images} contains 967 images across normal and three cancer types: adenocarcinoma, large-cell carcinoma, and squamous-cell carcinoma. Figure~\ref{fig:chestct} provides a visual representation of the four highlighted classes.

\subsubsection{Brain MRI}

\begin{figure}[h]
    \centering
    \subfigure[No tumor]{
        \includegraphics[width=0.20\textwidth]{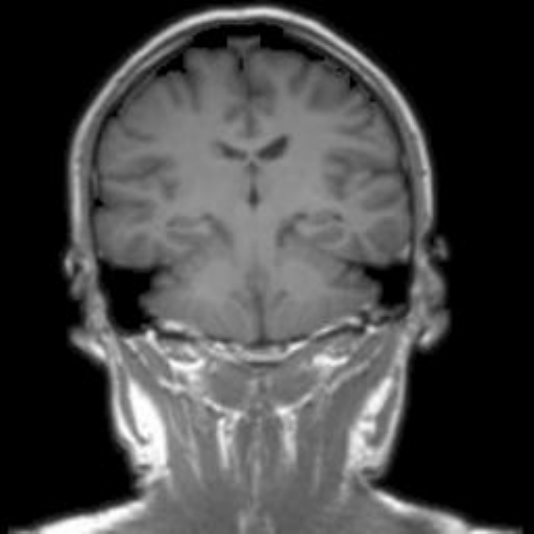}
    }
    \subfigure[Glioma tumor]{
        \includegraphics[width=0.20\textwidth]{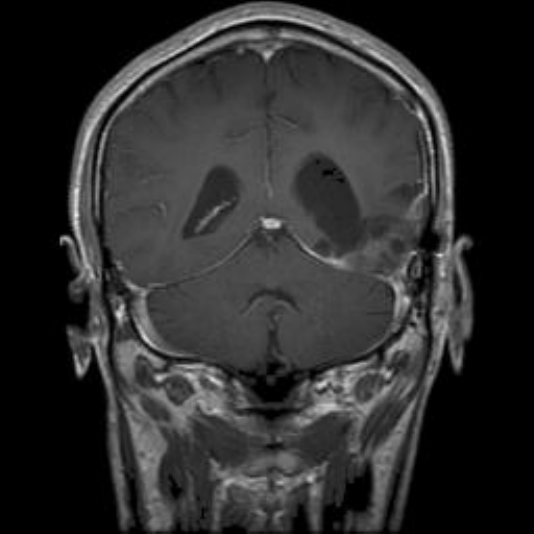}
    }
    \subfigure[Meningioma tumor]{
        \includegraphics[width=0.20\textwidth]{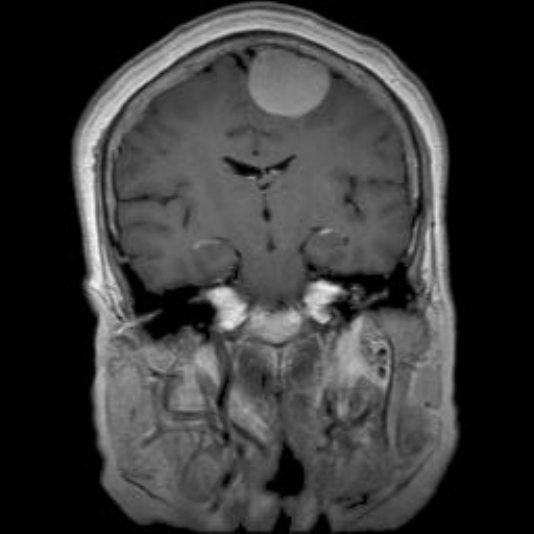}
    }
    \subfigure[Pituitary tumor]{
        \includegraphics[width=0.20\textwidth]{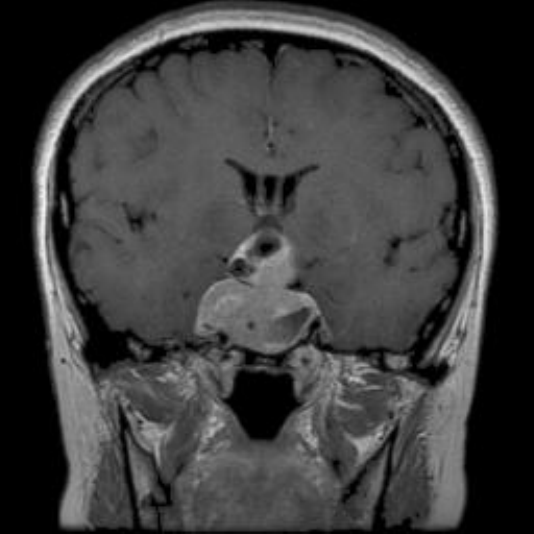}
    }
    \caption{Brain MRI scans.}
  \label{fig:brain_mri}
\end{figure}

Brain tumors pose significant health challenges, often impacting critical functions such as vision, balance, and cognition. Early detection is crucial for improving patient outcomes and quality of life. Through MRI scans, different conditions of the brain can be visually detected. The selected brain tumor dataset \footnote{https://www.kaggle.com/datasets/sami009mr/brain-tumor-dataset} contains 3,362 images across four categories: glioma, meningioma, pituitary tumors, and normal brains. Images illustrating the three aforementioned classes can be found in Fig.~\ref{fig:brain_mri}.

\subsubsection{Gastrointestinal endoscopy}

\begin{figure}[h]
    \centering
    \subfigure[Normal cecum]{
        \includegraphics[width=0.2\textwidth]{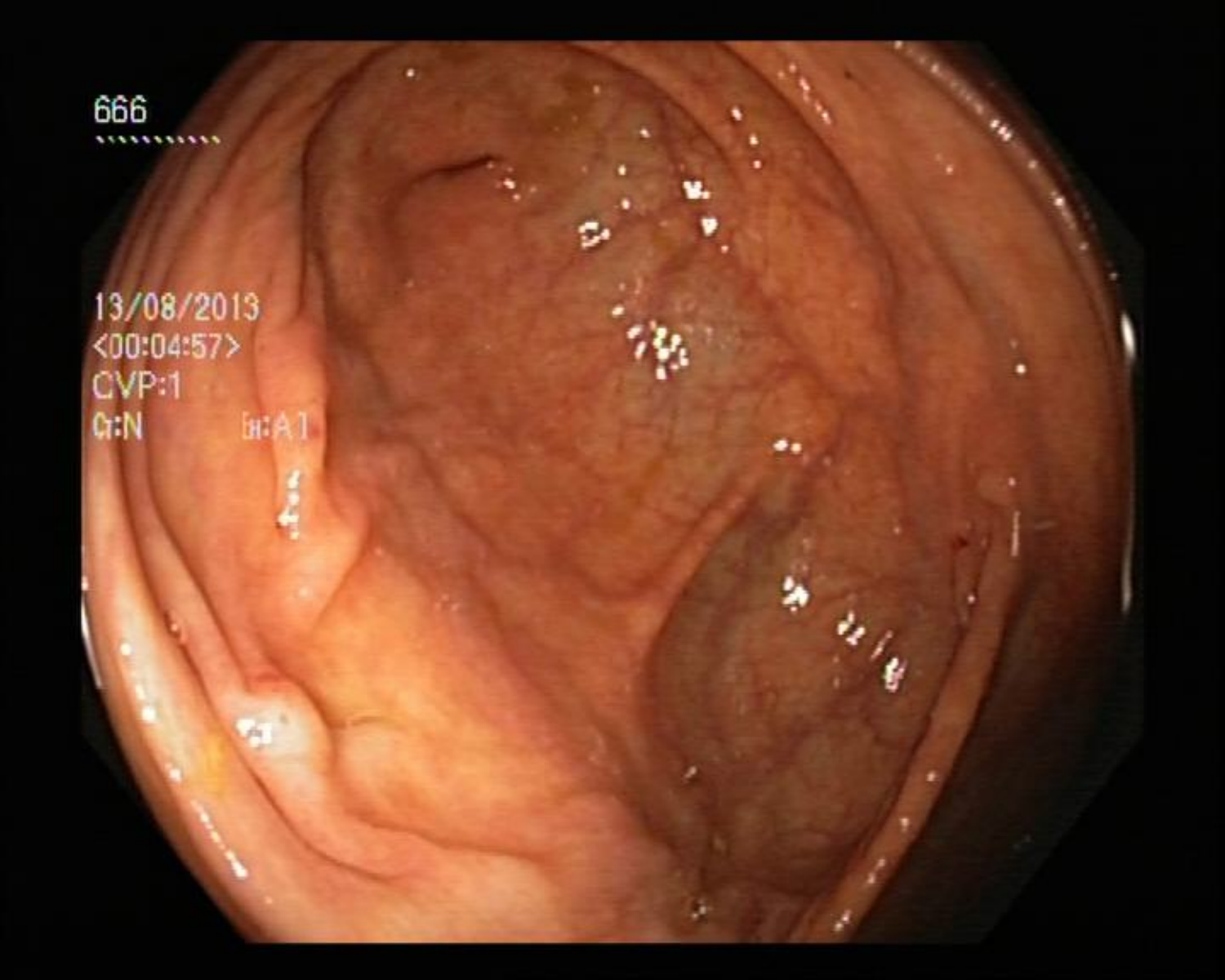}
    }
    \subfigure[Normal pylorus]{
        \includegraphics[width=0.2\textwidth]{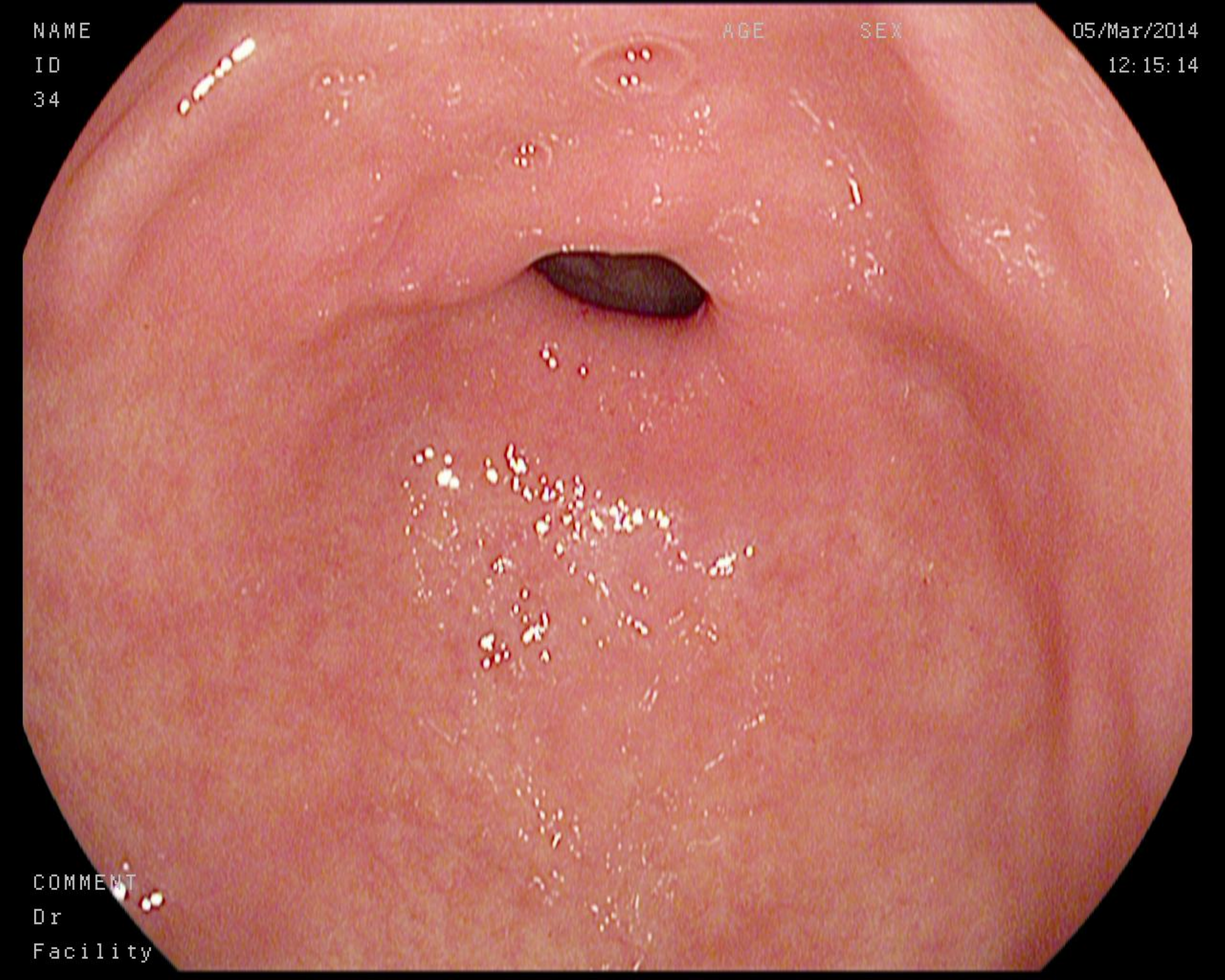}
    }
    \subfigure[Normal z-line]{
        \includegraphics[width=0.2\textwidth]{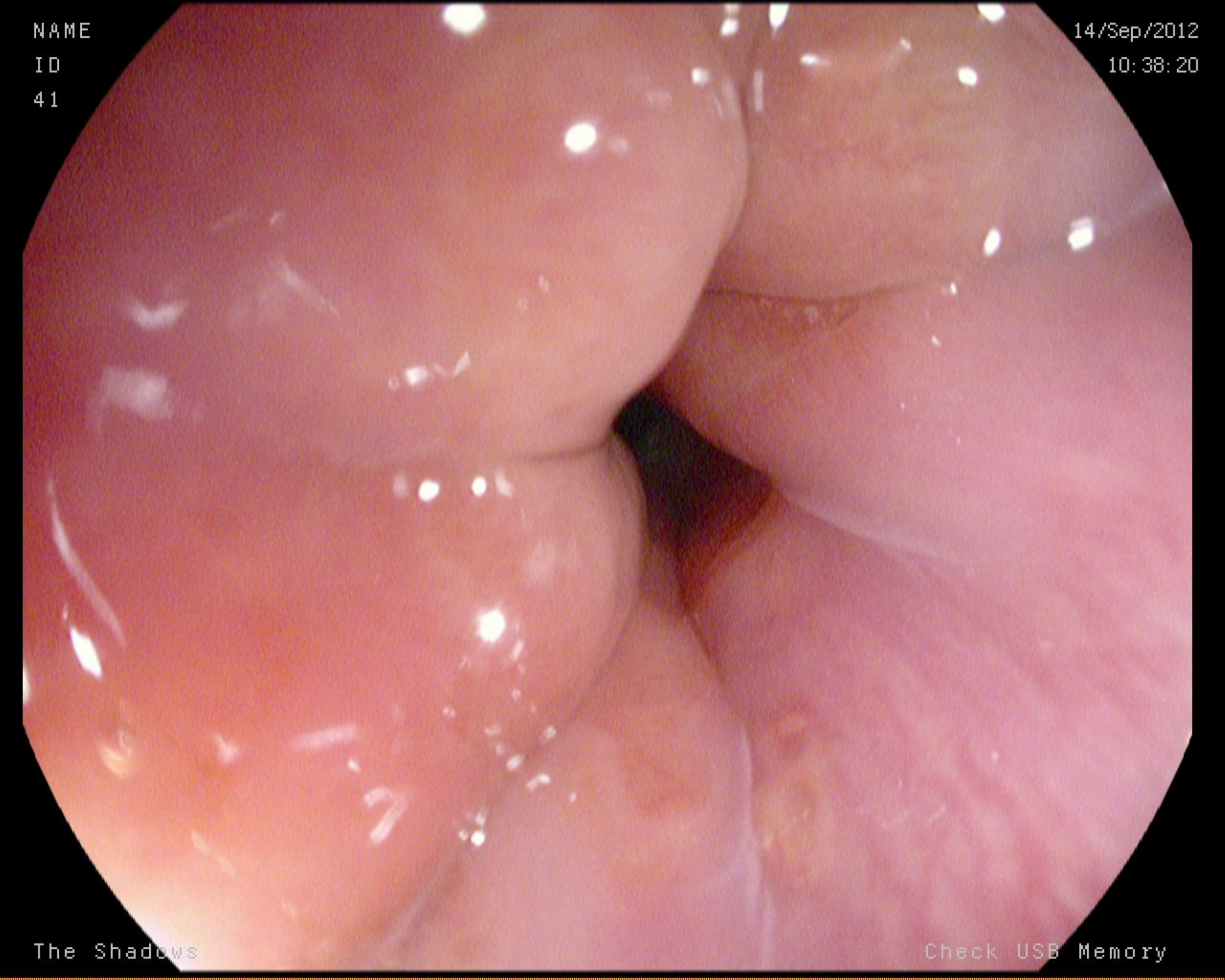}
    }
    \subfigure[Dyed lifted polyps]{
        \includegraphics[width=0.2\textwidth]{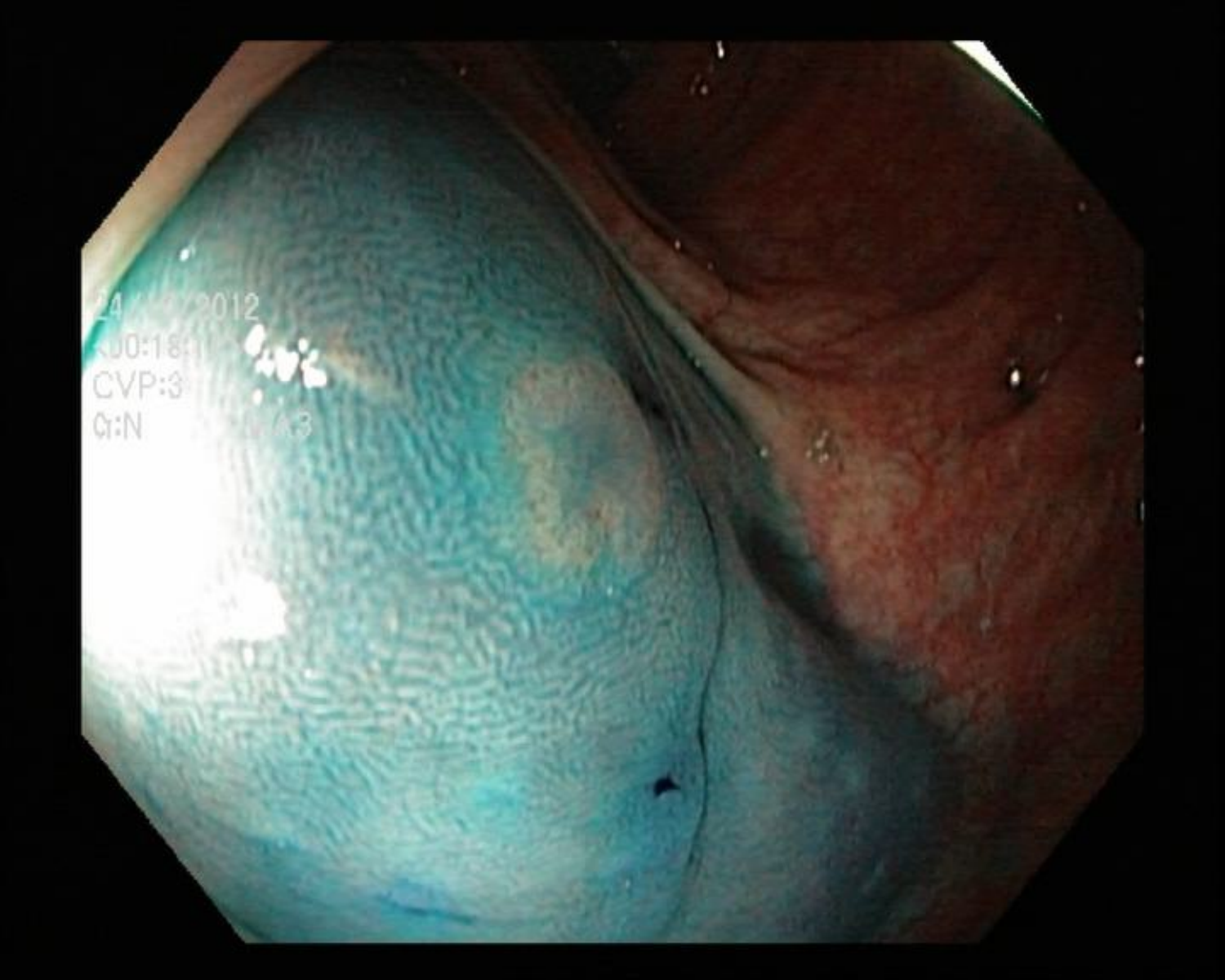}
    }
    \subfigure[Dyed resection margins]{
        \includegraphics[width=0.2\textwidth]{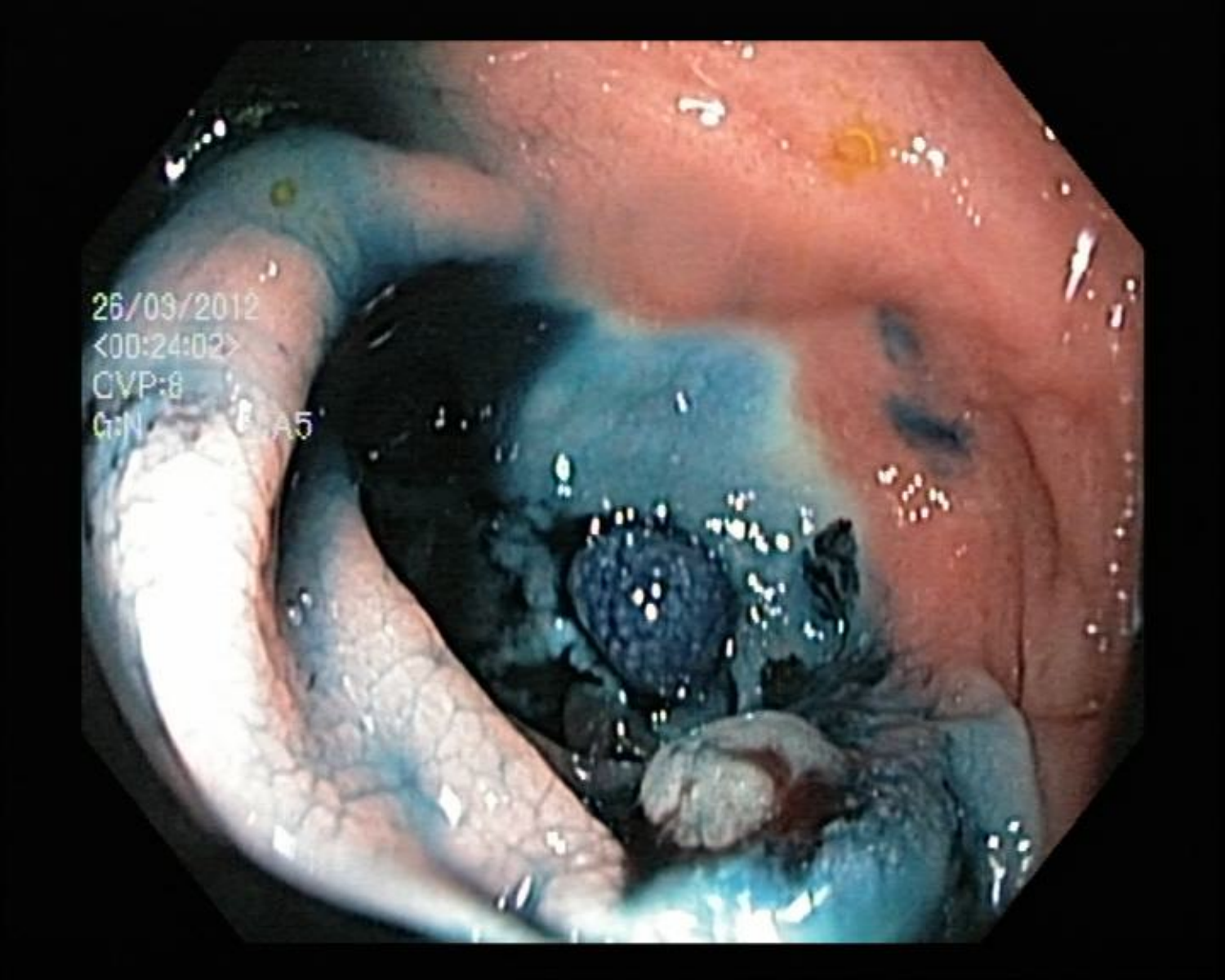}
    }
    \subfigure[Esophagitis]{
        \includegraphics[width=0.2\textwidth]{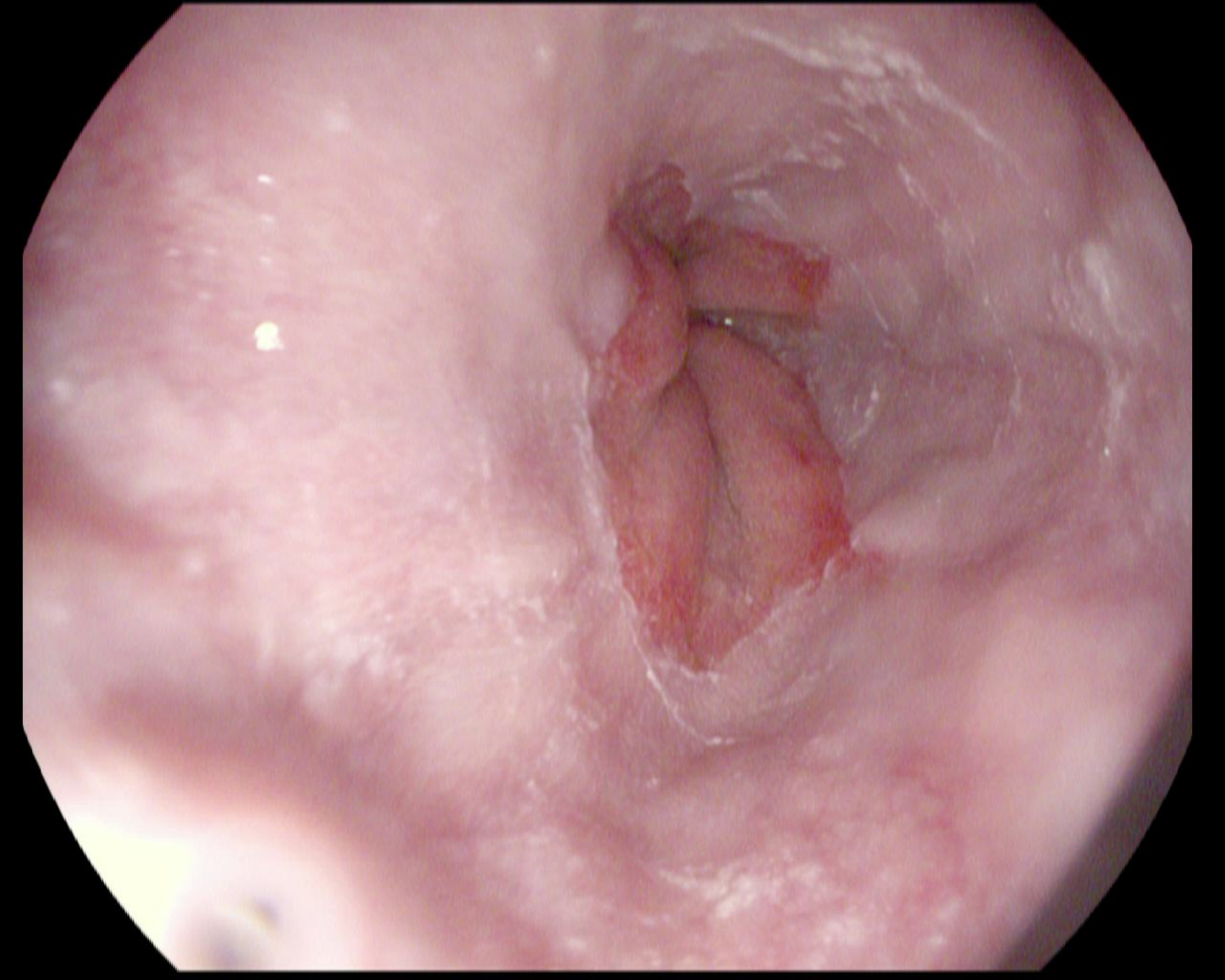}
    }
    \subfigure[Polyps]{
        \includegraphics[width=0.2\textwidth]{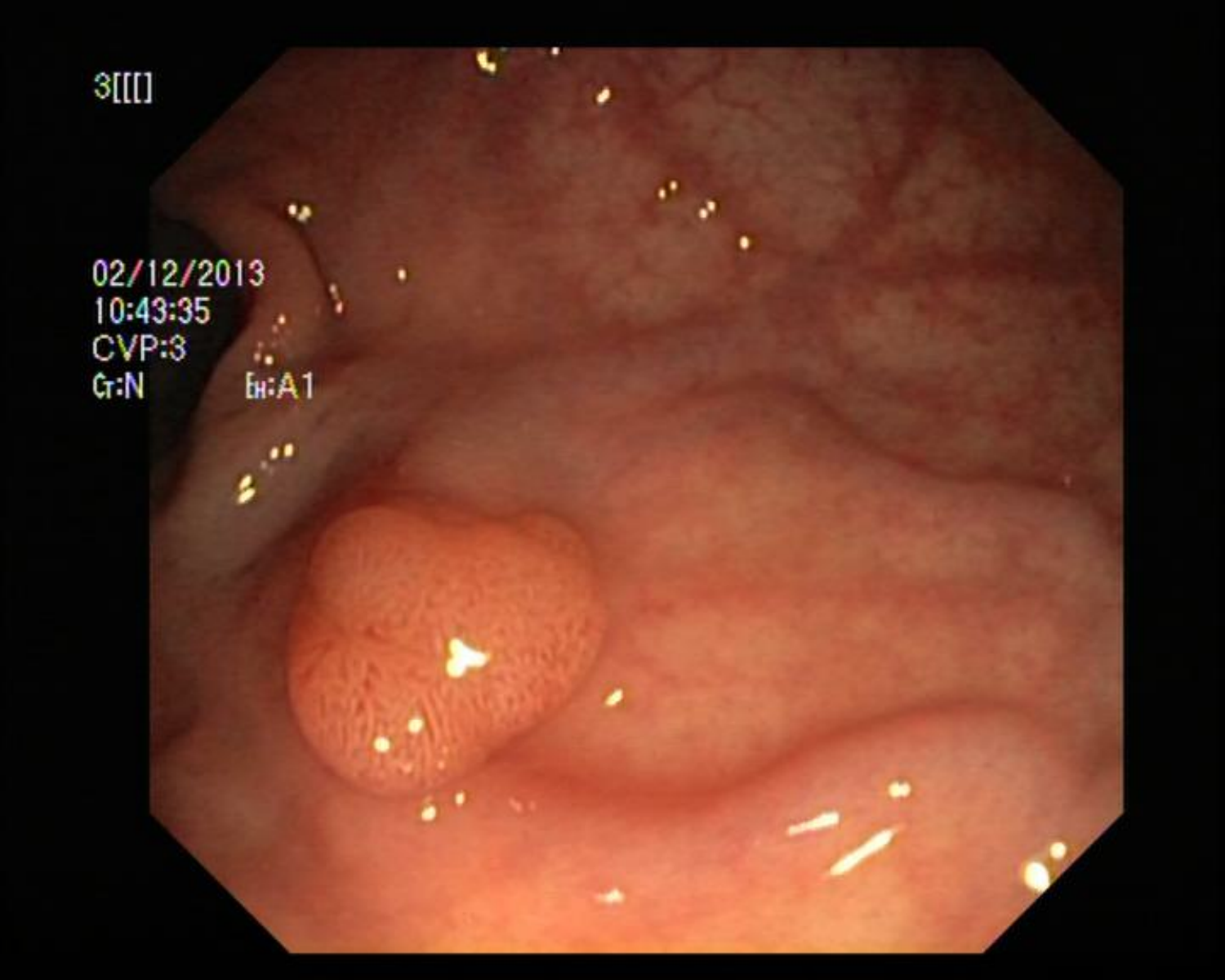}
    }
    \subfigure[Ulcerative-colitis]{
        \includegraphics[width=0.2\textwidth]{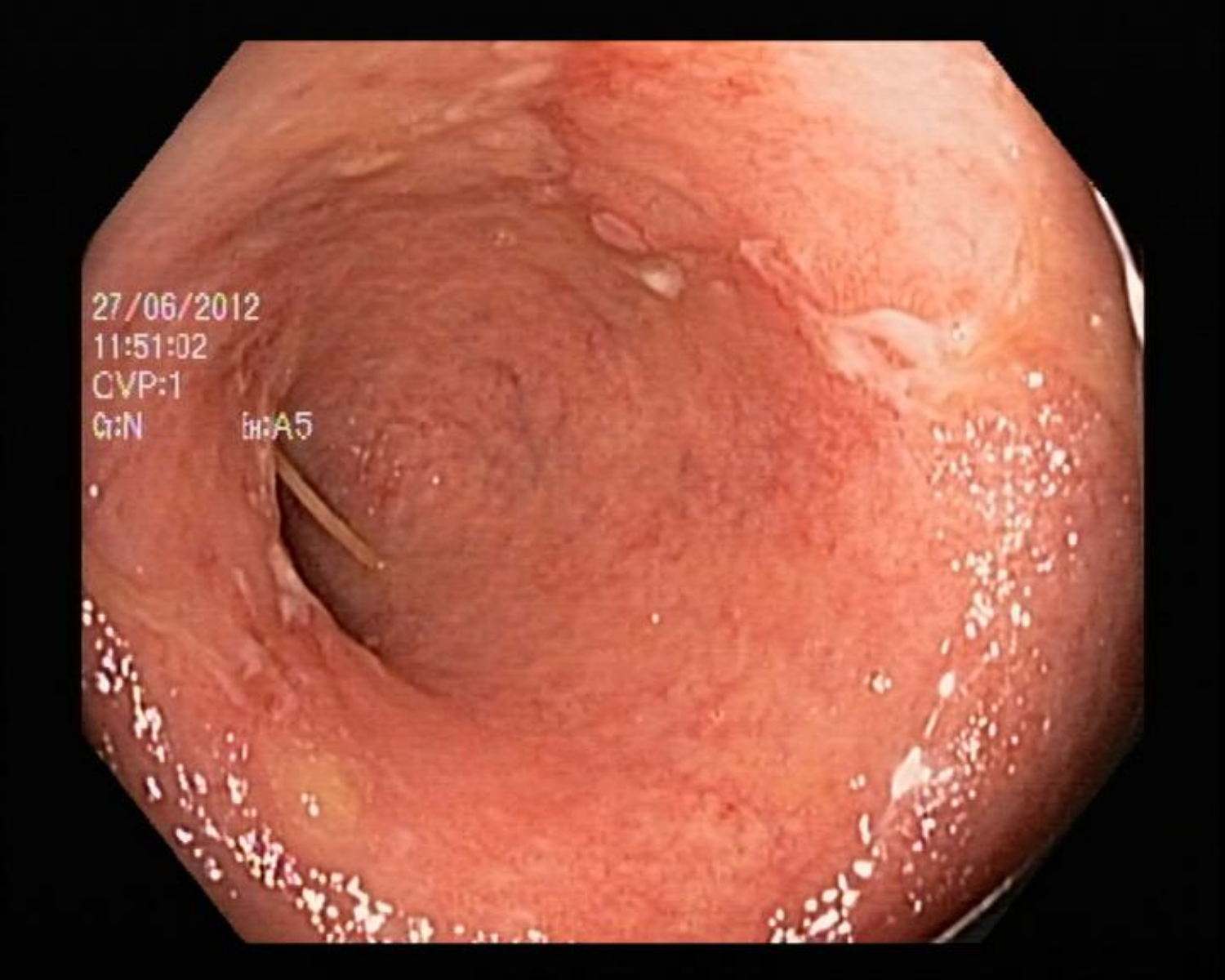}
    }
    \caption{Gastrointestinal endoscopy images.}
  \label{fig:kvasir}
\end{figure}

Endoscopies are used to detect diseases in the human digestive system. Images from inside the gastrointestinal tract can help physicians detect diseases early. The used dataset \footnote{https://www.kaggle.com/datasets/abdallahwagih/kvasir-dataset-for-classification-and-segmentation} contains eight image categories of the digestive system obtained through the endoscopy imaging technique, as shown in Fig.~\ref{fig:kvasir}. There are a total of $6000$ images that can be used in training and testing ML algorithms.

Table \ref{tab:datasets} provides a detailed summary of the datasets used in this study, highlighting the total number of images in each dataset, the distribution between training and testing sets, and the specific classes represented. Each dataset corresponds to a particular medical imaging domain with distinct classes relevant to diagnosing specific conditions. Furthermore, Table \ref{tab:datasets_specifications} complements this information by detailing the technical specifications of each dataset, including image types (e.g., grayscale, RGB), the range of image resolutions, and the file formats in which the data are stored.

\begin{figure}[!hb]
    \centering
    \subfigure[Histological biopsy ]{
        \includegraphics[width=0.168\textwidth]{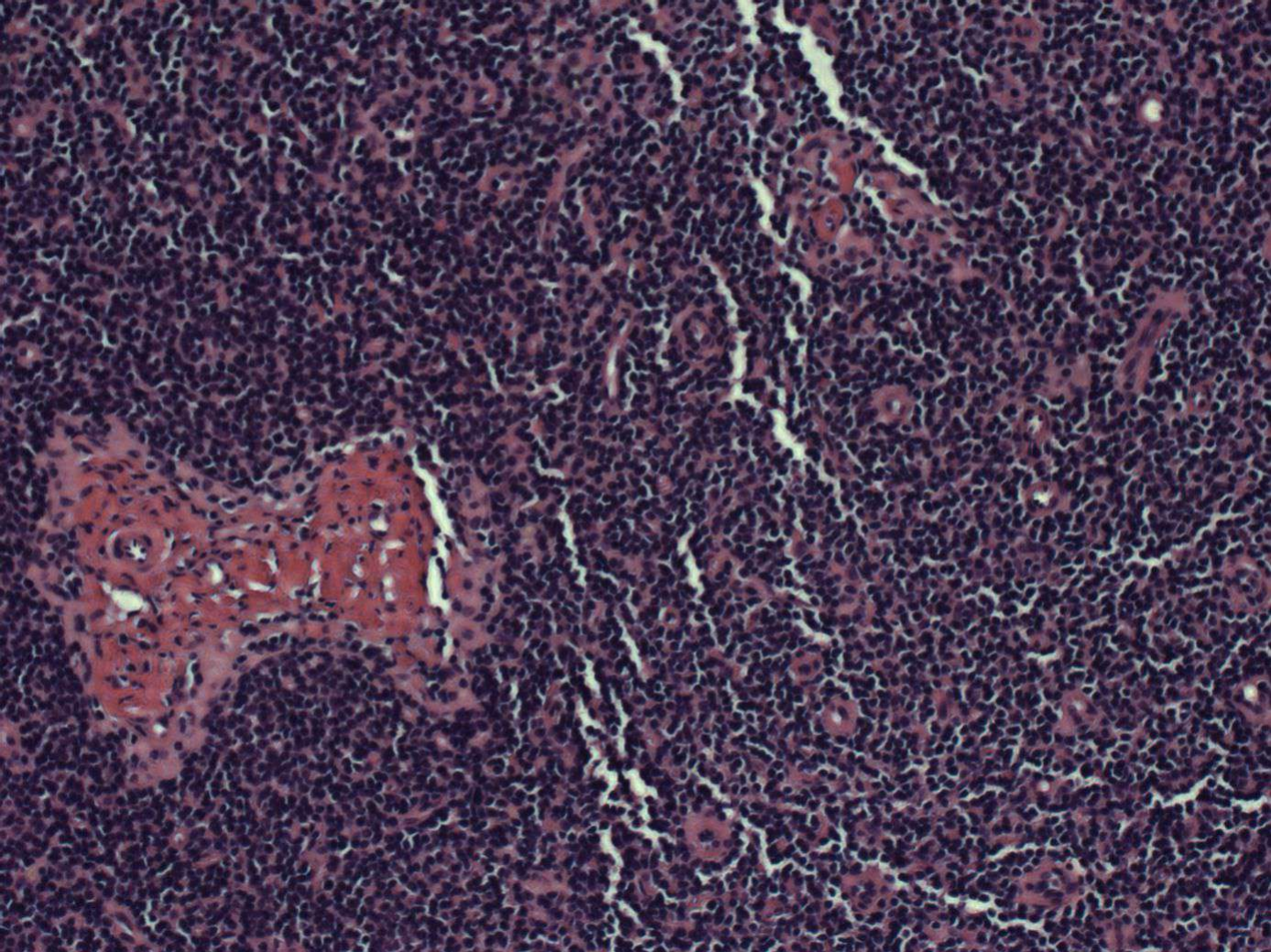}
    }
    \subfigure[Eye fundus]{
        \includegraphics[width=0.189\textwidth]{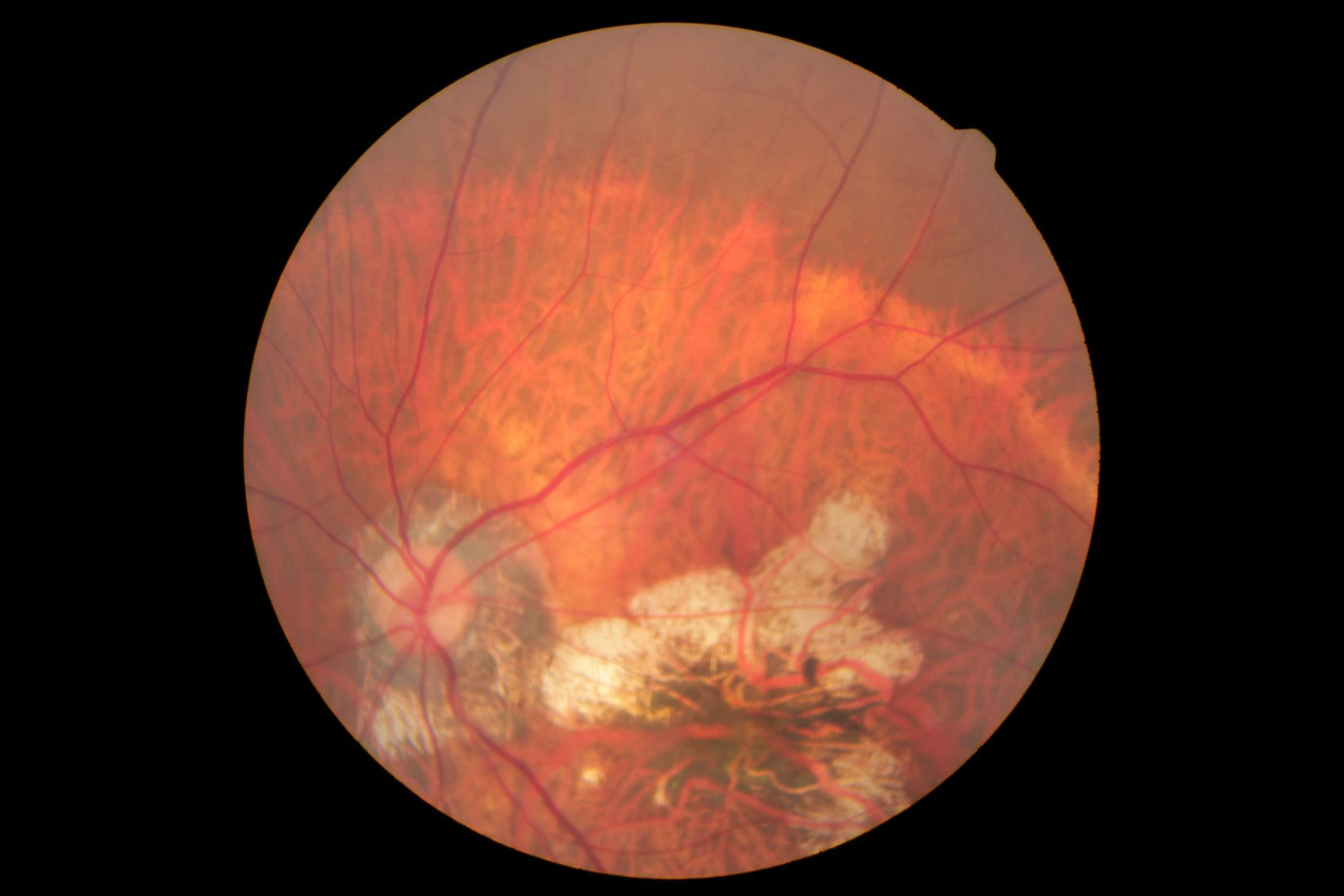}
    }
        \subfigure[Breast ultrasound]{
        \includegraphics[width=0.152\textwidth]{busi.pdf}
    }
    \subfigure[Chest CT]{
        \includegraphics[width=0.157\textwidth]{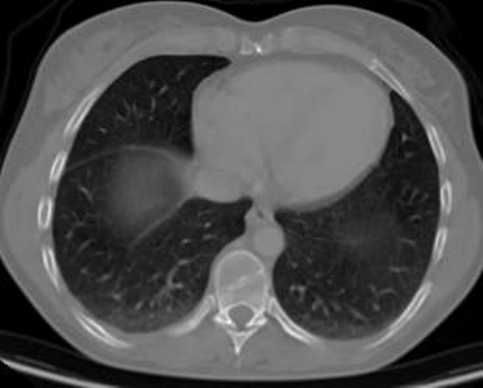}
    }\\
    \subfigure[Brain MRI]{
        \includegraphics[width=0.12\textwidth]{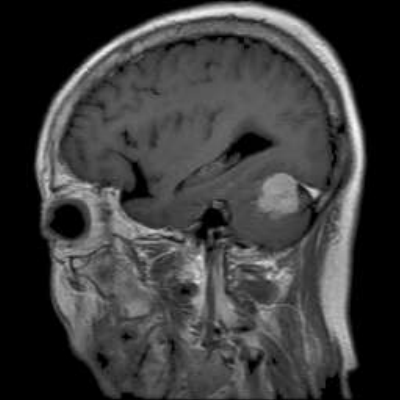}
    }
    \subfigure[Gastrointestinal endoscopy]{
        \includegraphics[width=0.15\textwidth]{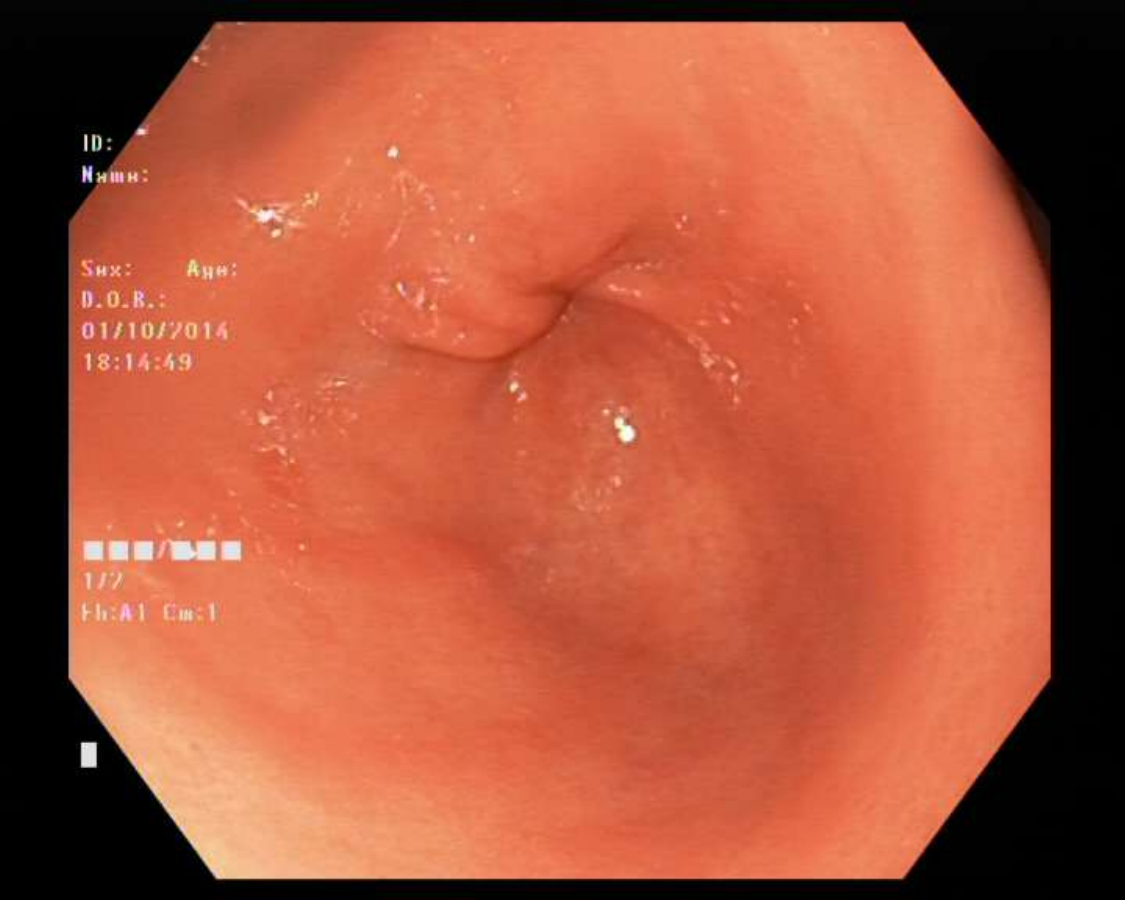}
    }
    \subfigure[Ocular alignment]{
        \includegraphics[width=0.3\textwidth]{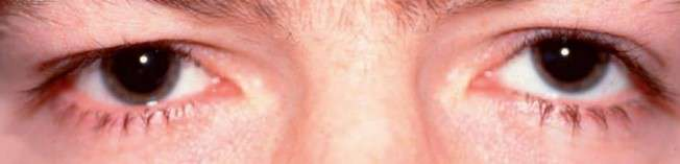}  
    }
    \caption{Example images of each dataset.}
  \label{fig:example_images}
\end{figure}

Additionally, Fig. \ref{fig:example_images} presents a representative image from each dataset, allowing for a visual comparison of the diverse tones, patterns, and textures present across the datasets. Such diversity in visual characteristics highlights the need for a robust model that can handle all image characteristics.

\begin{table}[!ht]
  \centering
  \caption{Selected datasets distribution and classes.}
  \scalebox{0.8}{
    \begin{tabular}{ccccp{15em}}
    \toprule
    \textbf{Dataset} & \textbf{Total images} & \textbf{Train images} & \textbf{Test images} & \multicolumn{1}{c}{\textbf{Classes}} \\
    \midrule
    Histological biopsy & 374   & 298   & 76    & Chronic lymphocytic leukemia\newline{}Follicular lymphoma\newline{}Mantle cell lymphoma \\
    \midrule
    Ocular alignment & 509   & 405   & 104   & Normal\newline{}Esotropia\newline{}Exotropia\newline{}Hypertropia\newline{}Hypotropia \\
    \midrule
    Eye fundus & 601   & 480   & 121   & Normal\newline{}Cataract\newline{}Glaucoma\newline{}Retinal disease \\
    \midrule
    Breast ultrasound  & 780   & 619   & 161   & Normal\newline{}Benign\newline{}Malign \\
    \midrule
    Chest CT & 967   & 613   & 354   & Normal\newline{}Adenocarcinoma\newline{}Squamous cell carcinoma\newline{}Large cell carcinoma \\
    \midrule
    Brain MRI & 3362  & 2688  & 674   & No tumor\newline{}Glioma tumor\newline{}Meningioma tumor\newline{}Pituitary tumor \\
    \midrule
    Gastrointestinal endoscopy & 4000  & 3200  & 800   & Normal cecum\newline{}Normal pylorus\newline{}Normal z-line\newline{}Dyed lifted polyps\newline{}Dyed resection margins\newline{}Esophagitis\newline{}Polyps\newline{}Ulcerative-colitis \\
    \bottomrule
    \end{tabular}%
    }
  \label{tab:datasets}%
\end{table}%

 \begin{table}[!ht]
  \centering
  \caption{Technical specifications of the selected datasets.}
  \scalebox{0.8}{
    \begin{tabular}{cp{3cm}cc}
    \toprule
    \textbf{Dataset} & \textbf{Type(s)} & \textbf{Size(s)} & \textbf{Format(s)}\\
    \midrule
    Histological biopsy & RGB & $1040 \times 1388$ & TIF   \\
    \midrule
    Ocular alignment & RGB & $37 \times 124\,-\,381 \times 1803$ & JPG  \\
    \midrule
    Eye fundus & RGB & $1224 \times 1848\,-\,1728 \times 2592$ & PNG \\
    \midrule
    Breast ultrasound  & Grayscale\newline RGB\newline RGBA & $335 \times 190\,-\,704 \times 948$ & PNG \\
    \midrule
    Chest CT & Grayscale\newline RGB\newline RGBA & $110 \times 172\,-\,874 \times 1200$ &  JPG, PNG \\
    \midrule
    Brain MRI & Grayscale & $167 \times 175\,-\,1446 \times 1375$ & JPG  \\
    \midrule
    Gastrointestinal endoscopy & RGB & $487 \times 332\,-\,1072 \times 1920$  &  JPG  \\
    \bottomrule
    \end{tabular}%
    }
  \label{tab:datasets_specifications}%
\end{table}

\subsection{Classical feature extractors}
A variety of classical methods for feature extraction are employed to benchmark and compare the performance of the MIAFEx framework. Below, a detailed overview of the selected techniques is provided, highlighting their key characteristics and relevance to the comparison.

\begin{itemize}
    \item \textbf{Histogram of Oriented Gradients (HOG)}
\end{itemize}
The Histogram of Oriented Gradients (HOG) \cite{dalal2005histograms} operates by dividing an image into blocks, computing the gradient magnitude and direction at each pixel, and constructing histograms of gradient orientations for these blocks. These histograms are normalized across blocks to achieve robustness against illumination changes and shadowing. The HOG technique is effective for detecting edges and shapes, making it popular in applications like object detection. However, its sensitivity to noise and limited ability to capture global image context reduce its effectiveness in complex scenarios, such as medical imaging, where subtle variations are critical.

\begin{itemize}
    \item \textbf{Scale-Invariant Feature Transform (SIFT)}
\end{itemize}
The Scale-Invariant Feature Transform (SIFT) \cite{el2013comparative} identifies key points in an image that remain consistent across changes in scale, rotation, and illumination. It achieves this by detecting extrema in the Difference of Gaussians (DoG) across multiple scales, assigning gradient-based orientations, and generating descriptors that summarize local image features. These descriptors are robust to transformations and are widely used in texture classification and object recognition. Despite its strength in analyzing local features, SIFT's high computational cost and inability to capture global patterns limit its application in medical imaging tasks that often require understanding broader contextual relationships.

\begin{itemize}
    \item \textbf{Local Binary Pattern (LBP)}    
\end{itemize}
Local Binary Patterns (LBP) \cite{he1990texture} encode local texture information by comparing pixel intensities in a neighborhood and creating a binary pattern for each pixel. The resulting patterns are aggregated into histograms, which serve as texture descriptors. LBP is computationally efficient and rotation-invariant, making it suitable for simple texture-based tasks. However, its sensitivity to noise and variations in illumination, along with its focus on local regions, makes it less robust for medical images that require capturing fine-grained structural features over larger contexts.

\begin{itemize}
    \item \textbf{Haralick Texture Features (HTFs) }    
\end{itemize}
Haralick Texture Features (HTFs) \cite{lofstedt2019gray} leverage the gray-level co-occurrence matrix (GLCM) to compute statistical measures, such as contrast, homogeneity, and entropy, that characterize spatial relationships between pixel intensities. These features are widely used in texture analysis for applications like cancer detection and tissue classification. While HTFs provide valuable insights into textural patterns, their computational demands and reliance on predefined spatial relationships make them less adaptable to diverse datasets and imaging conditions, as encountered in medical contexts.

\begin{itemize}
    \item \textbf{Gabor Filter (GF)}
\end{itemize}
Gabor Filters \cite{lee1999fingerprint} combine spatial and frequency domain analysis to capture textures and edge information. Each Gabor filter is defined by a Gaussian envelope modulating a sinusoidal wave, and a filter bank with multiple scales and orientations is used to extract localized features. Gabor filters excel in tasks like fingerprint recognition and texture segmentation, where their ability to capture directional features is crucial. However, they are computationally intensive, sensitive to parameter tuning, and may produce redundant information in complex or noisy images, limiting their efficiency in dynamic medical imaging scenarios.

\subsection{Tested classifiers}
Once the MIAFEx features were extracted from the datasets, they were classified using ML techniques. For comparison purposes, DL models (CNNs and the ViT) are used to perform the classification task on the same dataset and observe the performance of the MIAFEx features alongside different classifiers. These techniques and models are listed below.

\subsubsection{Machine learning (ML) classifiers}
\begin{itemize}
    \item \textbf{Logistic regression (LR)}
\end{itemize}
Logistic regression \cite{komarek2004logistic} is a statistical method used for binary classification tasks. It models the probability that a given input belongs to a specific class by applying the logistic function to a linear combination of the input features. Despite its simplicity, LR is widely used for its effectiveness, especially in binary classification problems.

\begin{itemize}
    \item \textbf{Random forest (RF)}
\end{itemize}
Random Forest \cite{rigatti2017random} is an ensemble learning method that constructs multiple decision trees during training and merges their predictions to improve accuracy and reduce overfitting. Each tree is trained on a different random subset of the data, and predictions are made based on the majority vote (classification) or average (regression) of all trees.

\begin{itemize}
    \item \textbf{Support vector machine (SVM)}
\end{itemize}
Support Vector Machine (SVM) \cite{chandra2021survey} is a powerful supervised learning algorithm used for classification and regression tasks. SVM aims to find the optimal hyperplane that maximizes the margin between different classes in the feature space. It is particularly effective for high-dimensional data and can handle both linear and non-linear classification using kernel functions.

\begin{itemize}
    \item \textbf{Extreme gradient boosting (XGBoost)}
\end{itemize}
XGBoost \cite{chen2015xgboost} is an efficient and scalable implementation of gradient boosting, an ensemble technique that builds multiple weak learners (typically decision trees) in sequence. Each tree corrects the errors of the previous one, and the final model combines all trees to make more accurate predictions. XGBoost is known for its high performance and speed, particularly in structured data tasks.

\begin{itemize}
    \item \textbf{Wrapper-based feature selection (WFS) and metaheuristics}
\end{itemize}
Wrapper-based feature selection is a technique that evaluates the performance of an ML model using different subsets of features to identify the most relevant ones \cite{kumar2022comprehensive}. It ``wraps'' the feature selection process around the learning algorithm to optimize the selected subset. Metaheuristic algorithms are often used to efficiently explore the search space of feature subsets, especially when dealing with large datasets, to improve the overall model's performance.

\subsubsection{Deep learning (DL) models}

\begin{itemize}
    \item \textbf{MobileNet-V2}
\end{itemize}
MobileNet-V2 \cite{sandler2018mobilenetv2} is a lightweight CNN designed for mobile and edge devices, intending to optimize both performance and efficiency. It uses depthwise separable convolutions to reduce the number of parameters and computations while maintaining accuracy. MobileNet-V2 introduces linear bottleneck layers and inverted residuals, allowing for effective feature extraction with minimal latency, making it suitable for real-time applications.

\begin{itemize}
    \item \textbf{DenseNet-161}
\end{itemize}
DenseNet-161 \cite{huang2017densely} is a deep CNN that enhances feature propagation and reuse through densely connected layers. In this architecture, each layer receives input from all previous layers, which improves gradient flow and reduces the risk of vanishing gradients. DenseNet-161 consists of 161 layers, allowing for more efficient training and higher accuracy, particularly in image classification tasks.

\begin{itemize}
    \item \textbf{ResNet-50}
\end{itemize}
ResNet-50 \cite{he2016deep} is a widely used deep residual network that tackles the vanishing gradient problem through skip connections, allowing gradients to flow through the network more effectively. Comprising 50 layers, ResNet-50 facilitates the training of very deep networks while maintaining high performance. Its architecture also enables the reuse of features, making it effective for various image recognition tasks and a popular choice for transfer learning.

\begin{itemize}
    \item \textbf{Vision Transformer (ViT)}
\end{itemize}
Vision Transformer (ViT) \cite{dosovitskiy2020image} is a revolutionary model that applies the transformer architecture, which was originally designed for natural language processing. Instead of using convolutional layers, ViT divides images into patches and processes them as sequences, leveraging self-attention mechanisms to capture global dependencies and context. This approach has proven to be highly effective on large-scale datasets, demonstrating that ViT usually outperforms traditional CNNs in vision tasks.

\subsection{Performance metrics}

Evaluating classification algorithms is a fundamental part of determining whether the classification problem they solve is being solved. To evaluate the algorithms in the classification task, the Accuracy, Precision, Recall, and F1-score metrics are used.   

Accuracy measures how many times the algorithm gets its predictions right out of the total number of records. Equation~\ref{Eq. Accuracy} is used to determine the accuracy of ML algorithms.

\begin{equation}\label{Eq. Accuracy}
    \text{Accuracy} =  \frac{\text{TP} + \text{TN}}{\text{TP} + \text{TN} + \text{FP} + \text{FN}} 
\end{equation}

\noindent where $\text{TP}$ represents the number of positive cases correctly identified as positive, $\text{TN}$ denotes the number of negative cases correctly identified as negative, $\text{FN}$ indicates the number of positive cases mistakenly classified as negative, and $\text{FP}$ refers to the number of negative cases incorrectly classified as positive.

On the other hand, Precision measures the algorithm's ability to predict true positives against the total number of positive records. If this value is detected as high, the algorithm is sensitive to true positives and will be effective in detecting any disease. Equation~\ref{Eq. Precision} shows the computation of the precision.

\begin{equation}\label{Eq. Precision}
    \text{Precision} = \frac{\text{TP}}{\text{TP} + \text{FN}}
\end{equation}

Equation~\ref{Eq. Recall} is used to obtain the Recall of a classification algorithm. The recall measures the proportion of all true positives that were correctly classified as positives. This metric is also known as the true positive rate.

\begin{equation}\label{Eq. Recall}
    \text{Recall} = \frac{\text{TP}}{\text{TP} + \text{TN}}
\end{equation}

Finally, the F1-score is a metric that allows measuring the quality of a model to predict imbalanced datasets. It is computed using Eq.~\ref{Eq. F1}.

\begin{equation}\label{Eq. F1}
    F1\text{-score} = 2\times\frac{\text{Precision}\times \text{Recall}}{\text{Precision}+\text{Recall}}
\end{equation}

Since this is a multiclass analysis, the weighted average of these metrics is used to provide a single comprehensive value that accounts for the performance across all classes. The weighted average ensures that the contribution of each class to the overall metric is proportional to the number of true instances in that class, hence preventing skewed results due to class imbalances. The weighted average for a metric $M$ across all classes can be calculated using the formula:

\begin{equation}
M_{\text{weighted}} = \frac{\sum_{i=1}^{n} N_i \cdot M_i}{\sum_{i=1}^{n} N_i}
\end{equation}

\noindent where $M_i$ is the metric value for class $i$, $N_i$ is the number of true instances in class, and $n$ is the total number of classes.

\subsection{Experimental setup}

Table \ref{tab:feature_extractors} provides an overview of the parameter configurations for the feature extractors employed in the study, which include HOG, SIFT, LBP, HTFs, and GF. It details key parameters such as the number of orientations, cell sizes, and other specific settings for each feature extraction method. 

Table \ref{tab:classifiers} presents the configuration for the classifiers used, which include SVM, RF, LR, and XGBoost. This table highlights important parameters such as regularization strength, kernel types, and the number of estimators for each classifier. These tables ensure a clear understanding of the settings used for both feature extraction and classification in the experiment.

The experimental configuration of the WFS method was performed by the $k$-nearest neighbors ($k$-NN) algorithm within the following metaheuristic algorithms \cite{lim2013performance}: Particle Swarm Optimization (PSO),  Differential Evolution (DE), and the Genetic Algorithm (GA). Each metaheuristic algorithm was applied with the number of particles set to 30 and a maximum of 200 iterations.

Regarding the MIAFEx and all DL models, the same configuration was applied. The training process ran for 50 epochs using the Nesterov-accelerated Adaptive Moment Estimation (Nadam) optimizer, with a learning rate of $1 \times 10^{-4}$, and the loss was computed using the cross-entropy loss function. To accelerate convergence and improve model performance, transfer learning was employed across all models.

For MIAFEx, ResNet-50, and ViT training, pre-trained weights from the \textit{ImageNet-21k} \cite{ridnik2021imagenet}  dataset (ViT-base-patch16-224) were utilized. This setup helps mitigate discrepancies in pretraining scale and enables a more balanced comparison between ViT-based methods and CNN baselines. In contrast, DenseNet-161 and MobileNetV2 were only available with pre-trained weights on ImageNet-1K\cite{you2018imagenet}, as no publicly released versions trained on ImageNet-21K currently exist. A batch size of 8 was used during training for all models to ensure consistent input size across experiments.


\begin{table}[ht]
  \centering
  \caption{Summary of feature extractor parameter configurations.}
  \scalebox{0.8}{
  \begin{tabular}{cccc}
    \midrule
    \textbf{Feature Extractor} & \textbf{Parameter} & \textbf{Value} & \textbf{Description} \\
    \midrule
    \multirow{4}{*}{HOG} 
    & Orientations      & 9                       & Number of bins in the histogram. \\
    & Pixels per cell   & $8 \times 8$            & Size of each cell in pixels. \\
    & Cells per block   & $2 \times 2$            & Number of cells per block. \\
    & Block normalization & L2-Hys                 & Normalization method for histograms. \\
    \midrule
    \multirow{2}{*}{SIFT} 
    & Keypoint detector & SIFT                    & Algorithm for detecting keypoints. \\
    & Descriptor length & Variable (padded to max)& Length of the descriptor vector. \\
    \midrule
    \multirow{3}{*}{LBP} 
    & Radius            & 1                       & Radius of the neighborhood. \\
    & Points            & $8 \times$ radius       & Number of sample points. \\
    & Method            & Uniform                 & Uniform patterns for histogram computation. \\
    \midrule
    \multirow{2}{*}{HTFs} 
    & Distances         & 1                       & Pixel distances for the GLCM computation. \\
    & Angles            & \{0, $\pi/4$, $\pi/2$, $3\pi/4$\} & Directions for computing GLCM. \\
    \midrule
    \multirow{4}{*}{Gabor Filter} 
    & Kernel size       & $21 \times 21$          & Size of the Gabor kernel. \\
    & Scales ($\sigma$) & \{1, 3\}                & Determines the width of the Gabor filter. \\
    & Orientations ($\theta$) & 4 (0 to $\pi$)      & Number of orientations to apply. \\
    & Wavelength ($\lambda$) & $\pi/4, \pi/2$    & Wavelengths of the sinusoidal factor. \\
    \bottomrule
  \end{tabular}
  }
  \label{tab:feature_extractors}
\end{table}

\begin{table}[ht]
  \centering
  \caption{Summary of classifier parameter configurations.}
  \scalebox{0.8}{
  \begin{tabular}{cccc}
    \midrule
    \textbf{Classifier} & \textbf{Parameter} & \textbf{Value} & \textbf{Description} \\
    \midrule
    \multirow{3}{*}{SVM} 
    & C                   & 1               & Regularization parameter. \\
    & kernel              & Radial basis function             & Specifies the kernel type. \\
    & degree              & 3                 & Degree of the polynomial kernel function. \\
    \midrule
    \multirow{4}{*}{RF} 
    & Estimators          & 100               & The number of trees in the forest. \\
    & Criterion           & Gini              & The function to measure the quality of a split. \\
    & Random state        & 42                & Seed used by the random number generator. \\
    \midrule
    \multirow{4}{*}{LR} 
    & C                   & 1.0               & Inverse regularization strength. \\
    & Solver              & LBFGS             & Algorithm to use in optimization. \\
    & Max iterations      & 1000              & Maximum number of iterations for solver. \\
    & Random state        & 42                & Seed used by the random number generator. \\
    \midrule
    \multirow{4}{*}{XGBoost} 
    & Estimators          & 100               & Number of boosting rounds. \\
    & Learning rate       & 0.1               & Step size at each iteration. \\
    & Max depth           & 6                 & Maximum depth of the tree. \\
    & Random state        & 42                & Seed used by the random number generator. \\
    \bottomrule
  \end{tabular}
  }
  \label{tab:classifiers}
\end{table}

\section{Results}
\label{sec:results}
This section presents the experimental results across the different medical imaging datasets described above. First, MIAFEx-extracted features are compared with classical feature extractors across a range of ML classifiers to evaluate relative performance. Next, the study assesses MIAFEx’s effectiveness in comparison to DL models, examining its behavior across diverse datasets to determine its robustness. Finally, an analysis of how classification performance varies with dataset size is provided, offering insights into the scalability and adaptability of each model.

\subsection{Classical feature extractors and MIAFEx performance with ML classifiers}
\begin{figure}[htbp]
    \centering
    \subfigure[Histological biopsy ]{
        \includegraphics[width=0.28 \textwidth]{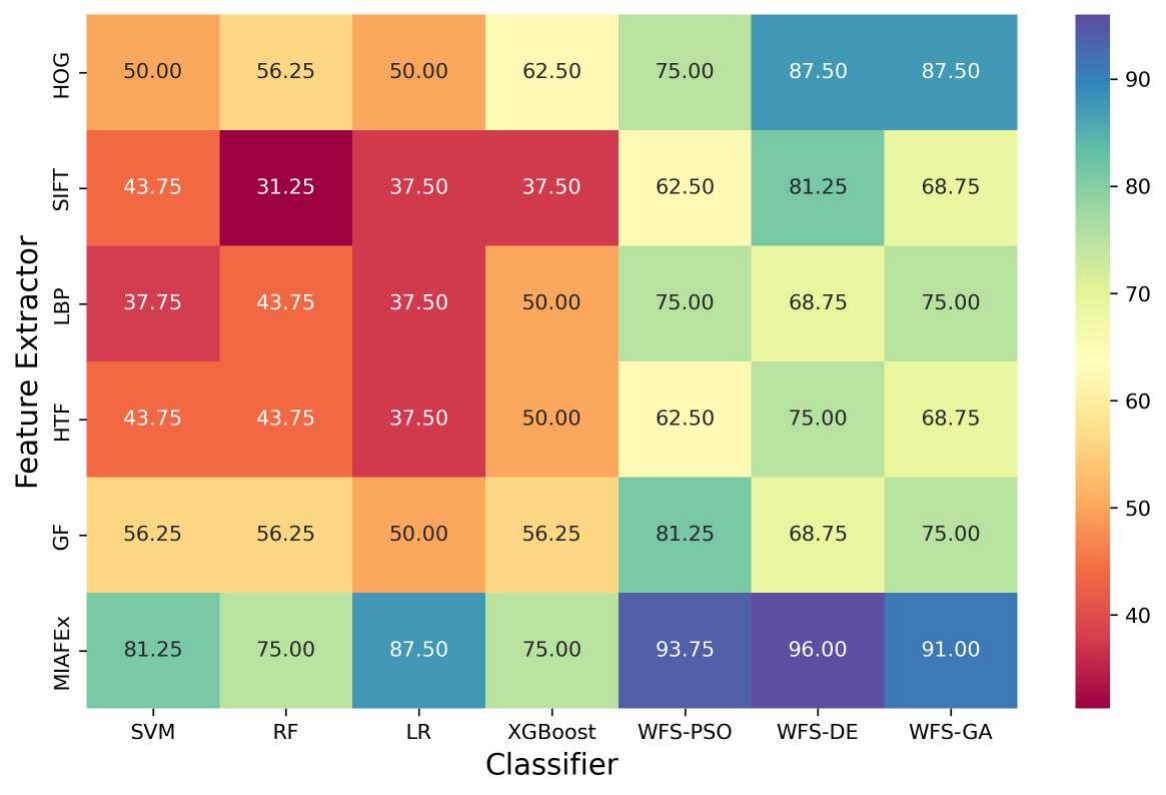}
    }
    \subfigure[Ocular alignment]{
        \includegraphics[width=0.28\textwidth]{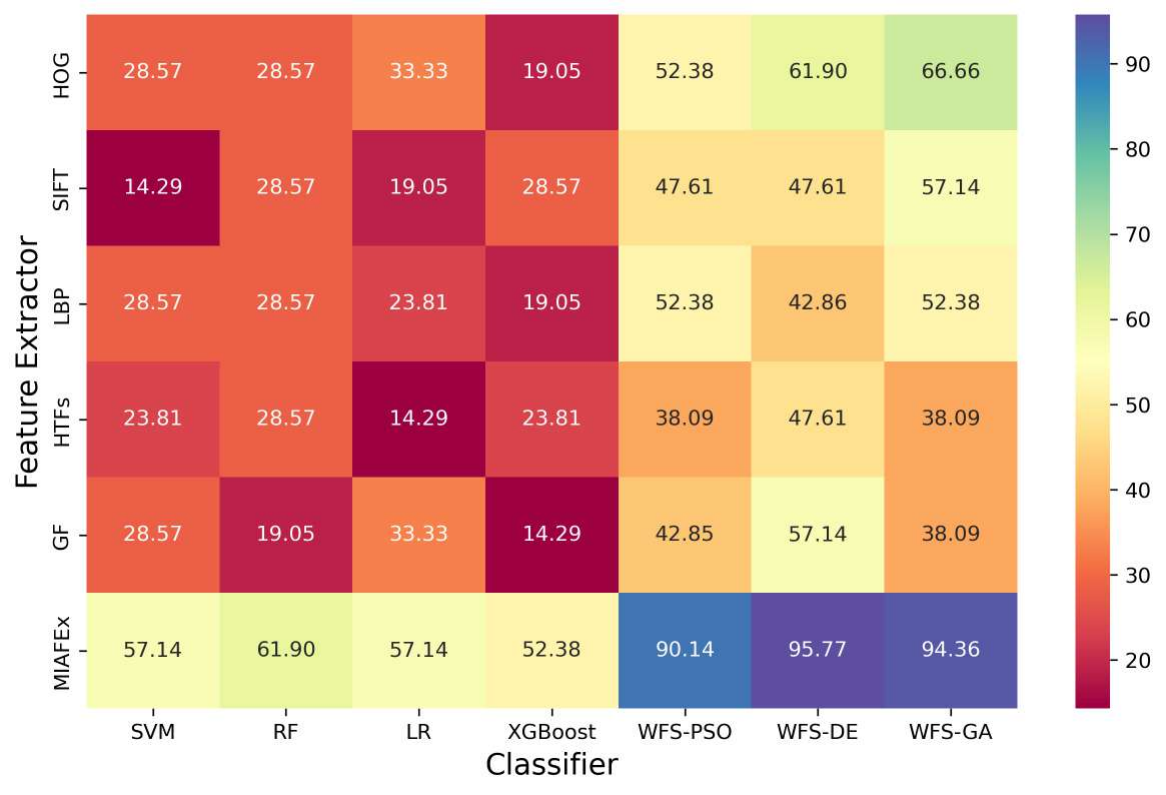}
    }
    \subfigure[Eye fundus]{
        \includegraphics[width=0.28\textwidth]{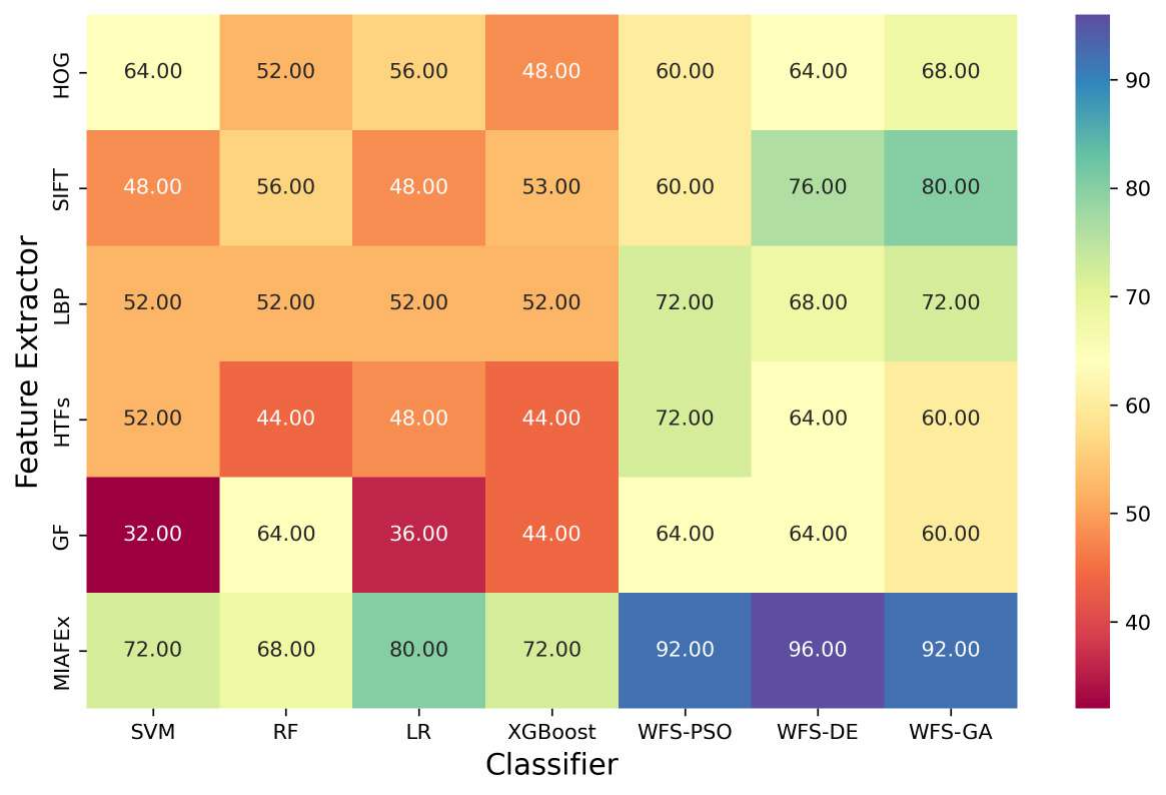}
    }\\
        \subfigure[Breast ultrasound ]{
        \includegraphics[width=0.28\textwidth]{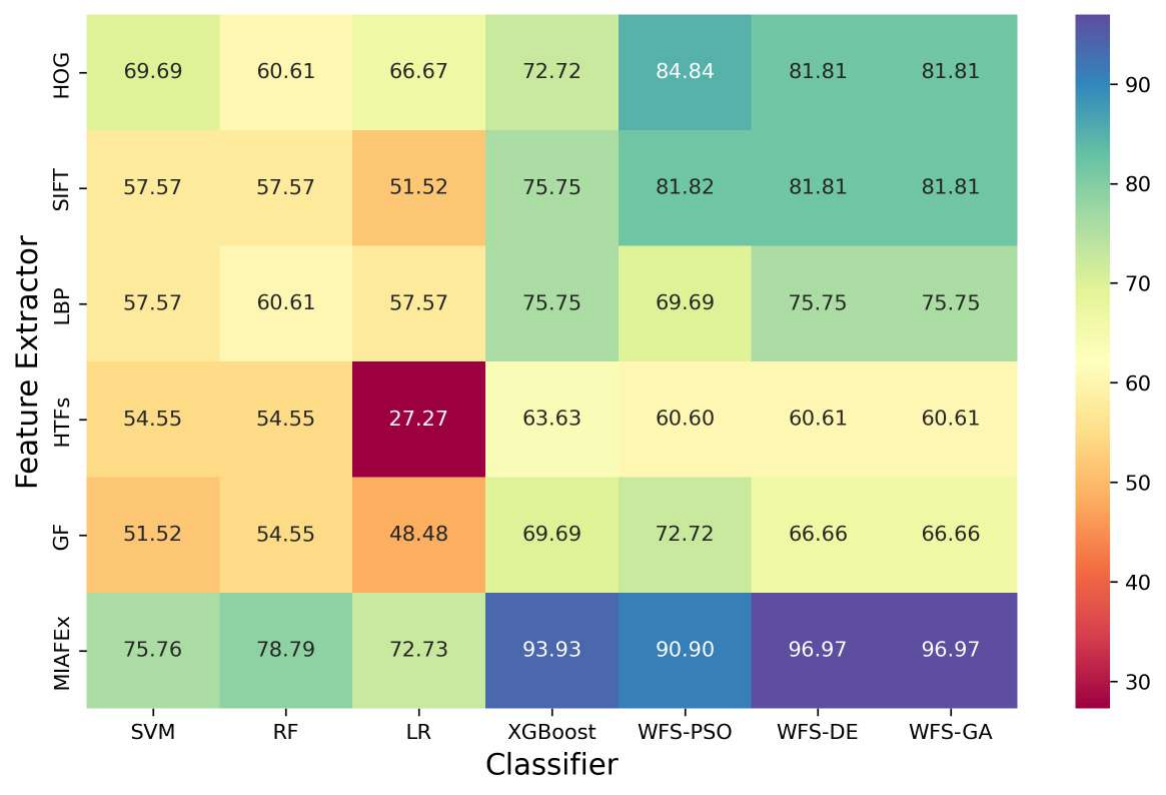}
    }
    \subfigure[Chest CT]{
        \includegraphics[width=0.28\textwidth]{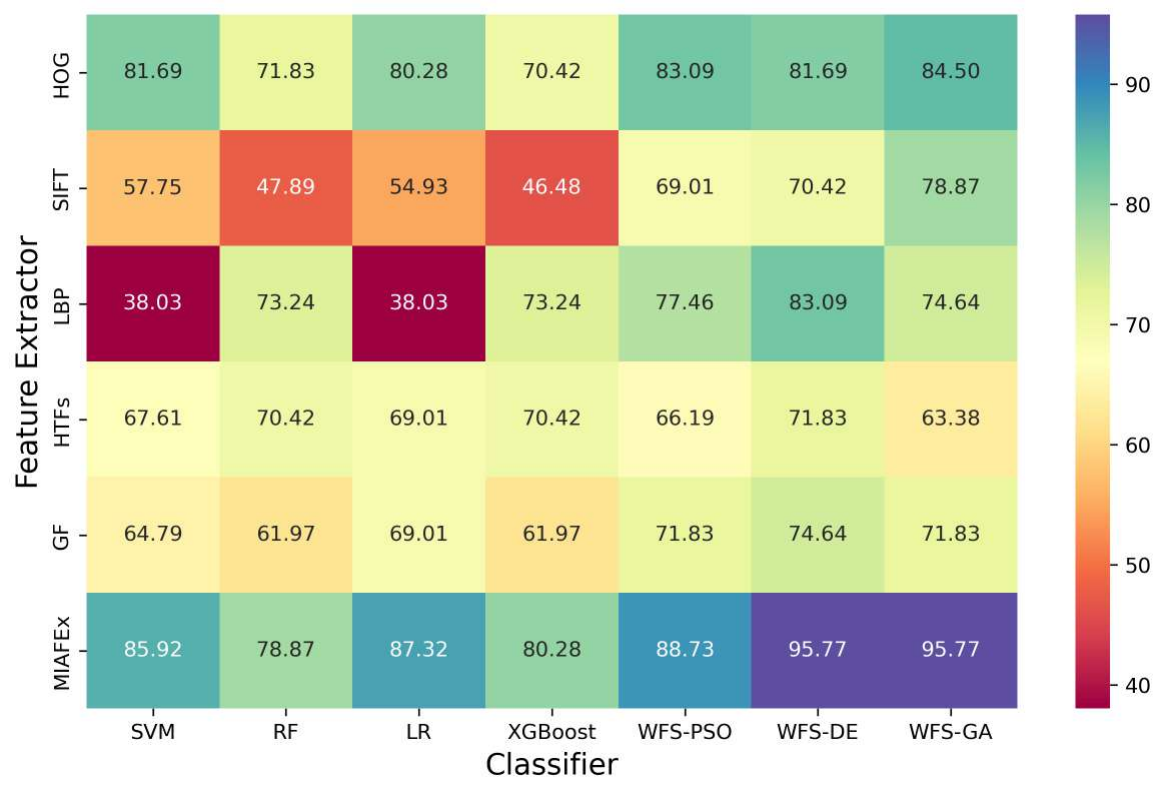}
    }
    \subfigure[Brain MRI]{
        \includegraphics[width=0.28\textwidth]{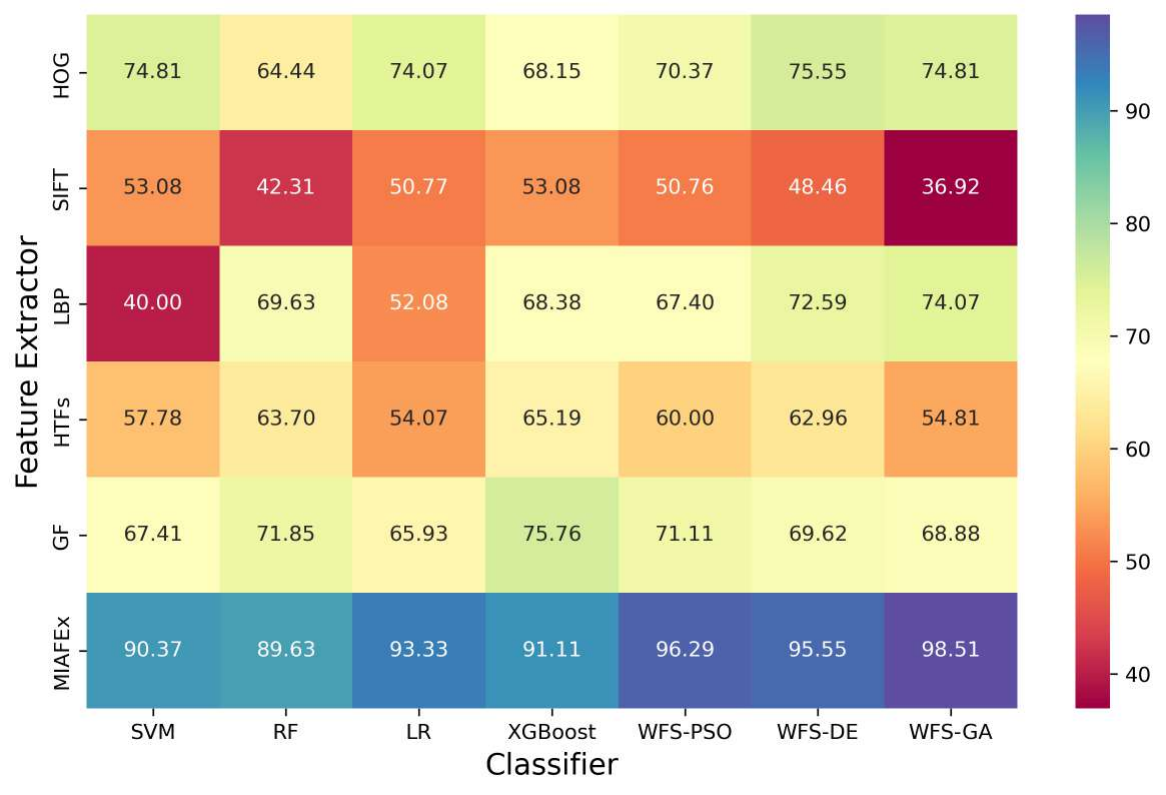}
    }
    \subfigure[Gastrointestinal endoscopy]{
        \includegraphics[width=0.28\textwidth]{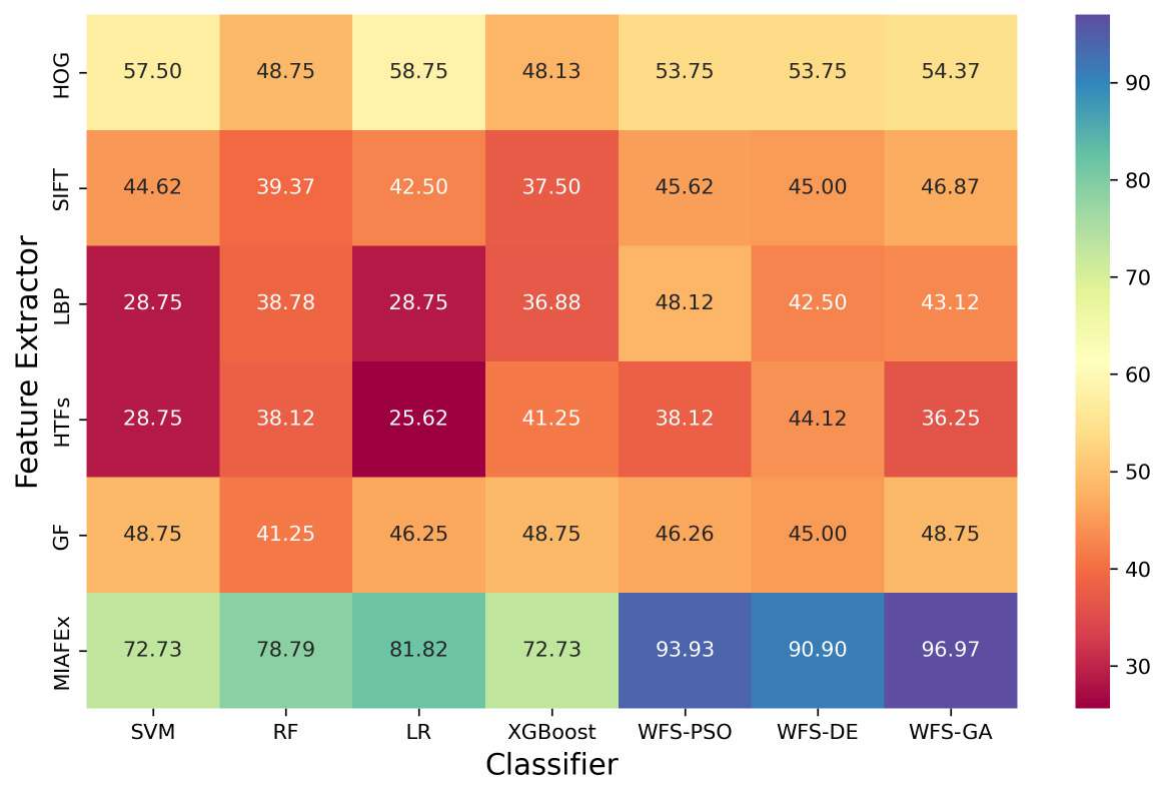}
    }
    \caption{Accuracy performance on different feature extractors and classifiers over all datasets.}
  \label{fig:heatmaps}
\end{figure}

The performance of the MIAFEx was evaluated in conjunction with various classifiers, and its efficacy in feature extraction was compared. The feature extraction techniques considered for comparison were HOG, SIFT, LBP, HTFs, and GF, and the classifiers tested include SVM, RF, LR, XGBoost, and evolutionary optimization techniques (WFS-PSO, WFS-DE, and WFS-GA). In Fig.~\ref{fig:heatmaps}, the accuracy heatmaps of all the combinations of feature extractor, classifier, and dataset are presented. Due to its extensive length, complete tables presenting the analysis of performance across all metrics are included in \ref{app1} for clarity and to maintain the flow of the main document.

The performance of MIAFEx as a feature extractor demonstrates a clear advantage over traditional methods across the different medical imaging datasets. In the Gastrointestinal Endoscopy dataset, the WFS-PSO method achieved 93.93\% accuracy, significantly outperforming other techniques, including MIAFEx with SVM (72.73\%) and traditional methods like HOG (57.50\%) or LBP (28.75\%). Similarly, WFS-DE and WFS-GA also provided strong results, reaching 90.90\% and 96.97\% accuracy, respectively. This trend is evident across other datasets as well. In Brain MRI, WFS-GA with MIAFEx reached 98.51\%, surpassing traditional methods and the MIAFEx-SVM combination (90.37\%). The Chest CT dataset followed a similar pattern, where WFS-DE reached 95.77\%, WFS-GA achieved 95.77\%, and MIAFEx with SVM yielded 85.92\%, highlighting the effectiveness of WFS methods and MIAFEx in boosting classification performance. In more challenging datasets like Strabismus, WFS-DE and WFS-GA achieved accuracies of 95.77\% and 94.36\%, respectively, while traditional methods like HOG and LBP performed at 28.57\% and 28.57\%. The Lymphoma dataset further emphasizes the strength of WFS methods, with WFS-DE and WFS-GA yielding accuracies of 96.00\% and 91.00\%, respectively, compared to MIAFEx with SVM (81.25\%). 

The dependence of classical descriptors on handcrafted heuristics, along with their susceptibility to noise and limited adaptability, significantly reduced their effectiveness in all datasets. The results of this study demonstrate the superiority of MIAFEx as a feature extractor for various medical imaging tasks, and its combination with ML classifiers and optimization techniques presents a promising approach for improving diagnostic accuracy in medical image analysis.  While the absolute performance varies across domains, the trend remains consistent, remarking the MIAFEx as a superior feature extraction approach. Moreover, the use of optimization techniques (WFS-PSO, WFS-DE, WFS-GA) further improved the overall performance.

\subsection{MIAFEx and DL models performance comparison}

\begin{figure}[!htb]
        \centering
        \subfigure[Histological biopsy ]{
            \includegraphics[width=0.31\textwidth]{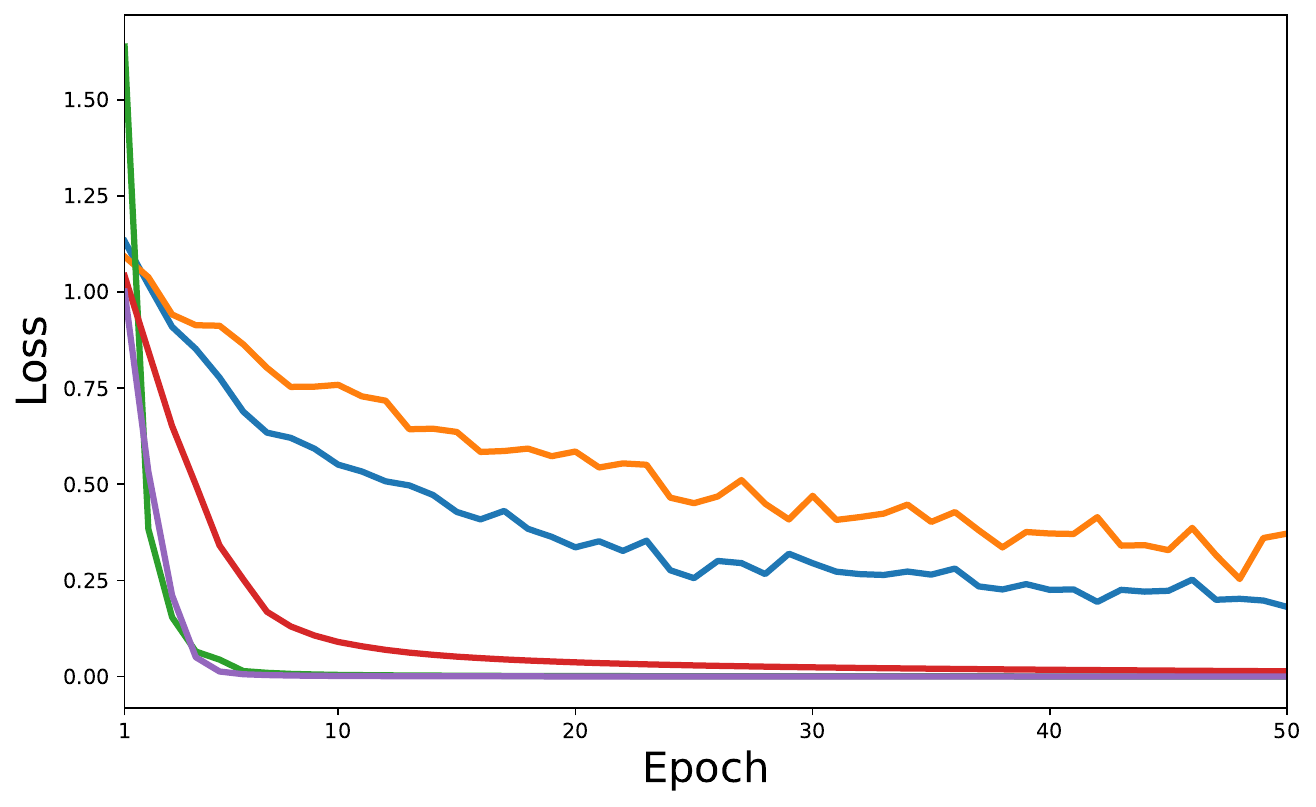}
        }
        \subfigure[Ocular alignment]{
            \includegraphics[width=0.31\textwidth]{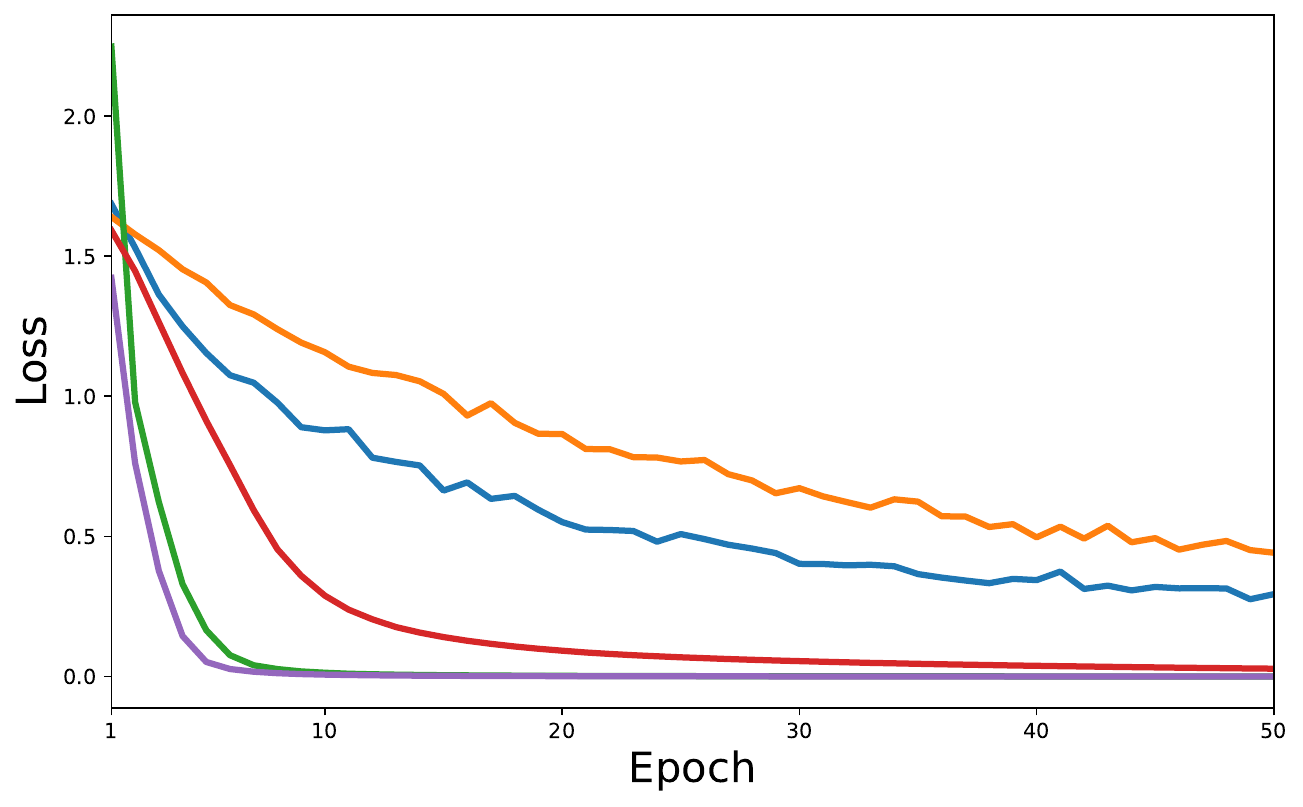}
        }
        \subfigure[Eye fundus]{
            \includegraphics[width=0.31\textwidth]{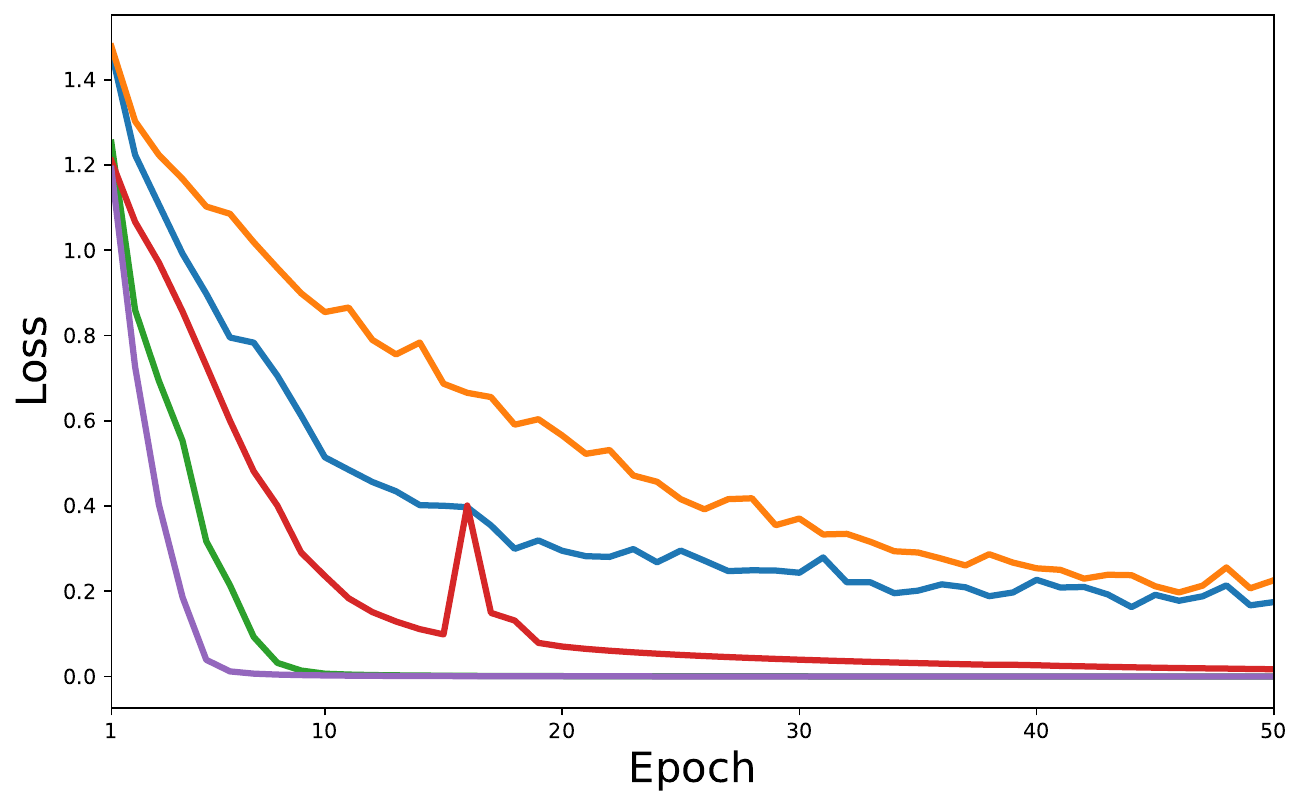}
        }\\
            \subfigure[Breast ultrasound ]{
            \includegraphics[width=0.31\textwidth]{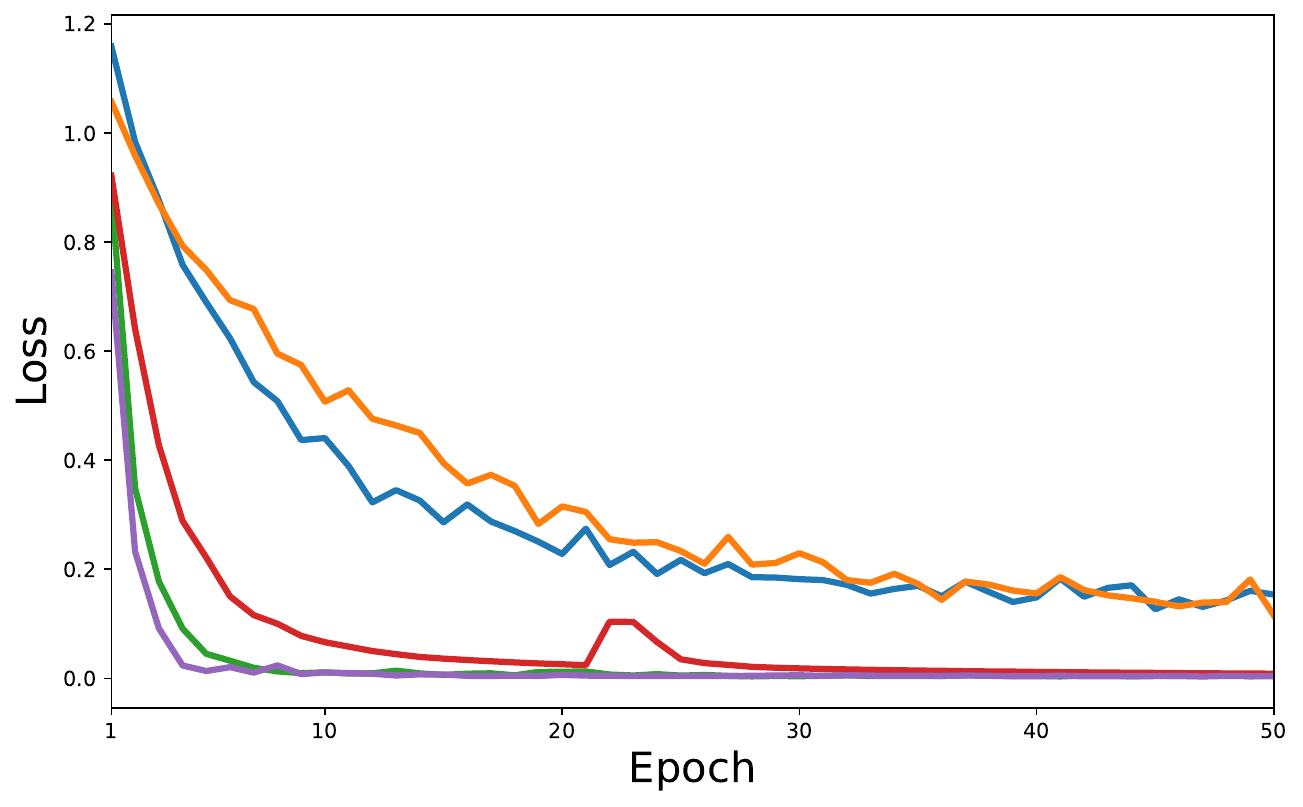}
        }
        \subfigure[Chest CT]{
            \includegraphics[width=0.31\textwidth]{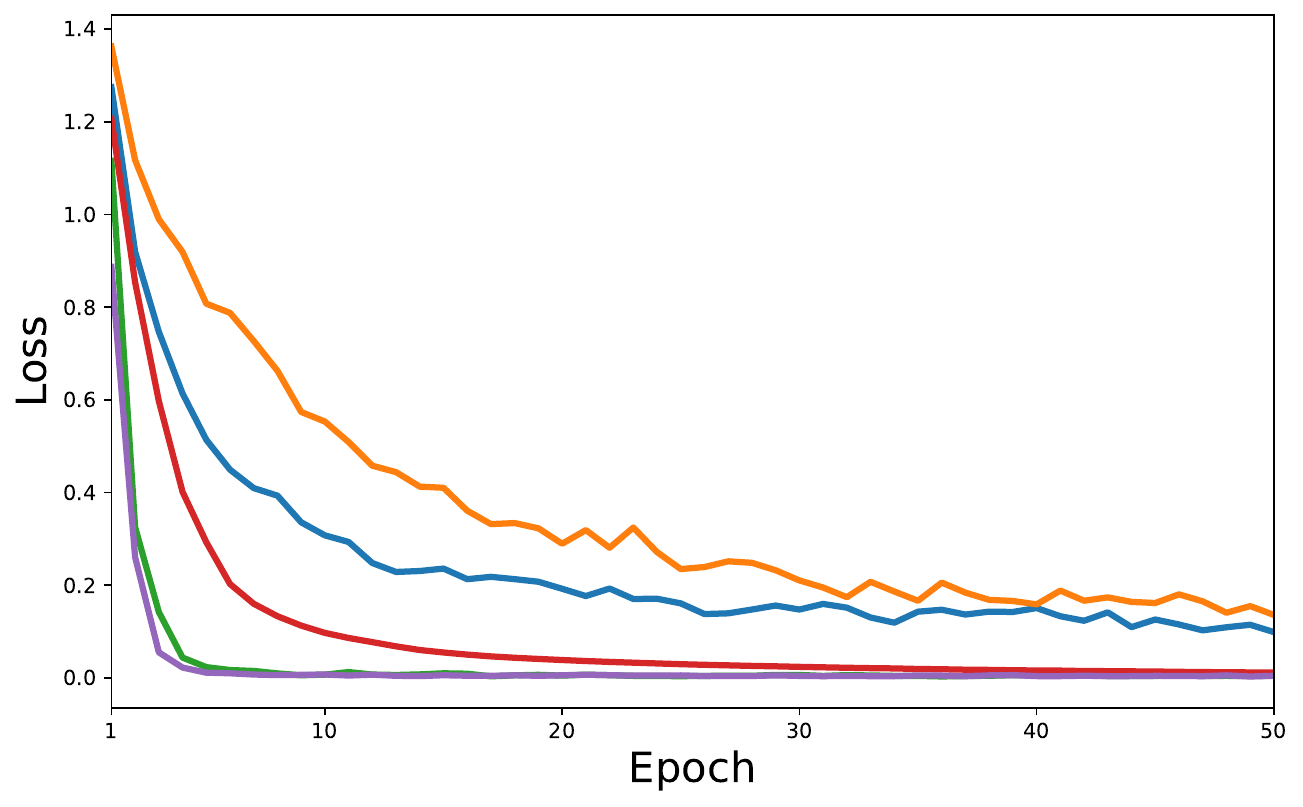}
        }
        \subfigure[Brain MRI ]{
            \includegraphics[width=0.31\textwidth]{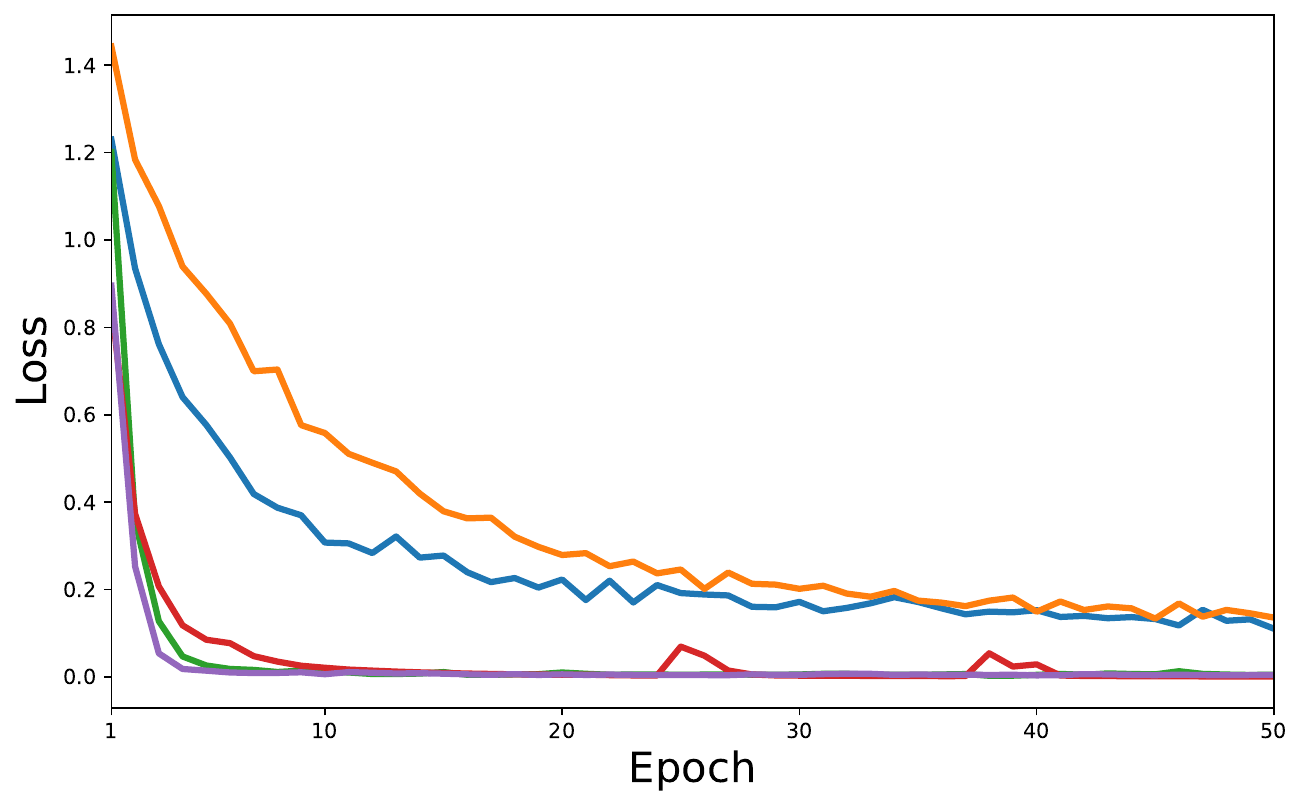}
        }\\
        \subfigure[Gastrointestinal endoscopy]{
            \includegraphics[width=0.31\textwidth]{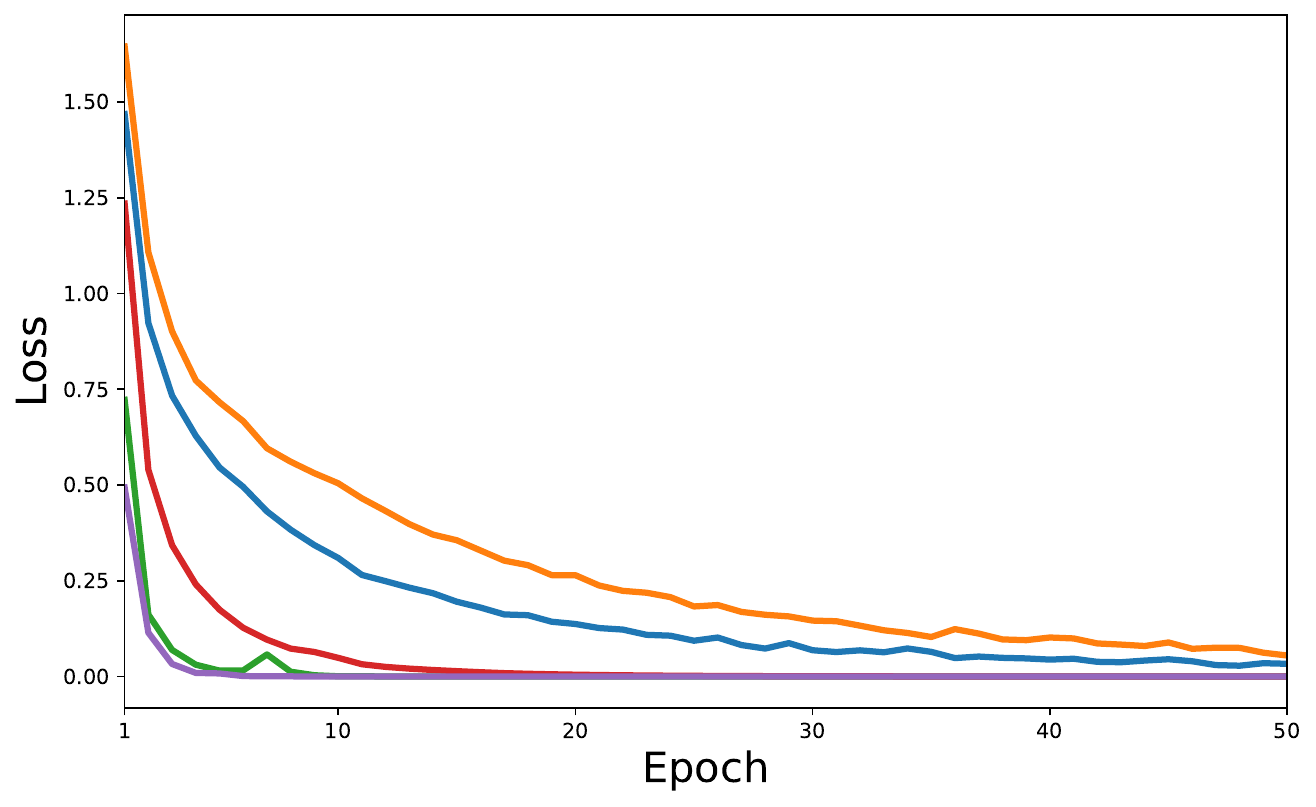}
        }
        \subfigure{
            \includegraphics[width=0.16\textwidth]{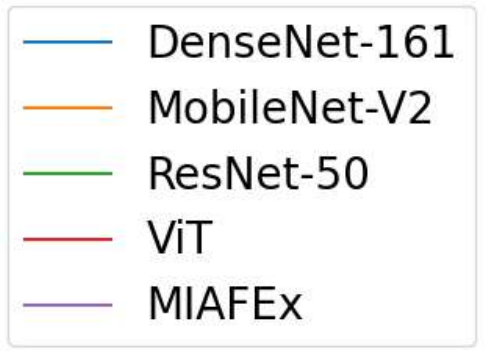}  
        }
        \caption{Training loss curves of the tested models and MIAFEx.}
      \label{fig:loss_curves}
    \end{figure}

First, the training performance of the MIAFEx and the DL models is compared through their loss curves to provide insights into the optimization behavior and convergence rates of each method. Figure \ref{fig:loss_curves} illustrates the training loss curves for MIAFEx and the tested DL models across all the datasets.

\newpage
Notably, MIAFEx demonstrates a rapid convergence across all datasets, reflecting its capacity to effectively learn discriminative features even in the presence of limited training data. In contrast, the curves for conventional DL models, such as MobileNet-V2 and ResNet-50, exhibit slower convergence and higher variability, particularly on smaller datasets like Histological Biopsy and Ocular Alignment. This suggests that these models are more prone to overfitting in these datasets. Moreover, the lower final loss values achieved by MIAFEx across most datasets underline its robustness and ability to capture salient features with minimal residual error. In larger datasets, such as Brain MRI and Gastrointestinal Endoscopy, models like DenseNet-161 and ViT demonstrate competitive performance, with loss curves nearing those of MIAFEx. However, MIAFEx maintains a slight edge in achieving faster convergence, indicating its efficiency in adapting to diverse datasets. 

Then, the performance of various DL models was evaluated and compared against the MIAFEx along with different ML classifiers. The results, summarized in Tables \ref{tab:combined} and \ref{tab:combined2}, provide information regarding the performance of these models in different dataset sizes.

\begin{table}[!ht]
  \centering
  \caption{Performance results of tested models on different datasets.}
  \scalebox{0.77}{
  \begin{tabular}{cccccc}
    \midrule
    \textbf{Dataset} & \textbf{Method} & \textbf{Accuracy} & \textbf{Precision} & \textbf{Recall} & \textbf{F1-Score} \\
    \midrule
    \multirow{10}{*}{Histological biopsy} 
     & MobileNet-V2   & 90.78 & 91.11 & 90.78 & 90.49 \\
    & DenseNet-161   & 77.63 & 78.83 & 77.63 & 77.53 \\
    & ResNet-50      & 73.68 & 73.34 & 73.68 & 72.93 \\
    & ViT            & 84.21& 84.20 & 84.21 & 83.56 \\
    & MIAFEx + LR             & 87.50 & 90.62& 87.50 & 86.83 \\
    & MIAFEx + RF             & 75.00 & 77.08 & 75.00 & 73.40 \\
    & MIAFEx + SVM            & 81.25 & 82.81 & 81.25 & 80.75 \\
    & MIAFEx + XGBoost        & 75.00 & 77.08 & 75.00 & 73.40 \\
    & MIAFEx + WFS-PSO        & 93.75 & 94.64 & 93.75 & 93.64 \\
    & MIAFEx + WFS-DE         & \textbf{96.00} & \textbf{96.29} & \textbf{96.00} & \textbf{95.79} \\
    & MIAFEx + WFS-GA         & 92.00 & 93.07 & 92.00 & 90.95 \\
    \midrule
    \multirow{10}{*}{Ocular alignment} 
     & MobileNet-V2   & 62.74 & 64.97 & 62.74 & 63.27 \\
    & DenseNet-161   & 63.72 & 64.75 & 63.72 & 63.05 \\
    & ResNet-50      & 62.74 & 62.05 & 62.74 & 61.72\\
    & ViT            & 62.74 & 62.27 & 62.74 & 62.23 \\
    & MIAFEx + LR             & 57.14 & 66.55 & 57.14 & 56.60 \\
    & MIAFEx + RF             & 61.90 & 49.39 & 61.90 & 54.76 \\
    & MIAFEx + SVM            & 57.14 & 61.24 & 52.38 & 50.18 \\
    & MIAFEx + XGBoost        & 52.38 & 60.00 & 52.38 & 51.32 \\
    & MIAFEx + WFS-PSO        & 90.14 & 90.79 & 90.14 & 90.13 \\
    & MIAFEx + WFS-DE         & \textbf{95.77} & \textbf{96.08} & \textbf{95.77} & \textbf{95.68} \\
    & MIAFEx + WFS-GA         & 94.36 & 94.47 & 94.37 & 94.37 \\
    \midrule
    \multirow{10}{*}{Eye fundus} 
    & MobileNet-V2   & 56.19 & 58.56 & 56.19 & 57.04 \\
    & DenseNet-161   & 66.11 & 64.87 & 66.11 & 65.22 \\
    & ResNet-50      & 57.85 & 56.91 & 57.50 & 57.24 \\
    & ViT            & 67.76 & 67.09 & 67.76 & 66.43 \\
    & MIAFEx + LR             & 80.00 & 83.00 & 80.00 & 79.42 \\
    & MIAFEx + RF             & 68.00 & 73.80 & 68.00 & 63.37 \\
    & MIAFEx + SVM            & 72.00 & 64.71 & 72.00 & 66.17 \\
    & MIAFEx + XGBoost        & 72.00 & 61.60 & 72.00 & 66.17 \\
    & MIAFEx + WFS-PSO        & 92.00 & 93.07 & 92.00 & 91.71 \\
    & MIAFEx + WFS-DE         & \textbf{96.00} & \textbf{96.29} & \textbf{96.00} & \textbf{95.79} \\
    & MIAFEx + WFS-GA         & 92.00 & 93.07 & 92.00 & 90.95 \\
    \midrule
    \multirow{10}{*}{Breast ultrasound} 
     & MobileNet-V2   & 86.95 & 87.45 & 86.95 & 87.00 \\
    & DenseNet-161   & 86.95 & 87.17 & 86.95 & 87.00 \\
    & ResNet-50      & 91.92 & 91.91 & 91.92 & 91.91 \\
    & ViT            & 88.81 & 89.22 & 88.81 & 88.80 \\
    & MIAFEx + LR             & 81.82 & 81.85 & 81.82 & 81.11 \\
    & MIAFEx + RF             & 78.79 & 84.50 & 78.79 & 76.76 \\
    & MIAFEx + SVM            & 75.76 & 78.16 & 75.76 & 74.52 \\
    & MIAFEx + XGBoost        & 72.73 & 72.73 & 72.73 & 72.73 \\
    & MIAFEx + WFS-PSO        & 93.93 & 94.09 & 93.94 & 93.81 \\
    & MIAFEx + WFS-DE         & 90.90 & 90.90 & 90.90 & 90.90 \\
    & MIAFEx + WFS-GA         & \textbf{96.97} & \textbf{96.97} & \textbf{97.27} & \textbf{96.88} \\
    \bottomrule
  \end{tabular}
  }
  \label{tab:combined}
\end{table}

\begin{table}[!ht]
  \centering
  \caption{Performance results of tested models on different datasets.}
  \scalebox{0.75}{
  \begin{tabular}{cccccc}
    \midrule
    \textbf{Dataset} & \textbf{Method} & \textbf{Accuracy} & \textbf{Precision} & \textbf{Recall} & \textbf{F1-Score} \\
    \midrule
    \multirow{10}{*}{Chest CT} 
    & MobileNet-V2   & 79.66 & 84.56 & 79.66 & 79.38 \\
    & DenseNet-161   & 88.98 & 89.19 & 88.98 & 88.99 \\
    & ResNet-50      & 77.40 & 78.45 & 77.40 & 77.30 \\
    & ViT            & 85.55 & 86.69 & 85.55 & 85.55 \\
    & MIAFEx + LR             & 87.32 & 87.54 & 87.32 & 87.11  \\
    & MIAFEx + RF             & 78.87 & 82.38 & 78.87 & 78.22 \\
    & MIAFEx + SVM            & 85.92 & 78.43 & 76.06 & 76.02 \\
    & MIAFEx + XGBoost        & 80.28 & 82.09 & 80.28 & 79.87 \\
    & MIAFEx + WFS-PSO        & 88.73 & 90.04 & 88.73 & 88.86 \\
    & MIAFEx + WFS-DE         & \textbf{95.77} & 95.98 & \textbf{95.77} & \textbf{95.78} \\
    & MIAFEx + WFS-GA         & \textbf{95.77} & \textbf{96.20} & \textbf{95.77} & \textbf{95.78} \\
    \midrule
    \multirow{10}{*}{Brain MRI} 
    & MobileNet-V2   & 81.35 & 85.33 & 81.35 & 81.29 \\
    & DenseNet-161   & 90.11 & 90.79 & 90.11 & 90.11 \\
    & ResNet-50      & 82.20 & 82.77 & 82.20 & 82.24 \\
    & ViT            & 96.78 & 96.80 & 96.78 & 96.78 \\
    & MIAFEx + LR            & 93.33 & 93.61 & 93.37 & 93.27  \\
    & MIAFEx + RF             & 89.63 & 89.67 & 89.63 & 89.59 \\
    & MIAFEx + SVM            & 90.37 & 90.74 & 90.37 & 90.37  \\
    & MIAFEx + XGBoost        & 91.11 & 91.36 & 91.11 & 91.36 \\
    & MIAFEx + WFS-PSO        & 96.29 & 96.29 & 96.29 & 96.29 \\
    & MIAFEx + WFS-DE         & 95.55 & 95.55 & 95.55 & 95.55 \\
    & MIAFEx + WFS-GA         & \textbf{98.51} & \textbf{98.51} & \textbf{98.51} & \textbf{98.51} \\
    \midrule
    \multirow{10}{*}{Gastrointestinal endoscopy} 
    & MobileNet-V2   & 91.87 & 91.88 & 91.87 & 91.84 \\
    & DenseNet-161   & 91.75 & 91.19 & 91.75 & 91.72 \\
    & ResNet-50      & 92.87 & 92.87 & 92.87 & 92.87 \\
    & ViT            & \textbf{93.45} & \textbf{93.59} & \textbf{93.45} & \textbf{93.45} \\
    & MIAFEx + LR             & 88.12 & 88.94 & 88.12 & 88.16 \\
    & MIAFEx + RF             & 76.88 & 77.17 & 76.88 & 76.52 \\
    & MIAFEx + SVM            & 81.88 & 81.70 & 81.88 & 81.58 \\
    & MIAFEx + XGBoost        & 75.00 & 75.63 & 75.00 & 75.04 \\
    & MIAFEx + WFS-PSO        & 83.75 & 84.98 & 83.75 & 83.50 \\
    & MIAFEx + WFS-DE         & 88.12 & 88.36 & 88.12 & 88.03 \\
    & MIAFEx + WFS-GA         & 87.50 & 88.63 & 87.50 & 87.19 \\
    \bottomrule
  \end{tabular}
  }
  \label{tab:combined2}
\end{table}

The results indicate that optimized models, particularly those using the MIAFEx framework combined with WFS-based algorithms (DE and GA), consistently outperform CNNs and ViT across multiple medical imaging datasets. This pattern is especially prominent in smaller datasets, such as histological biopsy and ocular alignment, where MIAFEx with optimization achieves notably high accuracies (up to 96\%) compared to the other models. For datasets with larger sample sizes, such as chest CT and brain MRI, ViT presents a competitive or even superior performance, but its performance is still below WFS-based methods. However, when the dataset size reaches 4000 images, the ViT model reaches higher performance than MIAFEx-based ones, suggesting that attention mechanisms in ViTs are highly effective for capturing complex textures in large datasets.

Particularly, for the Chest CT dataset, MIAFEx combined with wrapper-based feature selection using DE and GA achieved the highest accuracies (95.77\%), outperforming all deep learning baselines, including ResNet-50 (88.98\%) and ViT (85.55\%). This confirms the robustness of our refinement mechanism, particularly in medium-sized datasets with complex anatomical features. In the Brain MRI dataset, ViT performed strongly (96.78\%), yet MIAFEx combined with WFS-GA surpassed them, reaching 98.51\%. This result highlights the capacity of our approach to extract highly discriminative features even in relatively larger datasets, suggesting that the refinement mechanism generalizes well when integrated with classical classifiers and feature selection. Conversely, in the Gastrointestinal Endoscopy dataset, which contains 4000 images and eight classes, ViT achieved the best overall performance (93.45\%), slightly outperforming MIAFEx variants (e.g., 88.12\% with WFS-DE, and 87.50\% with WFS-GA). This suggests that in very large datasets, full end-to-end training of transformer-based models may offer better performance due to their global modeling capabilities.

Across all tested datasets, models using WFS-DE and WFS-GA optimizations maintain robust Precision, Recall, and F1-scores, emphasizing that feature selection through optimization significantly enhances model reliability in diverse imaging tasks. Overall, the combination of feature extraction and optimization proves advantageous, particularly in smaller datasets, while ViTs offer an alternative for large-scale medical datasets.

\subsection{Dataset size performance analysis}
\begin{figure}[!h]
    \centering
    \includegraphics[width=0.8\linewidth]{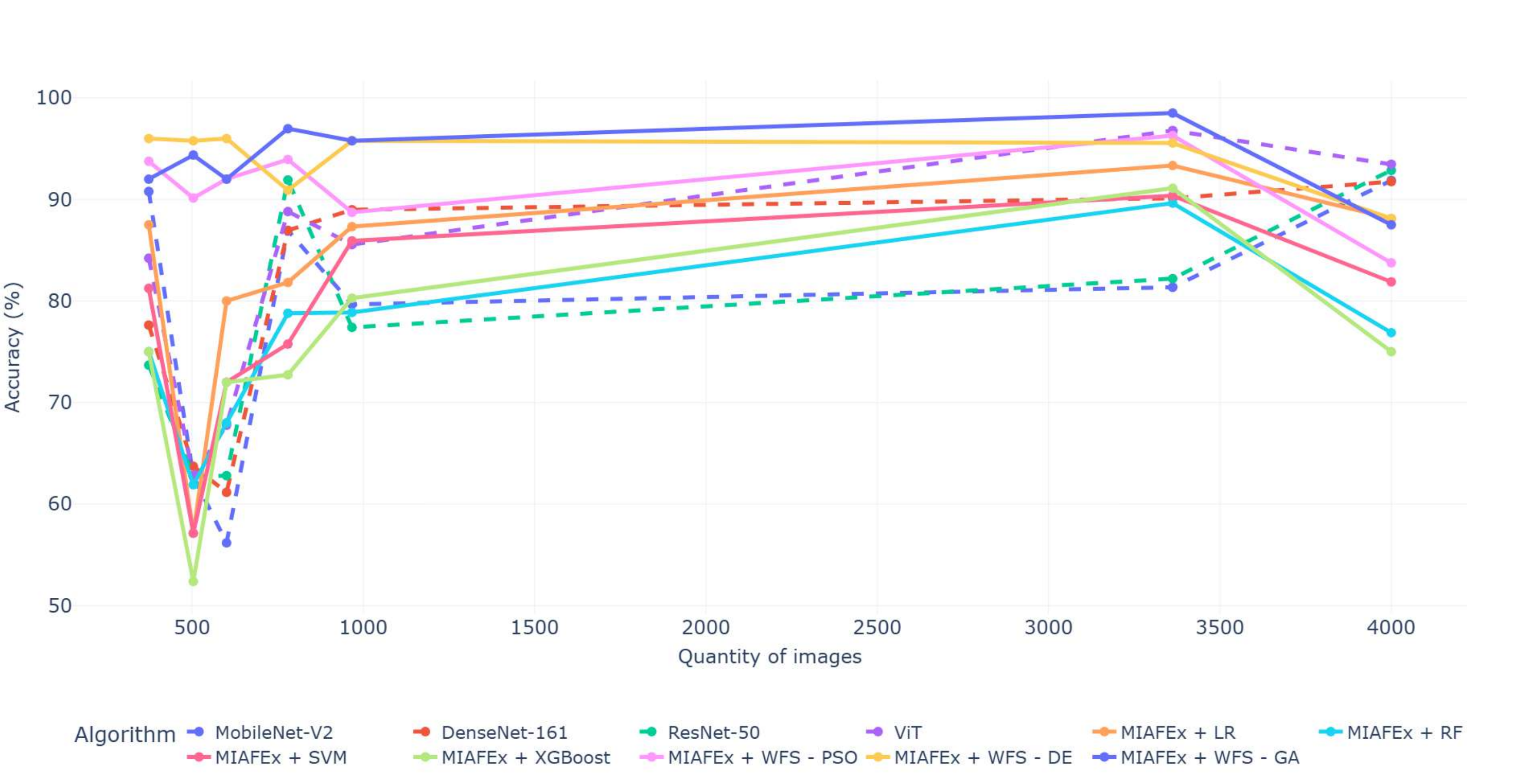}
    \caption{Classification performance of tested models on different image numbers.}
    \label{fig:class_number}
\end{figure}

To further study the impact of the dataset size on the tested models, the results in Fig. \ref{fig:class_number} provide an analysis of the accuracy performance of the tested models across different dataset sizes. 

For smaller datasets, particularly those with fewer than 1000 samples, the MIAFEx-based approaches demonstrate significantly higher performance compared to traditional models like MobileNet-V2, DenseNet-161, ResNet-50, and even the ViT model. For instance, with just 374 samples, MIAFEx with PSO achieves an impressive accuracy of 96\%, while the highest-performing CNN, MobileNet-V2, only reaches 90.78\%. This trend continues with datasets of 504 and 601 samples, where MIAFEx-based methods consistently outperform the CNNs and ViT, showcasing MIAFEx's capability to provide great resources on limited data effectively. However, as the dataset size increases beyond approximately 1000 samples, the performance trend changes. 

Overall, these results reinforce the central hypothesis of this work: MIAFEx offers strong and competitive feature extraction performance, particularly in small and medium medical datasets. In large-scale scenarios, while MIAFEx remains effective, pure transformer-based end-to-end models may have an advantage. These observations support the complementary nature of our hybrid approach.
Larger datasets see traditional CNNs and ViT start to yield superior accuracy, with ViT achieving an accuracy of 96.78\% at the 3362 images, but still being surpassed by MIAFEx + WFS-GA. When the dataset size reaches 4000 images, the ViT model and CNNs outperform all  MIAFEx-based models. This suggests that while MIAFEx excels in scenarios with smaller datasets, the larger datasets provide sufficient information for DL models to optimize their architectures and learning capacities, hence achieving greater overall accuracy.

A key limitation is the imbalance in pretraining datasets across architectures. While ViT models benefit from pretraining on ImageNet-21K, most classical CNN baselines are restricted to ImageNet-1K. Only ResNet-50 was evaluated under both settings to partially mitigate this discrepancy. This difference is important, as the pretraining dataset scale can significantly affect downstream performance.
These findings suggest that MIAFEx offers an efficient approach for applications with smaller medical datasets, while DL models remain advantageous when datasets include a large image quantity. This comparison highlights the importance of dataset size in model selection and suggests that a hybrid approach might be beneficial. Applying MIAFEx for smaller medical datasets provides higher performance than traditional feature extractors and DL models.

\color{black}
\subsection{Refinement mechanism impact}

Regarding the performance when combined with different classifiers, Table \ref{tab:ablation_study} presents a summary of the best-performing pairing for each dataset, contrasting the strongest pipeline based on raw [CLS] embeddings against its MIAFEx counterpart under identical data splits, optimizer, and evaluation protocol. Results of the complete per-class metrics and additional baselines across all classifiers (LR, RF, SVM, KNN, GB, WFS-PSO/DE/GA) are provided in Tables~\ref{tab:combined_ablation} and~\ref{tab:combined_ablation_2} (\ref{app2}) for completeness.

\begin{table}[htbp]
  \centering
  \caption{\textcolor{black}{Ablation study results across datasets for MIAFEx with and without refinement mechanism.}}
  \scalebox{0.8}{
  \begin{tabular}{ccccccc}
    \midrule
    \textbf{Dataset} & \textbf{Best configuration} & \textbf{Accuracy} & \textbf{Precision} & \textbf{Recall} & \textbf{F1-Score}\\
    \midrule
    \multirow{2}{*}{Histological biopsy}
    & Raw [CLS] embeddings + WFS-PSO        & 68.75 & 75.89 & 68.75 & 67.54 \\
    & MIAFEx + WFS-DE        & \textbf{96.00} & \textbf{96.29} & \textbf{96.00} & \textbf{95.79} \\
    \midrule
    \multirow{2}{*}{Ocular alignment}
    & Raw [CLS] embeddings + WFS-PSO        & 66.67 & 74.60 & 66.67 & 66.67 \\
    & MIAFEx + WFS-DE         & \textbf{95.77} & \textbf{96.08} & \textbf{95.77} & \textbf{95.68} \\
    \midrule
    \multirow{2}{*}{Eye fundus}
    & Raw [CLS] embeddings + WFS-GA          & 68.00 & 73.63 & 68.00 & 65.50 \\
    & MIAFEx + WFS-DE         & \textbf{96.00} & \textbf{96.29} & \textbf{96.00} & \textbf{95.79} \\
    \midrule
    \multirow{2}{*}{Breast ultrasound}
    & Raw [CLS] embeddings + WFS-PSO         & 75.76 & 78.83 & 75.76 & 74.16 \\
    & MIAFEx + WFS-GA        & \textbf{96.97} & \textbf{96.97} & \textbf{97.27} & \textbf{96.88} \\
    \midrule
    \multirow{2}{*}{Chest CT}
    & Raw [CLS] embeddings + WFS-GA          & 87.32 & 88.94 & 87.32 & 87.03 \\
    & MIAFEx + WFS-GA         & \textbf{95.77} & \textbf{95.98} & \textbf{95.77} & \textbf{95.78} \\
    \midrule
    \multirow{2}{*}{Brain MRI}
    & Raw [CLS] embeddings + WFS-GA          & 88.73 & 89.82 & 88.73 & 88.38 \\
    & MIAFEx + WFS-GA         & \textbf{98.51} & \textbf{98.51} & \textbf{98.51} & \textbf{98.51} \\
    \midrule
    \multirow{2}{*}{Gastrointestinal endoscopy}
    & Raw [CLS] embeddings + LR              & 85.00 & 85.68 & 85.00 & 85.02 \\
    & MIAFEx + WFS-DE         & \textbf{88.12} & \textbf{88.36} &\textbf{ 88.12 }& \textbf{88.03} \\
    \bottomrule
  \end{tabular}
  }
  \label{tab:ablation_study}
\end{table}

Across the seven datasets with reported values, the MIAFEx representation, typically combined with wrapper-based differential evolution or genetic search (WFS-DE/WFS-GA), consistently attains the top results, outperforming the best raw [CLS] pipelines (often WFS-PSO/WFS-GA).

The gains are largest on histological biopsy, ocular alignment, eye fundus, and breast ultrasound, where absolute improvements often reach $\approx$ 20–30 points, remain meaningful on already strong baselines such as chest CT and brain MRI ($\approx$ 8–10 points), and are modest yet consistent for gastrointestinal endoscopy ($\approx$ 3 points). Because the raw-CLS rows already pair features with the best wrapper (WFS-PSO/DE/GA or LR) for each dataset, the systematic uplift indicates that the learned element-wise refinement of the [CLS] representation is the principal driver of performance rather than the choice of optimizer, yielding balanced error reductions without trading precision for recall.  
\color{black}

\subsection{Comparison with CLIP-Based Feature Extraction}

To contextualize the performance of the proposed MIAFEx method, additional experiments were conducted using CLIP (\textit{Contrastive Language-Image Pretraining}) \cite{li2021supervision} as a baseline for feature extraction. CLIP is a robust multimodal model pre-trained on approximately 400 million image-text pairs from a variety of public sources, using contrastive learning to align visual and linguistic embeddings, enabling robust zero-shot capabilities across a wide range of natural image domains. The model used in this study corresponds to the ViT-B/32 variant from OpenAI.

\begin{figure}
    \centering
    \includegraphics[width=\linewidth]{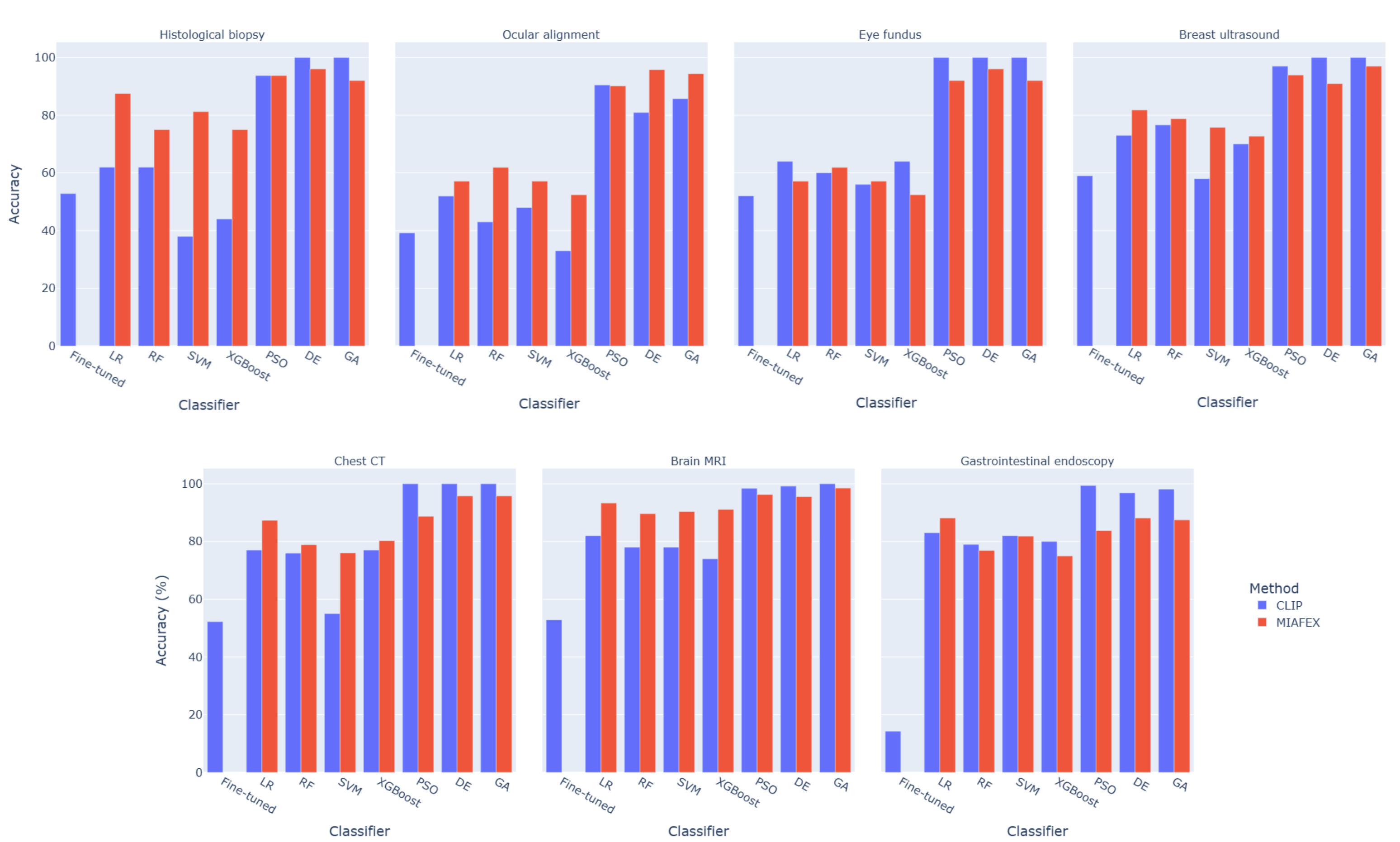}
    \caption{Accuracy comparison of CLIP vs MIAFEx across all datasets.}
    \label{fig:clip_miafex_accomp}
\end{figure}

Two evaluation strategies were implemented: (1) direct fine-tuning of CLIP on the target medical image datasets, and (2) extraction of frozen visual features using CLIP followed by downstream classification with traditional ML algorithms and metaheuristic-based feature selection. A summary of those experiment results is shown in Fig. \ref{fig:clip_miafex_accomp}.

The first approach (Fine-tuned) consistently failed to converge across all datasets. Attempts to fine-tune CLIP on the available medical datasets resulted in poor convergence and significantly reduced performance. This outcome is likely attributed to the limited size and specificity of the datasets, which contrasts starkly with the broad and large-scale training regimen that CLIP originally underwent. Fine-tuning a billion-parameter model on datasets with fewer than 500 samples per class introduces overfitting risks and optimization challenges. Medical image datasets used in this study are considerably smaller in size and do not provide sufficient diversity or semantic alignment to support such adaptation.

The second approach, utilizing CLIP as a fixed feature extractor, yielded variable results. When the extracted embeddings were directly classified using standard models such as LR, RF, SVM, and XGBoost, performance was low to moderate and dataset-dependent. However, when CLIP embeddings were processed through a metaheuristic feature selection pipeline (e.g., GA) followed by a $k$-NN, classification accuracy increased dramatically: frequently reaching 100\% across multiple datasets. When used as a frozen feature extractor and coupled with metaheuristic optimization techniques such as PSO, DE, or GA, CLIP achieved near-perfect performance on all datasets. In contrast, classical classifiers applied directly to CLIP features, such as SVM, RRF, and LR, exhibited moderate to poor results. This discrepancy suggests that while CLIP embeddings are semantically rich, they may not be linearly separable for medical tasks without a specialized feature selection or transformation stage.

In comparison, MIAFEx was developed specifically for medical imaging tasks and leverages feature extraction informed by supervised training on the ImageNet-21k dataset. Although its pretraining scale is considerably smaller than that of CLIP, MIAFEx achieved competitive or superior performance across most datasets when used in conjunction with classical classifiers. Furthermore, its transparency, reproducibility, and domain-aligned design offer practical advantages for medical applications, particularly where infrastructure constraints are essential.

It is also worth noting that CLIP’s pretraining process and training data are proprietary and not publicly reproducible, which imposes limitations on transparency and clinical adoption. By contrast, MIAFEx is designed with interpretability and accessibility in mind, fostering trust and deployment in healthcare environments.  While CLIP's embeddings exhibit strong generalization when refined by selection methods, their utility in direct classification tasks without further adaptation appears limited. Moreover, the proprietary and large-scale nature of CLIP's training corpus introduces reproducibility and interpretability concerns in clinical applications. In contrast, MIAFEx offers a lightweight, modular, and reproducible pipeline designed specifically for the constraints and requirements of medical imaging tasks.

These results highlight two key observations. First, CLIP exhibits strong representational capacity when augmented with domain-specific selection mechanisms, although this success is contingent on the availability of curated post-processing. Second, there remains a significant discrepancy in model provenance: CLIP benefits from opaque pre-training on hundreds of millions of natural images, while MIAFEx is explicitly tailored for low-resource medical settings using transparent and reproducible modules. Importantly, MIAFEx achieves high accuracy without requiring access to proprietary pre-training datasets or language supervision, ensuring greater feasibility for clinical translation.

A complete comparative summary of classification results for all datasets is presented in Tables~\ref{tab:clip_miafex} and \ref{tab:clip_miafex2}, annexed to \ref{app3}. CLIP-based pipelines achieve perfect accuracy only when combined with feature selection techniques, while raw embeddings remain less effective. In contrast, MIAFEx provides competitive performance under consistent settings.


\section{Conclusions}
\label{sec:conclusions}
The paper introduces the MIAFEx, a novel attention-based feature extraction method specifically designed to address dataset-related challenges in medical image classification. By using a refinement mechanism within the transformer encoder, MIAFEx optimizes the classification token to focus on the most relevant features of medical images. This approach addresses limitations of both classical feature extraction techniques and modern deep learning models, particularly in scenarios with limited training data and high intra-class variability, which are common in medical imaging.

Through comprehensive experiments across diverse medical imaging datasets, MIAFEx demonstrates notable improvements in accuracy, precision, recall, and F1-score over traditional feature extractors and popular DL methods, including CNNs and ViT. The integration of metaheuristic optimization techniques with MIAFEx further enhances classification performance, particularly in small datasets where standard models often struggle with overfitting. 

One of the main strengths of this method is its adaptability across a wide range of imaging modalities and dataset sizes. Experimental results demonstrated that MIAFEx consistently outperforms both classical handcrafted descriptors and deep learning models (CNNs and ViT) on small datasets. The incorporation of wrapper-based feature selection with metaheuristics further improves the classification accuracy, confirming MIAFEx’s potential as a reliable solution in data-constrained medical environments. However, some limitations must be acknowledged. The method’s performance advantage diminishes as the dataset size increases, where conventional ViTs and CNNs eventually surpass MIAFEx. Additionally, the refinement mechanism introduces additional parameters, which may require fine-tuning depending on the dataset characteristics.

From a broader perspective, this work provides a valuable tool in medical image analysis. MIAFEx addresses a critical gap in current methodologies. Its lightweight and adaptable design makes it particularly well-suited for integration into clinical decision-support systems, especially in settings where large-scale annotated datasets are unavailable or impractical to obtain.

In conclusion, MIAFEx proves to be an effective and versatile feature extraction method for medical image classification, with significant potential for early disease detection and diagnostic applications, setting a new benchmark for feature extraction in the field. Its ability to outperform existing methods on limited data suggests that MIAFEx could be highly impactful in real-world medical settings where data scarcity and variability are prevalent. Future work could explore further enhancements, scalability to larger datasets, and applicability across additional medical imaging domains.

\newpage
\appendix
\section{Performance of classical feature extractors and MIAFEx with ML classifiers}
\label{app1}
\begin{table}[htbp]
  \centering
  \caption{Performance on feature extraction on the histological biopsy dataset.}
  \scalebox{0.8}{
\begin{tabular}{cccccrccccc}
    \multicolumn{5}{c}{\textbf{SVM}}      &       & \multicolumn{5}{c}{\textbf{RF}} \\
\cmidrule{1-5}\cmidrule{7-11}    \textbf{Feature extractor} & \textbf{Accuracy} & \textbf{Precision} & \textbf{ Recall} & \textbf{ F1-Score} &       & \textbf{Feature extractor} & \textbf{Accuracy} & \textbf{Precision} & \textbf{ Recall} & \textbf{ F1-Score} \\
\cmidrule{1-5}\cmidrule{7-11}    \textbf{HOG} & 50.00 & 50.00 & 50.00 & 46.91 &       & \textbf{HOG} & 56.25 & 87.41 & 54.44 & 53.70 \\
\cmidrule{1-5}\cmidrule{7-11}    \textbf{SIFT} & 43.75 & 42.97 & 43.75 & 41.51 &       & \textbf{SIFT} & 31.25 & 35.04 & 31.25 & 32.85 \\
\cmidrule{1-5}\cmidrule{7-11}    \textbf{LBP} & 37.75 & 14.06 & 37.50 & 20.45 &       & \textbf{LBP} & 43.75 & 29.95 & 43.75 & 35.40 \\
\cmidrule{1-5}\cmidrule{7-11}    \textbf{HTFs} & 43.75 & 45.83 & 43.33 & 42.10 &       & \textbf{Haralick} & 43.75 & 42.06 & 43.33 & 41.90 \\
\cmidrule{1-5}\cmidrule{7-11}    \textbf{GF} & 56.25 & 56.25 & 56.25 & 56.25 &       & \textbf{Gabor} & 56.25 & 54.17 & 56.25 & 54.24 \\
\cmidrule{1-5}\cmidrule{7-11}    \textbf{MIAFEx} & \textbf{81.25} & \textbf{82.81} & \textbf{81.25} & \textbf{80.75} &       & \textbf{MIAFEx} & \textbf{75.00} & \textbf{77.08} & \textbf{75.00} & \textbf{73.40} \\
\end{tabular}%
}
  \label{tab:apphistol}%
\end{table}%

\begin{table}[htbp]
  \centering
  \caption{Performance on feature extraction on the ocular alignment dataset.}
  \scalebox{0.8}{
    \begin{tabular}{cccccrccccc}
    \multicolumn{5}{c}{\textbf{SVM}}      &       & \multicolumn{5}{c}{\textbf{RF}} \\
\cmidrule{1-5}\cmidrule{7-11}    \textbf{Feature extractor} & \textbf{Accuracy} & \textbf{Precision} & \textbf{ Recall} & \textbf{ F1-Score} &       & \textbf{Feature extractor} & \textbf{Accuracy} & \textbf{Precision} & \textbf{ Recall} & \textbf{ F1-Score} \\
\cmidrule{1-5}\cmidrule{7-11}    \textbf{HOG} & 28.57 & 21.43 & 26.00 & 23.05 &       & \textbf{HOG} & 28.57 & 16.89 & 26.00 & 20.32 \\
\cmidrule{1-5}\cmidrule{7-11}    \textbf{SIFT} & 14.29 & 15.83 & 14.29 & 13.82 &       & \textbf{SIFT} & 28.57 & 22.59 & 28.57 & 25.13 \\
\cmidrule{1-5}\cmidrule{7-11}    \textbf{LBP} & 28.57 & 17.78 & 28.00 & 17.27 &       & \textbf{LBP} & 28.57 & 34.00 & 28.00 & 30.00 \\
\cmidrule{1-5}\cmidrule{7-11}    \textbf{HTFs} & 23.81 & 34.29 & 23.81 & 25.32 &       & \textbf{HTFs} & 28.57 & 22.04 & 28.57 & 24.87 \\
\cmidrule{1-5}\cmidrule{7-11}    \textbf{GF} & 28.57 & 21.27 & 28.57 & 23.96 &       & \textbf{GF} & 19.05 & 22.22 & 19.05 & 19.77 \\
\cmidrule{1-5}\cmidrule{7-11}    \textbf{MIAFEx} & \textbf{57.14} & \textbf{61.24} & \textbf{52.38} & \textbf{50.18} &       & \textbf{MIAFEx} & \textbf{61.90} & \textbf{49.39} & \textbf{61.90} & \textbf{54.76} \\
\end{tabular}%
}
  \label{tab:appstrab}%
\end{table}%

\begin{table}[htbp]
  \centering
  \caption{Performance on feature extraction on the eye fundus dataset.}
  \scalebox{0.8}{
    \begin{tabular}{cccccrccccc}
    \multicolumn{5}{c}{\textbf{SVM}}      &       & \multicolumn{5}{c}{\textbf{RF}} \\
\cmidrule{1-5}\cmidrule{7-11}    \textbf{Feature extractor} & \textbf{Accuracy} & \textbf{Precision} & \textbf{ Recall} & \textbf{ F1-Score} &       & \textbf{Feature extractor} & \textbf{Accuracy} & \textbf{Precision} & \textbf{ Recall} & \textbf{ F1-Score} \\
\cmidrule{1-5}\cmidrule{7-11}    \textbf{HOG} & 64.00 & 62.73 & 64.00 & 55.70 &       & \textbf{HOG} & 52.00 & 27.04 & 52.00 & 35.58 \\
\cmidrule{1-5}\cmidrule{7-11}    \textbf{SIFT} & 48.00 & 41.25 & 48.00 & 43.81 &       & \textbf{SIFT} & 56.00 & 43.78 & 56.00 & 48.90 \\
\cmidrule{1-5}\cmidrule{7-11}    \textbf{LBP} & 52.00 & 27.04 & 52.00 & 35.58 &       & \textbf{LBP} & 52.00 & 50.58 & 52.00 & 50.56 \\
\cmidrule{1-5}\cmidrule{7-11}    \textbf{HTFs} & 52.00 & 27.04 & 52.00 & 35.58 &       & \textbf{HTFs} & 44.00 & 38.00 & 44.00 & 40.96 \\
\cmidrule{1-5}\cmidrule{7-11}    \textbf{GF} & 32.00 & 32.53 & 32.00 & 32.13 &       & \textbf{GF} & 64.00 & 56.84 & 64.00 & 58.05 \\
\cmidrule{1-5}\cmidrule{7-11}    \textbf{MIAFEx} & \textbf{72.00} & \textbf{64.71} & \textbf{72.00} & \textbf{67.31} &       & \textbf{MIAFEx} & \textbf{68.00} & \textbf{73.80} & \textbf{68.00} & \textbf{63.37} \\
\end{tabular}%
}
  \label{tab:appef}%
\end{table}%

\begin{table}[htbp]
  \centering
  \caption{Performance on feature extraction on the breast ultrasound dataset.}
  \scalebox{0.8}{
    \begin{tabular}{cccccrccccc}
    \multicolumn{5}{c}{\textbf{SVM}}      &       & \multicolumn{5}{c}{\textbf{RF}} \\
\cmidrule{1-5}\cmidrule{7-11}    \textbf{Feature extractor} & \textbf{Accuracy} & \textbf{Precision} & \textbf{ Recall} & \textbf{ F1-Score} &       & \textbf{Feature extractor} & \textbf{Accuracy} & \textbf{Precision} & \textbf{ Recall} & \textbf{ F1-Score} \\
\cmidrule{1-5}\cmidrule{7-11}    \textbf{HOG} & 69.69 & 72.64 & 69.69 & 69.06 &       & \textbf{HOG} & 60.61 & 61.46 & 60.61 & 48.35 \\
\cmidrule{1-5}\cmidrule{7-11}    \textbf{SIFT} & 57.57 & 47.76 & 57.57 & 50.96 &       & \textbf{SIFT} & 57.57 & 47.76 & 57.57 & 50.96 \\
\cmidrule{1-5}\cmidrule{7-11}    \textbf{LBP} & 57.57 & 33.15 & 57.57 & 42.07 &       & \textbf{LBP} & 60.61 & 64.77 & 60.61 & 58.08 \\
\cmidrule{1-5}\cmidrule{7-11}    \textbf{HTFs} & 54.55 & 32.39 & 54.55 & 40.64 &       & \textbf{HTFs} & 54.55 & 54.55 & 54.55 & 53.94 \\
\cmidrule{1-5}\cmidrule{7-11}    \textbf{GF} & 51.52 & 52.53 & 51.51 & 51.95 &       & \textbf{GF} & 54.55 & 55.56 & 54.55 & 54.49 \\
\cmidrule{1-5}\cmidrule{7-11}    \textbf{MIAFEx} & \textbf{75.76} & \textbf{78.16} & \textbf{75.76} & \textbf{74.52} &       & \textbf{MIAFEx} & \textbf{78.79} & \textbf{84.50} & \textbf{78.79} & \textbf{76.76} \\
\end{tabular}%
}
  \label{tab:appbusi}%
\end{table}%

\begin{table}[htbp]
  \centering
  \caption{Performance on feature extraction on the chest CT dataset.}
  \scalebox{0.8}{
    \begin{tabular}{cccccrccccc}
    \multicolumn{5}{c}{\textbf{SVM}}      &       & \multicolumn{5}{c}{\textbf{RF}} \\
\cmidrule{1-5}\cmidrule{7-11}    \textbf{Feature extractor} & \textbf{Accuracy} & \textbf{Precision} & \textbf{ Recall} & \textbf{ F1-Score} &       & \textbf{Feature extractor} & \textbf{Accuracy} & \textbf{Precision} & \textbf{ Recall} & \textbf{ F1-Score} \\
\cmidrule{1-5}\cmidrule{7-11}    \textbf{HOG} & 81.69 & 82.53 & 81.69 & 81.15 &       & \textbf{HOG} & 71.83 & 71.03 & 71.83 & 69.49 \\
\cmidrule{1-5}\cmidrule{7-11}    \textbf{SIFT} & 57.75 & 54.88 & 57.75 & 54.25 &       & \textbf{SIFT} & 47.89 & 38.49 & 47.89 & 41.25 \\
\cmidrule{1-5}\cmidrule{7-11}    \textbf{LBP} & 38.03 & 14.46 & 38.03 & 20.95 &       & \textbf{LBP} & 73.24 & 74.17 & 73.24 & 72.49 \\
\cmidrule{1-5}\cmidrule{7-11}    \textbf{HTFs} & 67.61 & 65.38 & 67.61 & 62.95 &       & \textbf{HTFs} & 70.42 & 68.92 & 70.42 & 69.15 \\
\cmidrule{1-5}\cmidrule{7-11}    \textbf{GF} & 64.79 & 65.23 & 64.79 & 63..37 &       & \textbf{GF} & 61.97 & 60.10 & 61.97 & 58.59 \\
\cmidrule{1-5}\cmidrule{7-11}    \textbf{MIAFEx} & \textbf{85.92} & \textbf{86.34} & \textbf{85.92} & \textbf{85.88} &       & \textbf{MIAFEx} & \textbf{78.87} & \textbf{82.38} & \textbf{78.87} & \textbf{78.22} \\
\end{tabular}%
}
  \label{tab:appchest}%
\end{table}%

\begin{table}[htbp]
  \centering
  \caption{Performance on feature extraction on the brain MRI dataset.}
  \scalebox{0.8}{
    \begin{tabular}{cccccrccccc}
    \multicolumn{5}{c}{\textbf{SVM}}      &       & \multicolumn{5}{c}{\textbf{RF}} \\
\cmidrule{1-5}\cmidrule{7-11}    \textbf{Feature extractor} & \textbf{Accuracy} & \textbf{Precision} & \textbf{ Recall} & \textbf{ F1-Score} &       & \textbf{Feature extractor} & \textbf{Accuracy} & \textbf{Precision} & \textbf{ Recall} & \textbf{ F1-Score} \\
\cmidrule{1-5}\cmidrule{7-11}    \textbf{HOG} & 74.81 & 74.65 & 74.81 & 74.17 &       & \textbf{HOG} & 64.44 & 68.85 & 64.44 & 62.55 \\
\cmidrule{1-5}\cmidrule{7-11}    \textbf{SIFT} & 53.08 & 57.50 & 53.08 & 56.21 &       & \textbf{SIFT} & 42.31 & 43.68 & 43.21 & 39.96 \\
\cmidrule{1-5}\cmidrule{7-11}    \textbf{LBP} & 40.00 & 42.78 & 40.00 & 33.79 &       & \textbf{LBP} & 69.63 & 69.44 & 69.63 & 69.22 \\
\cmidrule{1-5}\cmidrule{7-11}    \textbf{HTFs} & 57.78 & 57.39 & 57.78 & 55.83 &       & \textbf{HTFs} & 63.70 & 64.42 & 63.70 & 63.65 \\
\cmidrule{1-5}\cmidrule{7-11}    \textbf{GF} & 67.41 & 68.28 & 67.41 & 67.53 &       & \textbf{GF} & 71.85 & 71.61 & 71.85 & 71.50 \\
\cmidrule{1-5}\cmidrule{7-11}    \textbf{MIAFEx} & \textbf{90.37} & \textbf{90.74} & \textbf{90.37} & \textbf{90.37} &       & \textbf{MIAFEx} & \textbf{89.63} & \textbf{89.67} & \textbf{89.63} & \textbf{89.59} \\
\end{tabular}%
}
  \label{tab:appbrain}%
\end{table}%
\begin{table}[htbp]
  \centering
  \caption{Performance on feature extraction on the gastrointestinal endoscopy dataset.}
  \scalebox{0.8}{
    \begin{tabular}{cccccrccccc}
    \multicolumn{5}{c}{\textbf{SVM}}      &       & \multicolumn{5}{c}{\textbf{RF}} \\
\cmidrule{1-5}\cmidrule{7-11}    \textbf{Feature extractor} & \textbf{Accuracy} & \textbf{Precision} & \textbf{ Recall} & \textbf{ F1-Score} &       & \textbf{Feature extractor} & \textbf{Accuracy} & \textbf{Precision} & \textbf{ Recall} & \textbf{ F1-Score} \\
\cmidrule{1-5}\cmidrule{7-11}    \textbf{HOG} & 57.50 & 57.86 & 57.50 & 56.95 &       & \textbf{HOG} & 48.75 & 49.53 & 48.75 & 47.80 \\
\cmidrule{1-5}\cmidrule{7-11}    \textbf{SIFT} & 44.62 & 44.86 & 45.62 & 44.19 &       & \textbf{SIFT} & 39.37 & 38.35 & 39.37 & 38.28 \\
\cmidrule{1-5}\cmidrule{7-11}    \textbf{LBP} & 28.75 & 17.54 & 28.75 & 20.81 &       & \textbf{LBP} & 38.78 & 38.90 & 38.75 & 38.30 \\
\cmidrule{1-5}\cmidrule{7-11}    \textbf{HTFs} & 28.75 & 26.35 & 28.75 & 26.42 &       & \textbf{HTFs} & 38.12 & 38.41 & 38.12 & 37.76 \\
\cmidrule{1-5}\cmidrule{7-11}    \textbf{GF} & 48.75 & 49.02 & 48.75 & 48.00 &       & \textbf{GF} & 41.25 & 40.97 & 41.25 & 40.61 \\
\cmidrule{1-5}\cmidrule{7-11}    \textbf{MIAFEx} & \textbf{72.73} & \textbf{78.16} & \textbf{72.73} & \textbf{74.52} &       & \textbf{MIAFEx} & \textbf{78.79} & \textbf{84.50} & \textbf{78.79} & \textbf{76.76} \\
\end{tabular}%
}
  \label{tab:appkvasir}%
\end{table}%

\newpage
\section{Ablation study of the impact of the refinement mechanism}
\label{app2}

\begin{table}[!ht]
  \centering
  \caption{\textcolor{black}{Ablation study results on different datasets.}}
  \scalebox{0.77}{
  \begin{tabular}{cccccc}
    \midrule
    \textbf{Dataset} & \textbf{Method} & \textbf{Accuracy} & \textbf{Precision} & \textbf{Recall} & \textbf{F1-Score} \\
    \midrule
    \multirow{10}{*}{Histological biopsy} 
     & Raw [CLS] embeddings + LR           & 62.50 & 62.72 & 62.50 & 62.18 \\
    & Raw [CLS] embeddings + RF             & 62.50 & 61.46 & 62.50 & 58.93 \\
    & Raw [CLS] embeddings + SVM            & 43.75 & 46.25 & 43.75 & 42.19 \\
    & Raw [CLS] embeddings + XGBoost        & 50.00 & 50.00 & 50.00 & 49.21 \\
    & Raw [CLS] embeddings + WFS-PSO        & 68.75 & 75.89 & 68.75 & 67.54 \\
    & Raw [CLS] embeddings + WFS-DE         & 68.75 & 78.75 & 68.75 & 63.54 \\
    & Raw [CLS] embeddings + WFS-GA         & 68.75 & 75.52 & 68.75 & 67.37 \\
\bottomrule
  \end{tabular}
  }
  \label{tab:combined_ablation}
\end{table}

\begin{table}[!ht]
  \centering
  \caption{\textcolor{black}{Performance results of tested models on different datasets.}}  \scalebox{0.75}{
  \begin{tabular}{cccccc}
    \midrule
    \textbf{Dataset} & \textbf{Method} & \textbf{Accuracy} & \textbf{Precision} & \textbf{Recall} & \textbf{F1-Score} \\
    \midrule
    \multirow{10}{*}{Chest CT} 
    & Raw [CLS] embeddings + LR              & 73.24 & 73.51 & 73.24 & 72.57 \\
    & Raw [CLS] embeddings + RF              & 66.20 & 72.88 & 66.20 & 61.92 \\
    & Raw [CLS] embeddings + SVM             & 59.15 & 54.34 & 59.15 & 50.29 \\
    & Raw [CLS] embeddings + XGBoost         & 54.93 & 50.77 & 54.93 & 51.97 \\
    & Raw [CLS] embeddings + WFS-PSO         & 87.32 & 88.24 & 87.32 & 87.21 \\
    & Raw [CLS] embeddings + WFS-DE          & 87.32 & 88.26 & 87.32 & 86.89 \\
    & Raw [CLS] embeddings + WFS-GA          & 87.32 & 88.94 & 87.32 & 87.03 \\
\bottomrule
  \end{tabular}
  }
  \label{tab:combined_ablation_2}
\end{table}

\newpage
\section{Performance of CLIP and MIAFEx with ML classifiers}
\label{app3}

\begin{table}[htbp]
  \centering
  \caption{Performance comparison of CLIP and MIAFEx with different classifiers.}
   \scalebox{0.75}{
    \begin{tabular}{ccccccc}
    \midrule
    \textbf{Dataset} & \textbf{Method} & \textbf{Classifier} & \textbf{Accuracy} & \textbf{Precision} & \textbf{Recall} & \textbf{F1}-Score \\
    \midrule
    \multirow{15}[0]{*}{Histological biopsy} & \multirow{8}[0]{*}{CLIP} & Fine-tuned & 52.82 & 48.3  & 52.82 & 39.86 \\
          &       & LR    & 62.00 & 62.00 & 62.00 & 62.00 \\
          &       & RF    & 62.00 & 64.00 & 62.00 & 59.00 \\
          &       & SVM   & 38.00 & 14.00 & 38.00 & 20.00 \\
          &       & XGBoost & 44.00 & 47.00 & 44.00 & 44.00 \\
          &       & PSO   & 93.75 & 94.64 & 93.75 & 93.64 \\
          &       & DE    & \textbf{100.00} & \textbf{100.00} & \textbf{100.00} & \textbf{100.00} \\
\bottomrule
    \end{tabular}}%
  \label{tab:clip_miafex}%
\end{table}%

\begin{table}[htbp]
  \centering
  \caption{Performance comparison of CLIP and MIAFEx with different classifiers.}
  \scalebox{0.75}{
    \begin{tabular}{ccccccc}
    \midrule
    \textbf{Dataset} & \textbf{Method} & \textbf{Classifier} & \textbf{Accuracy} & \textbf{Precision} & \textbf{Recall} & \textbf{F1-Score} \\
    \midrule
    \multirow{15}[0]{*}{Breast ultrasound} & \multirow{8}[0]{*}{CLIP} & Fine-tuned & 59.00 & 61.63 & 59.00 & 44.42 \\
          &       & LR    & 73.00 & 74.00 & 73.00 & 72.00 \\
          &       & RF    & 76.60 & 80.10 & 76.60 & 72.10 \\
          &       & SVM   & 58.00 & 33.00 & 58.00 & 42.00 \\
          &       & XGBoost & 70.00 & 71.10 & 70.00 & 68.80 \\
          &       & PSO   & 96.97 & 97.27 & 96.97 & 96.88 \\
          &       & DE    & \textbf{100.00} & \textbf{100.00} & \textbf{100.00} & \textbf{100.00} \\
\bottomrule
    \end{tabular}}%
  \label{tab:clip_miafex2}%
\end{table}%

\newpage
\bibliographystyle{elsarticle-num}
\bibliography{references}

\end{document}